%% file: main.tex
\documentclass[journal]{IEEEtran}
\usepackage{color}
\usepackage{xcolor}

\ifCLASSOPTIONcompsoc
\usepackage[nocompress]{cite}
\else
\usepackage{cite}
\fi

\usepackage[utf8]{inputenc}
\usepackage{longtable}
\usepackage{graphicx}
\usepackage{color}

\usepackage[cmex10]{amsmath}

\usepackage{array}
\usepackage{mdwmath}
\usepackage{mdwtab}
\usepackage{url}
\usepackage{bm}
\usepackage{hyperref}
\usepackage{wrapfig,booktabs}
\usepackage{multirow}
\usepackage{booktabs}
\usepackage{amssymb}
\usepackage{lscape}

\usepackage{rotating}
\usepackage{cleveref}

\usepackage{xtab}
\usepackage{longtable}
\usepackage{subfigure}

\usepackage{tikz}
\usepackage{pifont}

\usepackage{nicefrac}

\usepackage{algorithm}
\usepackage{algorithmic}

\DeclareGraphicsExtensions{.pdf,.eps}

\input{notations}

\begin{document}
\bstctlcite{IEEEexample:BSTcontrol}


\title{Advances and Trends in the 3D Reconstruction of the Shape and Motion of Animals}

\author{Ziqi Li, \and Abderraouf Amrani, \and Shri Rai, and Hamid Laga
	
	\IEEEcompsocitemizethanks{\IEEEcompsocthanksitem This work was supported by the Australian Research Council through the Discovery under Grant DP220102197. \IEEEcompsocthanksitem Ziqi Li is with the School of Information Technology, Murdoch University (Australia). Email: Ziqi.Li@murdoch.edu.au.
		\IEEEcompsocthanksitem 
		Abderraouf Amrani is with the School of Information Technology, Murdoch University (Australia). Email: a.amrani@murdoch.edu.au, 
		\IEEEcompsocthanksitem Shri Rai is with the School of Information Technology, Murdoch University (Australia). Email: S.Rai@murdoch.edu.au. 
            \IEEEcompsocthanksitem Hamid Laga is with the School of Information Technology, Murdoch University (Australia). Email: H.Laga@murdoch.edu.au. 
	}
	\thanks{Manuscript received XXX; revised December yyy.}
}

\markboth{XXX}%
{Name \MakeLowercase{\etalnospace}: title}

\IEEEcompsoctitleabstractindextext{
\begin{abstract}
Reconstructing the 3D geometry, pose, and motion of animals is a long-standing problem, which has a wide range of applications, from biology, livestock management, and animal conservation and welfare to content creation in digital entertainment and Virtual/Augmented Reality (VR/AR). Traditionally, 3D models of real animals are obtained using 3D scanners. These, however, are intrusive, often prohibitively expensive, and difficult to deploy in the natural environment of the animals. In recent years, we have seen a significant surge in deep learning-based techniques that enable the 3D reconstruction, in a non-intrusive manner, of the shape and motion of dynamic objects just from their RGB image and/or video observations. Several papers have explored their application and extension to various types of animals. This paper surveys the latest developments in this emerging and growing field of research. It categorizes and discusses the state-of-the-art methods based on their input modalities, the way the 3D geometry and motion of animals are represented, the type of reconstruction techniques they use, and the training mechanisms they adopt. It also analyzes the performance of some key methods, discusses their strengths and limitations, and identifies current challenges and directions for future research. 
\end{abstract}

}

\maketitle

\IEEEdisplaynotcompsoctitleabstractindextext

\IEEEpeerreviewmaketitle

\section{Introduction}

\IEEEPARstart{3}{D} Reconstruction of the shape and motion of animals is a fundamental problem in machine learning, computer vision, and graphics. It has a wide range of applications, from digital content creation for the entertainment industry and virtual/augmented reality applications to wildlife conservation, livestock management, animal welfare monitoring, and biological studies. For instance, detailed 3D models of animals can help biologists and conservationists monitor animal health, behavior, and population dynamics, particularly for endangered species~\cite{sinha2023common, Kanazawa2015Learning3D}. Analyzing and understanding how the 3D shape of animals evolves and changes in response to various factors can be correlated to their health and can be used to study the way they respond to environmental changes and guide intervention strategies. Realistic 3D models of animals can also be used to populate virtual environments and enhance the visual experience in virtual and augmented reality~\cite{yang2023Reconstructing, cheng2023virtual}. 3D reconstruction can also empower robots in technology-enabled farms~\cite{yang2023Reconstructing, li2024learning} to better interact with animals and navigate the shared environments.

Traditionally, the 3D reconstruction of animals is based on either active techniques, such as laser scanners and photometric stereo, which are intrusive, or passive techniques such as multi-view stereo. These technologies are difficult to deploy in the environments where animals naturally live. They also require the subjects to remain static during the acquisition process. However, unlike humans, animals are non-cooperative. They continuously move and deform in a non-rigid way, making accurate 3D reconstruction using these traditional techniques very challenging. 

\begin{figure}[t]
  \centering
    \includegraphics[trim={0cm 0cm 0cm 0},clip, width=\linewidth]{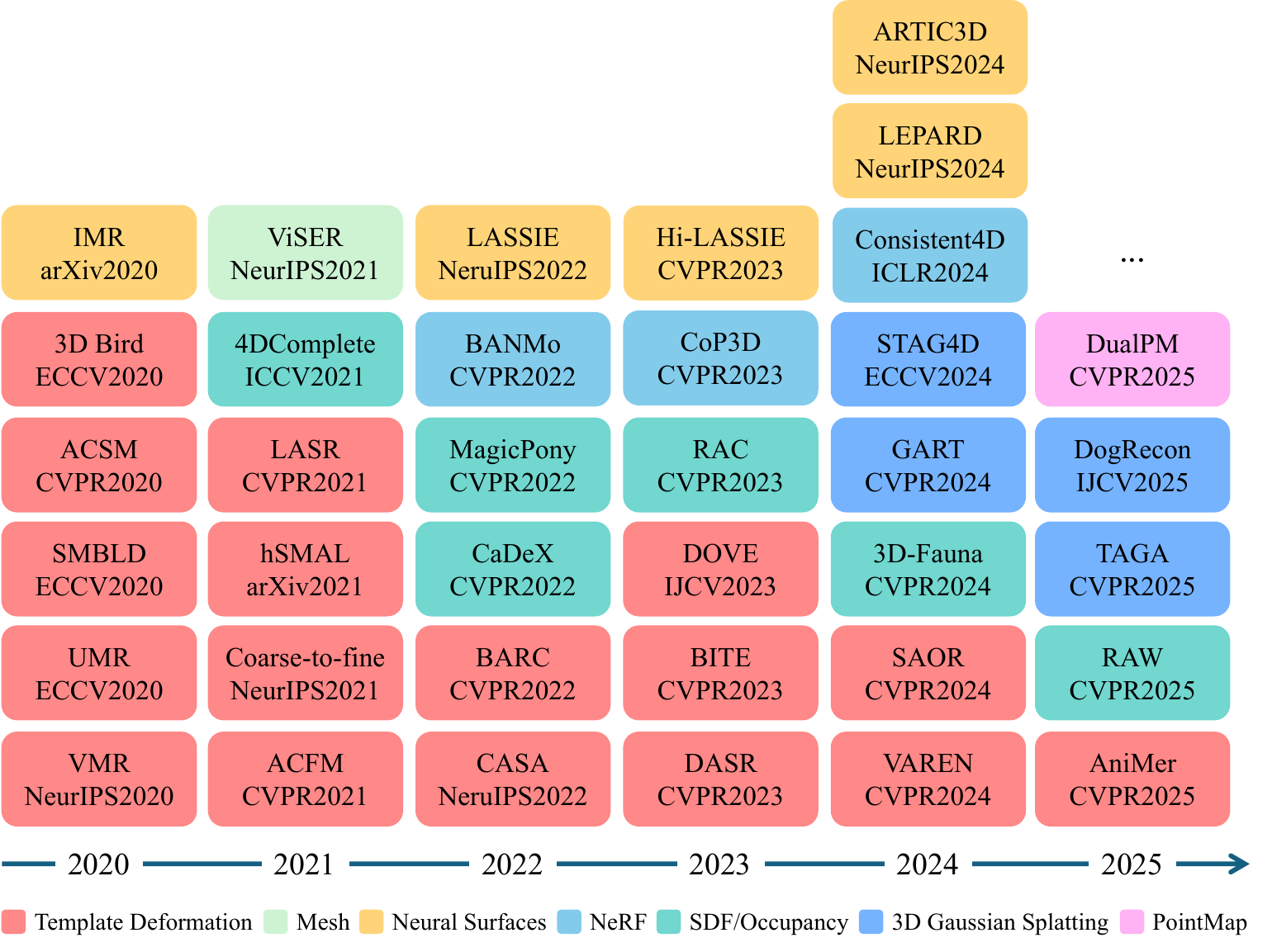}
  \caption{Representative recent papers that tackled the problem of 3D reconstruction of the shape and motion of animals from RGB images and videos. }
  \label{fig:papers}
\end{figure}

\begin{figure*}[t]
  \centering
    \includegraphics[trim={0 10cm 3.4cm 0},clip, width=\linewidth]{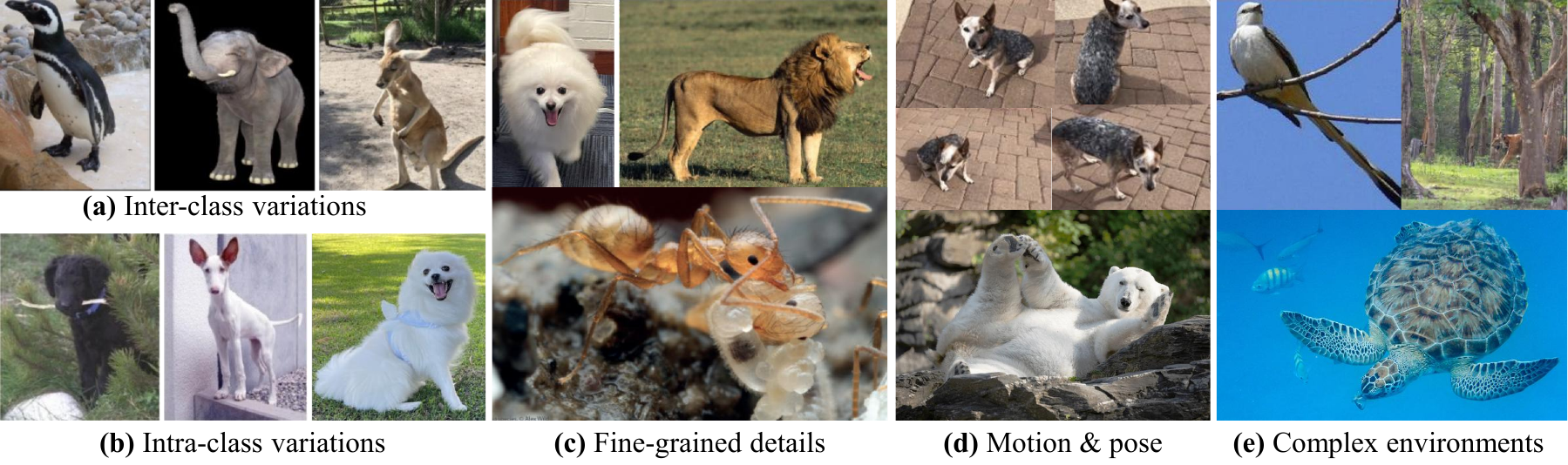}
  \caption{Illustration of the different challenges in reconstructing the 3D shape, pose, and motion of animals in-the-wild.}
  \label{fig:challenges}
\end{figure*}


In recent years, we have seen a surge in deep learning techniques that enable non-intrusive 3D reconstruction of the geometry and motion of various objects from RGB images and videos. Several papers have explored the usage and extension of these methods to the 3D reconstruction of animals. \Cref{fig:papers} shows that the variety of techniques being explored has grown steadily in the past five years, highlighting the importance and the continuous growing interest in the topic in the machine learning, computer vision, and graphics communities~\cite{li2024learning, yang2022banmo, yang2021viser}. This paper reviews and summarizes the latest progress in this continuously growing field of research. It aims to \textbf{(1)} provide researchers and practitioners with the background knowledge necessary to understand this rapidly evolving field, \textbf{(2)} describe the core (new) techniques that (re-)shaped it, and \textbf{(3)} discuss the recent progress and trends in the field. We classify the state-of-the-art  based on their input modalities, the way the 3D shape and motion of animals are represented, the reconstruction techniques and neural architectures they use, and the training strategies they adopt. We also analyze and compare the performance of key methods on existing benchmark datasets. 

Several papers surveyed the 3D reconstruction problem~\cite{han2019image, laga2020survey, Fahim2021SingleView3R, Jin20203DRU, Dalal2024GaussianS3, Tretschk2022StateOT,yunus2024recent}; however, only a few of them discussed the issues related to 3D animal reconstruction. For example, Tretschk \etal~\cite{Tretschk2022StateOT} discussed the differences between reconstructing animals and  other objects, and the special considerations one needs to take into account when dealing with animals. The paper, however, only reviewed parametric model-based methods and does not provide a comprehensive review of latest techniques such as neural implicit representations and Gaussian splatting-based methods. Yunus \etal~\cite{yunus2024recent} focus on general nonrigid 3D shapes. Animals can be seen as a special case of nonrigid 3D shapes. However,  we believe that a dedicated survey is necessary to cover all the challenges inherent to animals. To the best of our knowledge, this is the first comprehensive survey that covers a wide range of 3D animal shape and motion reconstruction methods, beyond parametric models. 

\subsection{Challenges}
\label{sec:challenges}


Reconstructing, from RGB images and/or videos, the 3D geometry, pose, and motion of animals is an ill-posed problem that shares the same challenges as the general 3D reconstruction problem but also raises new challenges that are specific to animals.~\Cref{fig:challenges} summarizes these challenges. 

Unlike human bodies, which have a fixed skeletal structure, animals show a \textbf{high level of inter-class and intra-class geometric and structural variability}. As shown in~\Cref{fig:challenges}-(a), different species exhibit significant differences in both their overall body structure and in the shape of their specific body parts. This is referred to as inter-class variability. Also, animals within the same species exhibit substantial differences,  \eg due to breed variations as shown in the dog images of~\Cref{fig:challenges}-(a). Species-specific methods address this challenge by training one parametric model per species.  Ideally, however, one would require species-agnostic reconstruction techniques that accurately model the rich intra-species and inter-species variability in the 3D shape and structure of animals.

Animals typically have \textbf{complex 3D geometry} with \textbf{fine-grained details} of high biological importance, \eg fur in dogs and lions, and tiny details in insects such as ants (see~\Cref{fig:challenges}-(c)). Such fine details are also crucial for achieving high levels of realism in animal simulation. They are, however, very difficult to reconstruct from just RGB images and videos, especially when the animals are moving and interacting with other objects.


Animals are \textbf{highly dynamic}, \textbf{unpredictable}, and \textbf{non-cooperative}. They undergo complex \textbf{non-rigid motion and deformation} and exhibit natural camouflage (see~\Cref{fig:challenges}-(d)). This not only makes their detection and tracking challenging but also results in high levels of \textbf{(self-)occlusions}. When dealing with humans, these problems are often addressed by using shape and motion priors, \eg statistical shape models such as 3D morphable models~\cite{blanz1999morphable,egger20203dmorphable} or SMPL~\cite{loper2015smpl}. Such priors, however, need to be learned from accurate ground-truth 3D models captured by 3D scanning devices. Similarly, deep learning models need to be trained with pairs of images/videos and their corresponding ground-truth 3D models. While this can be done for static objects or dynamic objects that are cooperative, such as humans, animals are \textbf{not cooperative}. They cannot be brought to digitization studios in numbers and asked to remain still for an extended time to enable their accurate 3D scanning. This results in a \textbf{scarcity of groundtruth 3D data} that can be used to train deep learning models or learn statistical shape priors for animals. 

Finally, compared to other objects, animals live in \textbf{complex environments} and thus one has no control over the lighting conditions and background complexity, including occlusions by other animals that are in the vicinity; see~\Cref{fig:challenges}-(e). Some environments, \eg underwater or aerial settings, even introduce distortions when captured using standard optics~\cite{li2024learning, sinha2023common}.


\subsection{Overview and taxonomy} 
\label{sec:taxonomy}
In this paper, we formulate the problem of reconstructing the 3D shape and motion of animals as  that of learning a function $\neuralsurfacefunc_{\Params}$ of the form:
\begin{equation}
    \begin{split}
        \neuralsurfacefunc_{\Params}:&  \inputdomain \times \rplus \to \outputdomain, \\
        & (\domainpoint, \thetime)  \to \point = \neuralsurfacefunc_{\Params}(\domainpoint,\thetime; \conditions).
    \end{split}
    \label{eq:3Drec_formulation}
\end{equation}

\noi Here, $\domainpoint$ is a point in the input spatial domain  $\inputdomain$, and $\thetime \in \rplus$ is time. The function $\neuralsurfacefunc$, which has parameters $\Params$, outputs the shape properties, such as geometry and appearance, at $(\domainpoint, \thetime) $. The output domain $\outputdomain$ can be a combination of a geometry space $\geometryspace$, an appearance space $\appearancespace$, and the space of any other attributes that one aims to recover, \eg pose.  The function $\neuralsurfacefunc$  can be further conditioned on some observations $\conditions$, which can be one or multiple RGB  images or videos captured with a single or multi-view cameras. They can also be depth maps or partial point clouds captured with depth sensors. These conditions can be provided as input to the function $\neuralsurfacefunc$ or enforced during training. 

\begin{figure}[!t]
  \centering
    \includegraphics[trim={0 13.3cm 12cm 0},clip, width=\linewidth]{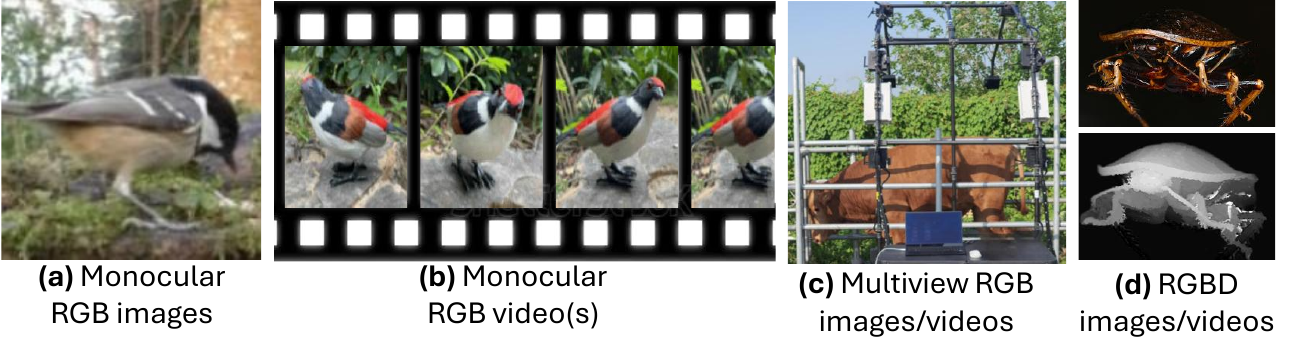}
   
  \caption{Input modalities used for 3D and 4D reconstruction of animals.}
  \label{fig:input_modalities}
\end{figure}


This paper reviews the state-of-the-art 3D animal reconstruction methods and classifies them based on \textbf{(1)} the nature of the observations $\conditions$, hereinafter referred to as input modalities (Section~\ref{sec: Input Modalities}), and \textbf{(2)} the way the 3D shape and motion of animals are represented (Sections~\ref{sec:explicit_representations},~\ref{sec:parametric_models},~\ref{sec:implicit_representations},~\ref{sec:gaussian_avatars}). The latter is the key to any 3D reconstruction method's success. In fact, what is needed is a compact representation that is suitable for processing and regression using neural networks, can capture complex geometries and topologies at different levels of detail, and can be efficiently learned from data. We will first review explicit representations (Section~\ref{sec:explicit_representations}), then  template and model-based methods (Section~\ref{sec:parametric_models}), implicit representations (Section~\ref{sec:implicit_representations}), and finally Gaussian Splatting-based methods (Section~\ref{sec:gaussian_avatars}). Section~\ref{sec:generation} reviews some latest methods that focus on generating animal shape and motion either from images of an environment or text prompts. Section~\ref{Sec: Training Strategies} reviews commonly used training strategies and summarizes existing datasets used  for training and performance evaluation. Section~\ref{sec:discussion} discusses the performance of some key representative methods, while Section~\ref{Sec:Research gaps} provides a summary of the remaining key challenges and potential directions for future research. We conclude in Section~\ref{sec:conclusion}.

    


\section{Input} 
\label{sec: Input Modalities}

  

\subsection{Input modalities}

As summarized in~\Cref{fig:input_modalities}, the input to 3D animal body shape and motion reconstruction can be an RGB image, a depth map $\depthmap$, or a combination of both, hereinafter denoted by RGB-D. It can also be an RGB(-D) video. Depth maps   can be either captured with depth sensors or reconstructed from RGB images using deep learning-based methods such as Depth Anything, a foundational monocular depth estimator introduced by Yang \etal~\cite{yang2024depth}. In all cases, the input can be captured by a single camera in a fixed location or by using multiple cameras sparsely or densely distributed around the scene. The former case is known as \textbf{monocular 3D reconstruction} while the latter is referred to as \textbf{multi-view 3D reconstruction}.  Multi-view RGB and RGB-D images provide a better coverage of the overall scene and are less prone to self-occlusions. The images, however, need to be captured with calibrated and synchronized cameras, which can be expensive and difficult to achieve in the natural environment of animals. 

Due to their practicality, a large proportion of the recent literature focused on monocular 3D reconstruction from RGB(-D) images. This, however, is a highly ill-posed and challenging problem, especially when dealing with highly complex and nonrigid objects such as animals, which exhibit high levels of self-occlusions. During training, many existing methods require additional supervisory signals, which can be obtained through various techniques; see Section~\ref{Sec: Training Strategies}. At runtime, these methods can perform 3D reconstruction from monocular RGB images or videos. 


Despite extensive research, the performance of monocular RGB image-based 3D reconstruction lags significantly behind multi-view-based 3D reconstruction. However, deploying multiple sensors in the wild is not practical.  Some methods address this issue by using \textbf{in-the-wild images of a specific animal species}, instead of multi-view images of each individual animal, to learn common structures across animals within that species~\cite{yao2021lassie, yao2023artic3d, yao2023hi-lassie}. The key insight is that capturing multiple RGB images of the same animal is challenging; finding, from online repositories, various images of different animals from the same species in diverse poses, backgrounds, lighting conditions, and texture variations is relatively easy. Since, in general, animals within the same species share some common characteristics, one can exploit these shared characteristics to train neural networks to learn common structures and infer them at runtime from a monocular RGB image.

Other methods leverage the latest developments in \textbf{generative AI to synthesize, from a single input image, multi-view images} of the same scene and use them as input to a multi-view 3D reconstruction pipeline to recover the full 3D geometry of animals. For instance, Liu \etal~\cite{liu2024one} use the pre-trained 2D diffusion model of~\cite{liu2023zero1to3} to generate multi-view images from a single input image. Liu \etal~\cite{liu2024one2345}  extended the method by using a single-view image to refine the pre-trained 2D diffusion model so that it can generate consistent multi-view images, thereby generating high-fidelity textured 3D models of animals. While promising, these methods can only synthesize novel views from angles that are slightly different from the input views. When the target angle significantly deviates from the angle of the input view, these methods generate plausible images only when the animal is in a neutral pose and thus preserves its symmetry. These significantly limit the performance of the 3D reconstruction process.

3D reconstruction can also be performed from monocular or multi-view \textbf{RGB(-D) video streams}~\cite{li2021hsmal,yang2022banmo,sinha2023common}, instead of static images. The advantage of using videos is two-fold; \First, they can be used to recover the motion of the animal, in addition to its 3D shape. \Second, regions and animal parts that are occluded in one frame can become visible in other frames. Thus, with efficient fusion methods, one can address the (self-)occlusion issues faced in image-based 3D reconstruction. This, however, comes with additional challenges. For instance, one needs to ensure temporally consistent reconstruction across frames. Also, animal body shapes deform in a non-rigid way. Thus, to exploit information across frames, one needs to non-rigidly align and fuse the observations across frames.

\subsection{Input encoding}

\begin{figure}[t]
    \centering
    \includegraphics[trim={0 3.8cm 0cm 0},clip,width=\linewidth]{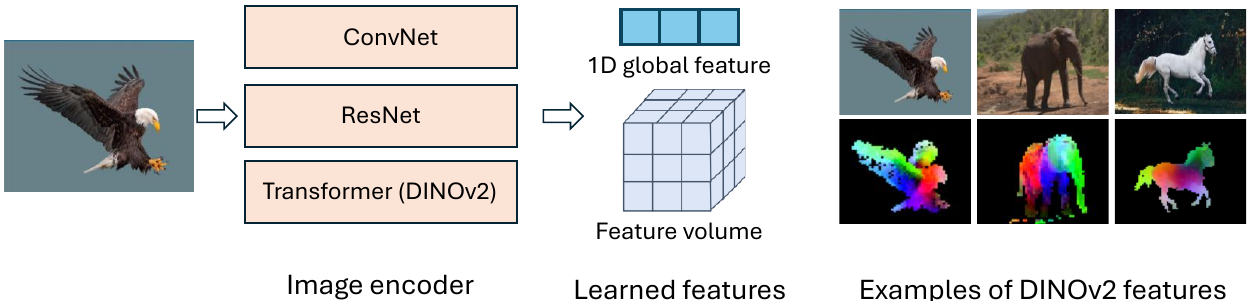}
    \caption{Input encoding using various types of encoders. Here, we show examples of features learned using DINOv2~\cite{caron2021dino,oquab2023dinov2} - we show the first three principal components highlighted in distinct colors.}
    \label{fig:dino}
\end{figure}

In general, the input modalities are not fed directly into the reconstruction module but encoded into features, which are then processed by the reconstruction, or decoding, module. The feature space is referred to as the latent space. 
A good encoding should map the modalities that represent the same 3D objects to nearby points in the latent space, be robust to small changes in the input, and remain invariant to extrinsic factors such as camera pose.  Traditionally~\cite{han2019image}, the encoding stage is implemented using fully convolutional networks~\cite{li2021complete,yang2022banmo,wu2023dove,liu2024one,sinha2023common} or residual networks such as ResNet18~\cite{Kanazawa2018CategorySpecific,kulkarni2020articulation,tulsiani2020implicit,li2020self,wu2022casa} or the deeper ResNet34~\cite{ruegg2022barc}  or ResNet50~\cite{Zuffi2019Safari,biggs2020who}. 

In recent years, there has been increasing interest in the use of pre-trained Vision Transformers (ViT) such as self-distillation with NO labels (DINO)~\cite{caron2021dino} and DINOv2~\cite{oquab2023dinov2}, as encoders. DINO is a self-supervised learning approach that enables ViTs to capture semantic parts in images without needing labeled data. As shown in~\Cref{fig:dino}, DINO can extract meaningful features that represent various parts of animals, which is crucial for reconstructing 3D skeletons and shapes from images. More importantly, DINO features are very stable under drastic changes in imaging conditions such as lighting and camera pose. This is particularly important for in-the-wild 3D/4D animal reconstruction. For instance, methods such as~\cite{yao2021lassie,yao2023hi-lassie,li2024learning,liu2024one2345,liu2024lepard} used the DINO-ViT encoder to extract image features that are stable across multi-view. Wu \etal~\cite{wu2023magicpony} combined the DINO-ViT with additional convolutional networks to extract detailed image features. Yao \etal~\cite{yao2023hi-lassie} applied DINO-ViT to discover 2D and 3D parts for high-fidelity articulated skeleton reconstruction, while Li \etal~\cite{li2024learning} leveraged DINO features to build deformable 3D models of animals across multiple species.

\section{Explicit representations}
\label{sec:explicit_representations}

\begin{figure*}[t]
  \centering
    \includegraphics[trim={0 11.2cm 0 0},clip, width=\linewidth]{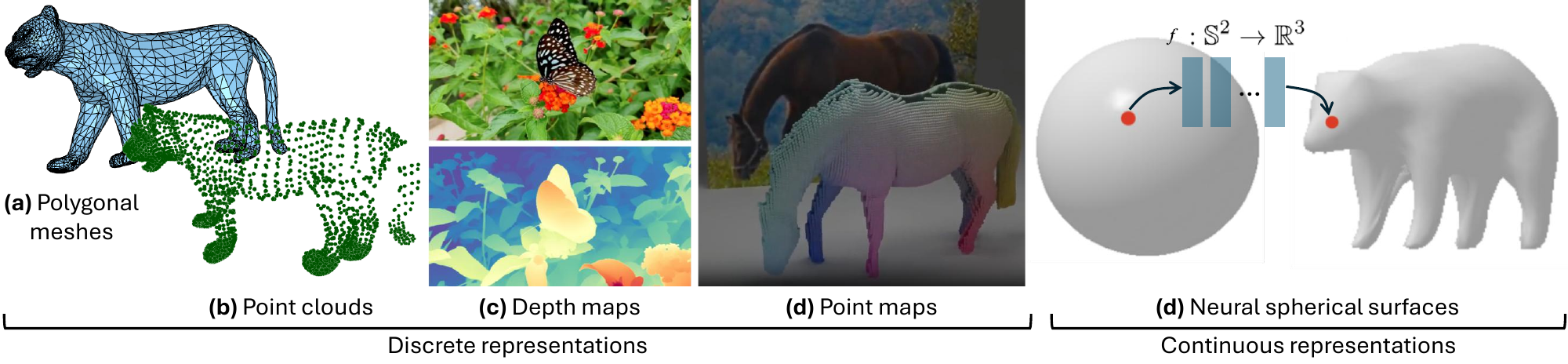}
   
  \caption{Various explicit representations used for deep learning-based 3D reconstruction of animals.}
  \label{fig:explicit_representations}
\end{figure*}

Explicit representations define the 3D surface of an object as a differentiable 2-manifold embedded in the ambient 3D Euclidean space:
\begin{equation}
    \begin{split}
        \neuralsurfacefunc:&  \inputdomain  \to \geometryspace = \rthree, \text{ such that }  \domainpoint  \to \point = \neuralsurfacefunc(\domainpoint).
    \end{split}
    \label{eq:explicit}
\end{equation}

\noi The function $\neuralsurfacefunc$ maps a point $\domainpoint \in \inputdomain$ to a 3D point $\point \in \geometryspace=\rthree$.  Polygonal meshes and point clouds (Section~\ref{sec:polygonal}), originally designed for visualization, are the most commonly used explicit representations. However, they are discrete and unstructured. Thus, they are difficult to regress using neural networks. This issue has been addressed using neural surfaces (Section~\ref{sec:neural_surfaces}).


\subsection{Discrete, explicit representations}
\label{sec:polygonal}

\subsubsection{Representations}
A polygonal mesh (\Cref{fig:explicit_representations}-(a))  can be seen as a piecewise linear approximation of the function  $\neuralsurfacefunc$ of Equation~\eqref{eq:explicit}  where the domain $\domain$ is defined as the union of $\nfaces$ triangular faces $\{\face_i\}_{i=1}^{\nfaces}$. Each triangular face $\face_i$ is defined by its three vertices $(\vertex_1^i, \vertex_2^i, \vertex_3^i)$ with $\vertex_j^i \in \rthree$. Any point $\point$ on the surface of the object can be explicitly defined using its barycentric coordinates with respect to the vertices of the triangular face to which it belongs. Thus, the surface is fully represented with its vertices and faces.

Triangular meshes are the most commonly used representation in computer vision and graphics due to their efficiency and compatibility with graphics engines. However, they cannot capture the internal details of a 3D shape. Polygonal meshes, however, are a particular case of simplicial complex. An $n$-simplex is a geometric object with $\nvert+1$ vertices, which lives in an $n$-dimensional space (and cannot fit in any space of smaller dimension): a $0$-simplex defines a point, a $1$-simplex defines a segment, a $2$-simplex defines a triangle, while a $3$-simplex defines a tetrahedron. Thus, a 3D volume can be represented using a $3$-simplicial complex, which is the collection of $3$-simplices (or tetrahedra). 


A \textbf{point cloud} (\Cref{fig:explicit_representations}-(b)) is a $0$-simplicial complex that can be represented using a set of  3D locations $\{\point_i \in \rthree\}_{i=1}^\npoints$. Unlike triangular meshes, point clouds are light in memory requirements and can represent the fine details of complex objects. Their main issue is that a 3D point does not have the notion of surface. In other words, a 3D point in isolation does not capture the local surface properties. As such, computing loss functions such as the Chamfer Distance (CD) during the optimization as well as the rendering of the reconstructed point clouds requires surface extraction, which is not differentiable.  This issue has been addressed by abstracting the local geometry around a point using spheres~\cite{lassner2021pulsar} or surfels~\cite{pfister2000surfels}. A sphere is defined by its center, radius, and opacity. Since it is three-dimensional, it captures the 3D volume. A \textbf{surfel}, on the other hand,  is a disk tangent to the surface at a given point. It has a center, a radius, a unit normal direction, and an appearance~\cite{pfister2000surfels,mihajlovic2021deepsurfels}. Thus, it characterizes the local neighborhood of a surface at the given point with an oriented disk.

\begin{wrapfigure}{r}{0.6\linewidth}   
    \includegraphics[trim={9.5cm 0.4cm 9.5cm 0},clip, width=\linewidth]{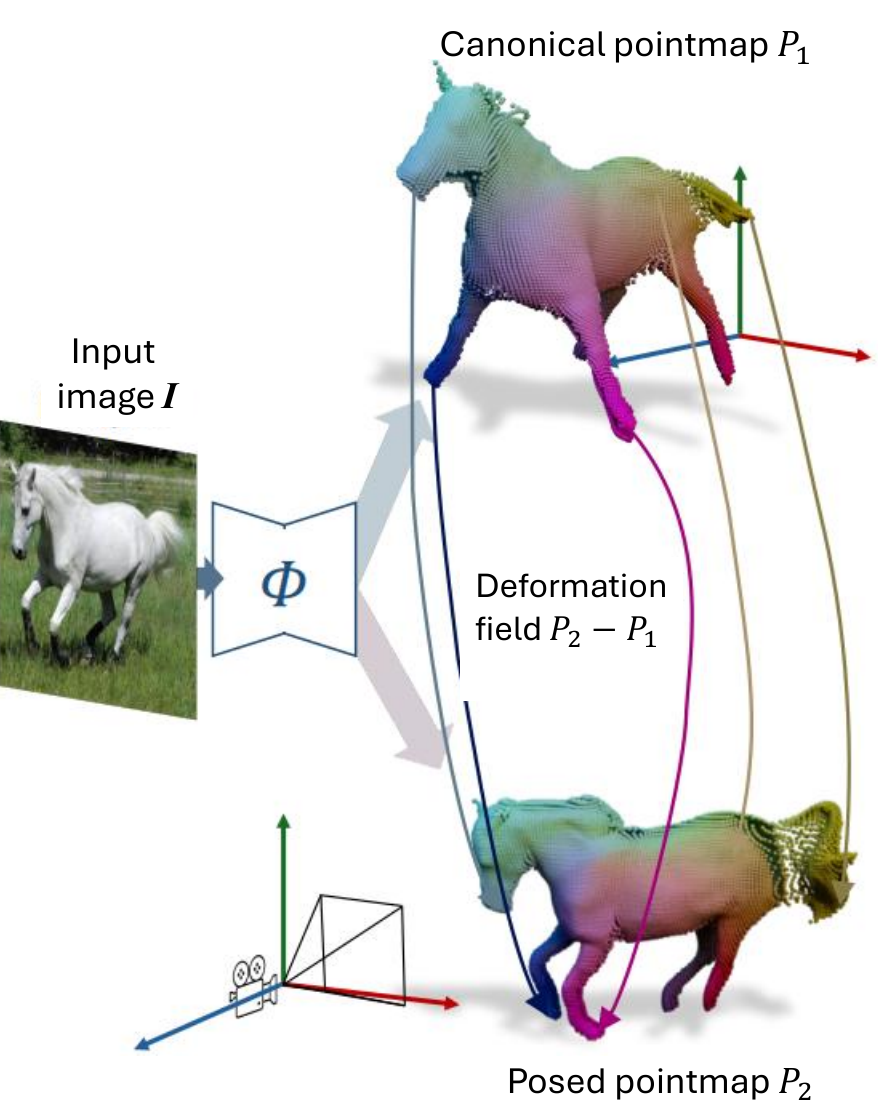}
    \caption{Animal reconstruction using point map-based representations~\cite{kaye2025dualpm}. Image adapted from~\cite{kaye2025dualpm}.}
    \label{fig:PointmapReconstruction}
\end{wrapfigure}

Polygonal meshes and point clouds are irregular data structures, and thus, they cannot be directly processed with convolutional operations. On the other hand, \textbf{depth maps} (\Cref{fig:explicit_representations}-(c)), also referred to as $2.5$D representation,  have the same regular structure as an RGB image but only store partial information, and rely on the availability of the camera's intrinsics, which are very difficult to estimate especially when dealing with animals in the wild. To address this issue, DUSt3R~\cite{wang2024dust3r} introduced the concept of \textbf{pointmap} (\Cref{fig:explicit_representations}-(d)), which is a dense 2D field of 3D points $\pointmap \in \real^{\width\times\height\times 3}$. In association with its corresponding RGB image of resolution $\width\times\height$, it defines a one-to-one mapping between image pixels and 3D scene points. Similar to depth maps, a pointmap is a partial representation.  Thus, one requires a collection of pointmaps to represent a complete 3D object or scene. However, unlike depth maps, pointmaps are independent of the intrinsic parameters of the cameras. Thus, the idea behind it is to store the 3D object or scene in multiple canonical views so that one can work in image space. Introduced by Wang \etal~\cite{wang2024dust3r} for geometric 3D vision tasks, it has been used for image matching~\cite{leroy2024grounding}, dynamic 3D reconstruction from monocular videos~\cite{zhang2025monst3r}, and 3D shape and pose reconstruction~\cite{kaye2025dualpm}.

\subsubsection{Reconstruction}

To the best of our knowledge, few papers have utilized polygonal meshes or unstructured point clouds for the 3D reconstruction of the shape, pose, and motion of animals. Often, in these methods, a dynamic 3D object is represented as the deformation of a canonical configuration. An example is DynamicFusion~\cite{newcombe2015dynamicfusion}, which extends KinectFusion by reconstructing dynamic 3D scenes in the form of a point cloud in a canonical pose and a dense volumetric 6D motion field that warps the estimated canonical geometry to each video frame. Although DynamicFusion-based methods can achieve plausible results in reconstructing dynamic objects in general, they require point clouds as input and are, in general, slow as they rely on non-linear optimizations.

In recent years, we have seen a surge of interest in the use of pointmaps for dynamic scene reconstruction. Kaye \etal~\cite{kaye2025dualpm} introduced DualPM, dual posed-canonical pointmaps for 3D shape and pose reconstruction of dynamic objects such as animals. Unlike the original pointmaps~\cite{wang2024dust3r}, which reconstructs static 3D objects, Kaye \etal~\cite{kaye2025dualpm} handle dynamic 3D objects by predicting two pointmaps (see~\Cref{fig:PointmapReconstruction}): the first one, denoted by $\pointmap_1$, is similar to~\cite{wang2024dust3r} as it assigns for each pixel in the input image, its corresponding 3D point. The second pointmap, referred to as the canonical pointmap and denoted by $\pointmap_2$, is defined in a canonical space where the object is in a canonical (neutral) pose. Thus, it is invariant to the object's pose and deformation. The two maps implicitly encode the object's deformation. 

To predict the dual pointmaps, Kaye \etal~\cite{kaye2025dualpm} took an input image and coded it using a pre-trained network~\cite{zhang2023tale}
from which the canonical point map $\pointmap_2$ is predicted. The method then conditions the prediction of $\pointmap_1$ on $\pointmap_2$. The method also predicts per-pixel confidence scores for each of the two maps. 

The main limitation of point maps is that they only predict the visible portion of the object. To predict complete 3D objects,  Kaye \etal~\cite{kaye2025dualpm} introduced the concept of an \textbf{amodal pointmap}, which associates each pixel of the image, within the mask of the object of interest, to the sequence of 3D points that intersect the camera ray through that pixel in order of increasing distance from the camera center. To predict the amodal point map, Kaye \etal~\cite{kaye2025dualpm} used an image-based layered representation. In this representation, the first layer encodes the visible points, and each consecutive layer encodes the next set of points occluded by the previous layer. Note that the method is trained and supervised with synthetic data. 
  
\subsection{Neural surfaces}
\label{sec:neural_surfaces}

Although discrete representations such as polygonal meshes and point clouds are the gold standard when it comes to 3D scanning and rendering, they raise several challenges when processing them using deep neural networks. \textbf{First}, they are not organized into regular grids. Thus, unlike images, they are not suitable for processing using convolutional operations. Pointmaps address this issue but, similar to depth maps, they represent partial information. \textbf{Second}, representing fine-grained details, using discrete representations, would require high-resolution meshes and point clouds, which will induce a shape space of very high dimension. \textbf{Finally},  the same 3D object or scene can be represented using different, but equivalent,  polygonal meshes or point clouds of different resolutions and topologies. 

In this section, we review continuous explicit representations, which aim to address some of the issues faced by discrete representations.


\subsubsection{Representation}
Instead of using a discrete surface representation, other methods treat $\inputdomain$ in Equations~\eqref{eq:3Drec_formulation} and~\eqref{eq:explicit} as a continuous domain and then parameterize the function $\neuralsurfacefunc$, which maps points in $\inputdomain \subset \rthree$ to shape properties, using neural networks, \eg MLPs, exploiting their power as universal approximations; see~\Cref{fig:explicit_representations}-(d).  Termed \emph{neural surfaces}~\cite{zhang2021ners}, this representation has many benefits: it is continuous and can represent surfaces in a resolution-free or discretization-agnostic manner. Thus, in theory, it can be discretized at any arbitrary resolution. Also, the neural networks that represent such surfaces are data-efficient at training. Unlike the discrete methods discussed in Section~\ref{sec:polygonal}, which require full 3D models as ground-truth, neural surfaces only require sample points from the parameterization domain paired with their corresponding ground-truth shape properties. Thus, they are data-efficient since a single 3D model can result in a large number of training samples.

\begin{figure*}[t]
    \centering
    \includegraphics[trim={0cm 5.1cm 0cm 0},clip, width=\linewidth]{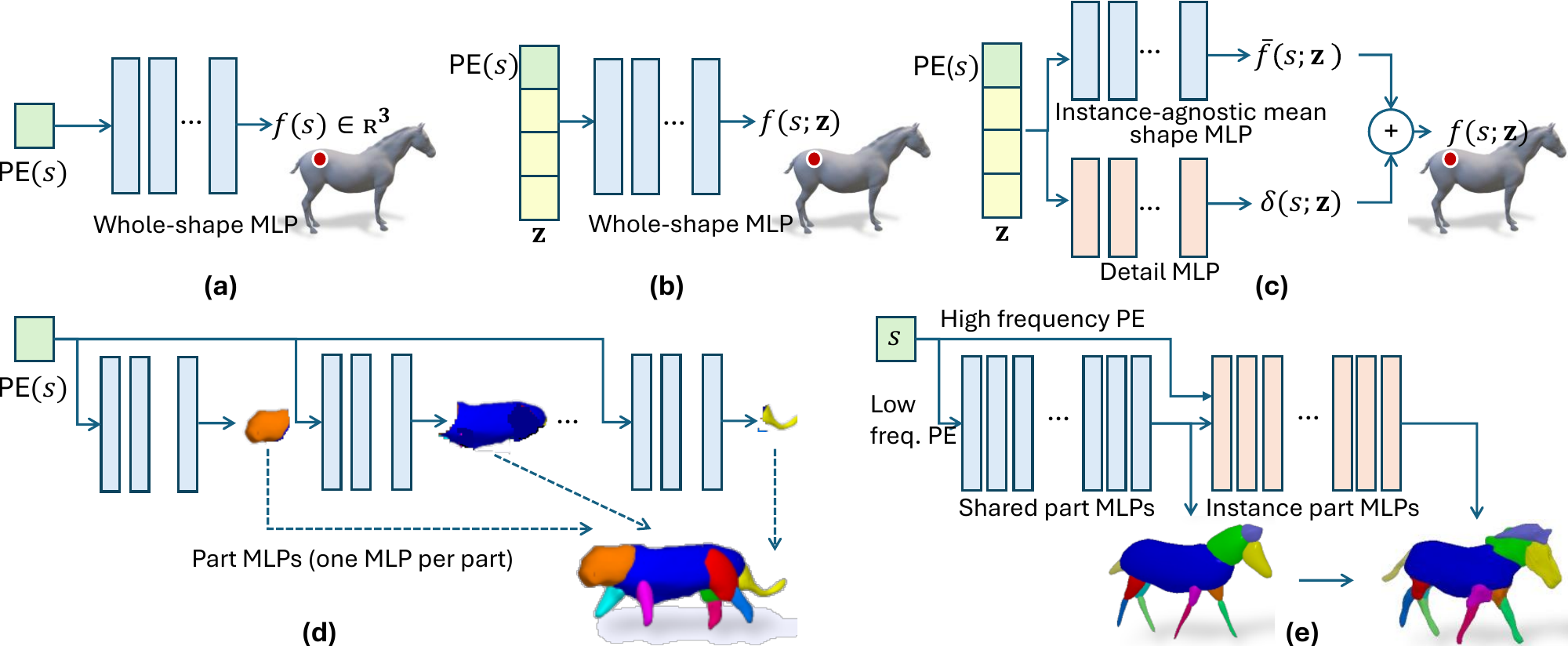}
    \caption{Neural architectures for neural surface-based 3D animal representation and reconstruction. \textbf{(a)} Object-specific neural representation~\cite{tulsiani2020implicit,morreale2021neural,zhang2021ners,williamson2024neural}. \textbf{(b)} Object-agnostic neural representation~\cite{tulsiani2020implicit}.  \textbf{(c)} Combining an instance-agnostic mean shape $\mean{\neuralsurfacefunc}$ and an instance-dependent deformation that capture instance-specific surface details~\cite{tulsiani2020implicit}.  \textbf{(d)} Partwise neural surfaces~\cite{yao2021lassie,yao2023artic3d}. \textbf{(e)} Neural surfaces decomposed into shared (low-frequency) and instance-specific (high-frequency) components~\cite{yao2023hi-lassie}. PE refers to point (or frequency) encoding.}
    \label{fig:neuralsurfaces}
\end{figure*}

Existing 3D and 4D methods that build on this general formulation differ in the nature of the domain $\domain$. Methods such as~\cite{tulsiani2020implicit,morreale2021neural,zhang2021ners,williamson2024neural}  define $ \domain$ as the unit sphere $\stwo$; see~\Cref{fig:neuralsurfaces}-(a). In other words:
\begin{equation}
        \neuralsurfacefunc: \stwo  \to \rthree, \text{ such that } \domainpoint  \to \point = \neuralsurfacefunc(\domainpoint).
    \label{eq:spherical_explicit}
\end{equation}

\noi This representation, however, is object-specific. To model different possible shapes across instances of a category, Tulsiani \etal~\cite{tulsiani2020implicit} make the shape function additionally dependent on a latent code $\latentcode\in \real^{\featuredim}$; see~\Cref{fig:neuralsurfaces}-(b):  
\begin{equation}
        \neuralsurfacefunc: \stwo \times \real^{\featuredim} \to \rthree, \text{ such that } (\domainpoint, \latentcode)  \to \point = \neuralsurfacefunc(\domainpoint, \latentcode).
    \label{eq:general_spherical_explicit}
\end{equation}

\noi To benefit from commonalities within animal species, Tulsiani \etal~\cite{tulsiani2020implicit} further decomposed $\neuralsurfacefunc$ into a combination of an instance-agnostic mean shape $\mean{\neuralsurfacefunc}$ and an instance-dependent deformation (\Cref{fig:neuralsurfaces}-(c)):
\begin{equation}
     \neuralsurfacefunc(\domainpoint, \latentcode) = \mean{\neuralsurfacefunc}(\domainpoint, \latentcode) + \deformation(\domainpoint, \latentcode) \text{ with } \domainpoint \in \stwo; \latentcode \in \real^{\featuredim}.
 \end{equation}


\noi Instead of representing the entire shape with a single function, other methods~\cite{yao2021lassie,yao2023hi-lassie,yao2023artic3d} decomposed a complex 3D shape into parts, following its skeletal structure, and represent each part using a spherically-parameterized neural surface~\cite{zhang2021ners}. These parts are then combined and blended together using rigid transformations based on the skeletal structure. Specifically, each part is scaled, rotated, and translated according to its corresponding bone in the skeleton, ensuring connectivity between parts under various articulations. In this representation, one can use instance specific MLPs, with one MLP per part as shown in~\Cref{fig:neuralsurfaces}-(d), or instance-agnostic MLPs (one per part) that learn the mean shape, followed by instance specific part MLPs; see~\Cref{fig:neuralsurfaces}-(e).

\subsubsection{Reconstruction}

Existing methods implement the function $ \neuralsurfacefunc$ as an MLP that takes a point $\domainpoint\in \stwo$, or its frequency encoding, and regresses its corresponding surface point $\point\in \rthree$
~\cite{tulsiani2020implicit,morreale2021neural,zhang2021ners,williamson2024neural,nizamani2025dynamic}. The main difference among these methods is the nature of the MLP and the way it is trained.

For instance, Tulsiani \etal~\cite{tulsiani2020implicit} take an image pixel and predict its location on $\stwo$ using a pre-trained ResNet-18. The method then uses another four-layer MLP to predict the mapping between the sphere and 3D. Zhang \etal~\cite{zhang2021ners} dropped the convolutional block. Instead, they first transform a point $\domainpoint\in \stwo$  into its Cartesian coordinates. The problem of learning the function $\neuralsurfacefunc$ is then formulated as that of learning, using an MLP, a deformation field  that deforms the sphere to the target shape. This approach, however, fails to model complex shapes. In particular, it struggles to fit highly non-convex regions while ensuring that the surface remains a manifold.

Several, often complementary, strategies have been proposed to handle complex shapes, especially those with highly non-convex regions. For instance, Nizamani \etal~\cite{nizamani2025dynamic}  use multiple residual MLP blocks to estimate step-wise deformations. The idea is to ensure that the fitting is performed in a step-wise manner and that each step ensures that the deformation field is diffeomorphic.  Tulsiani \etal~\cite{tulsiani2020implicit} decompose the shape of an animal into \textbf{(1)} a base shape that is common to all instances of an animal species and \textbf{(2)} instance-specific (high-frequency) shape details. It then uses a two-level neural surface. The first one represents the base shape and is shared by all instances within the species. The instance-specific details are modelled using a separate neural surface that captures the high-frequency details unique to each animal instance. This two-level representation captures detailed and accurate 3D shapes while preserving class-specific priors learned from all images.

LASSIE~\cite{yao2021lassie} and subsequent methods such as  Hi-LASSIE~\cite{yao2023hi-lassie} and ARTIC3D~\cite{yao2023artic3d} decompose a 3D shape of an animal into parts, following its skeleton structure. They then represent each part with a deformable neural surface~\cite{zhang2021ners}. These parts are blended together to form a complete, consistent surface. 

More formally, LASSIE~\cite{yao2021lassie} takes as input $20$ to $30$ in-the-wild images of an animal in varying poses and a generic 3D skeleton. It then uses a first MLP that fits a geometric primitive, represented using a neural surface, per body part. During the optimization, the method also poses the generic skeleton to each input image and centers the estimated body parts to the nodes of the skeleton and scales them to the size of the skeleton bones. A second network, termed deformation MLP, then deforms the individual geometric primitives to exactly match the 3D shape of the body parts. 

Hi-LASSIE~\cite{yao2023hi-lassie} and ARTIC3D~\cite{yao2023artic3d} follow a similar pipeline as LASSIE but with two major differences. First, LASSIE~\cite{yao2021lassie} assumes that a generic 3D skeleton is given. Hi-LASSIE~\cite{yao2023hi-lassie} and ARTIC3D~\cite{yao2023artic3d}, on the other hand, automatically discover the skeleton structure from a collection of animals of the same species but in different poses. To find an initial 3D skeleton, Hi-LASSIE~\cite{yao2023hi-lassie} first extracts a 2D skeleton from a reference image. It then lifts it to 3D by finding symmetric parts and separating them in the 3D space with respect to the symmetry plane. Second,  Hi-LASSIE~\cite{yao2023hi-lassie} initializes the geometry of each part using the neural surfaces of basic primitives and then optimizes their geometry using two networks. The first one is a set of part MLPs (one per part) but shared across all instances of the species. This enables the recovery of the geometry that is shared across the species. The second one is a part MLP but instance specific. Thus, it is used to recover instance-specific geometric details. 

To improve the 3D reconstruction, ARTIC3D~\cite{yao2023artic3d} further enhances the input images with occlusions/truncation via 2D diffusion to obtain cleaner mask estimates and semantic features. Further, it performs diffusion-guided 3D optimization to estimate shape and texture that are of high-fidelity and faithful to the input images.

\section{Template deformation-based methods}
\label{sec:parametric_models}

\begin{figure*}[htp]
\centering
    \begin{tabular}{@{}c@{}}
        { \includegraphics[width=0.3\linewidth]{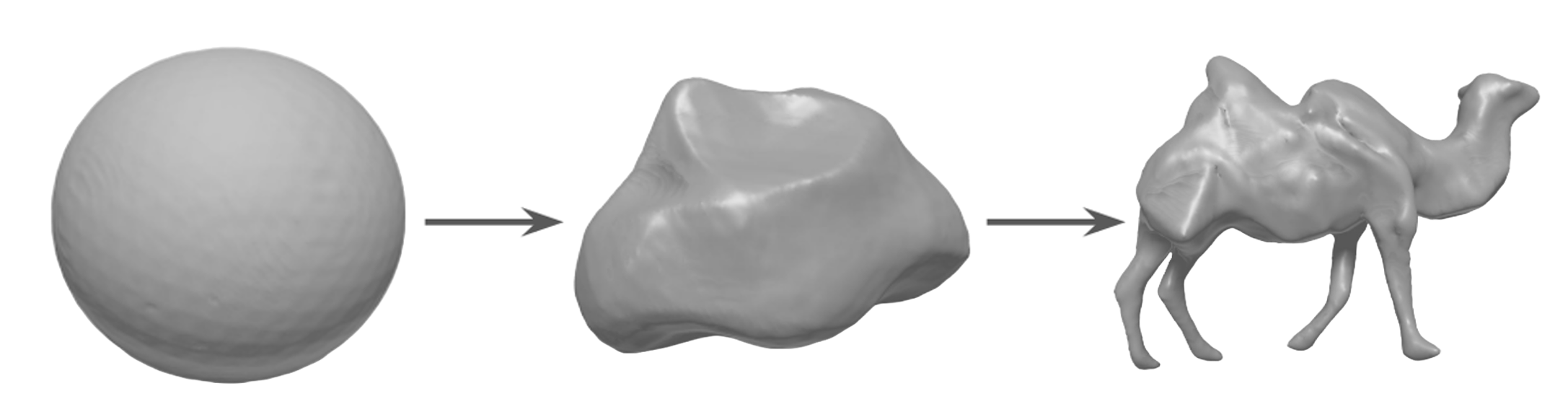}}\\ 
        \textbf{(a)} Free-form template deformation.
    \\
    {\includegraphics[trim={0 9cm 0 0},clip, width=0.3\linewidth]{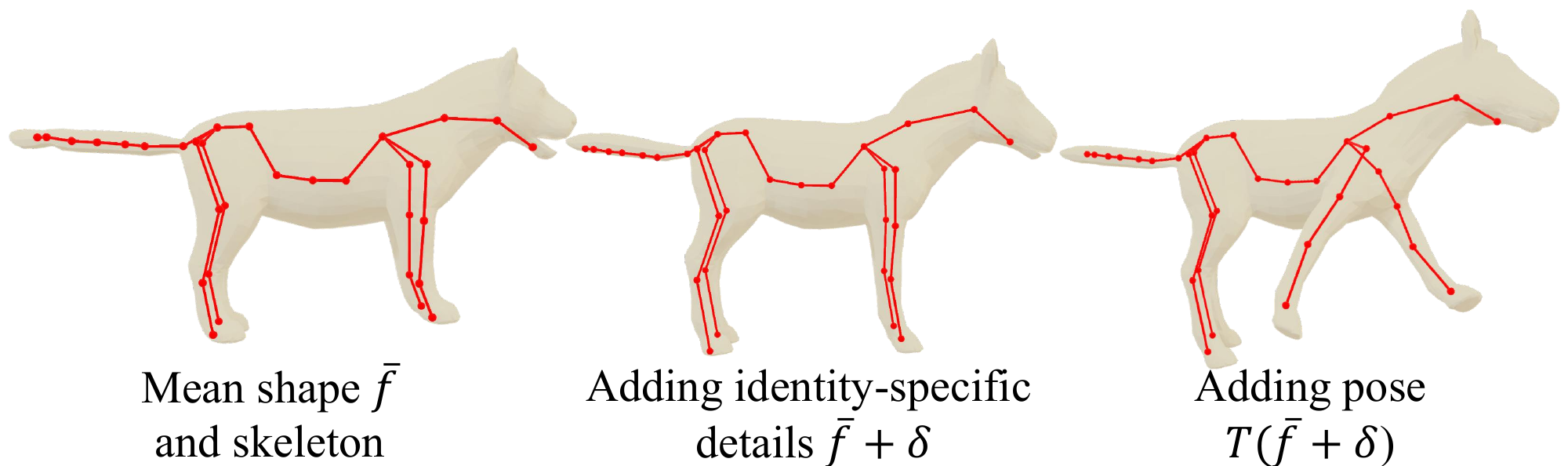}}\\ 
     \textbf{(c)} SMAL~\cite{zuffi2017menagerie} model.
    \end{tabular}
    \begin{tabular}{@{}c@{}}
     {\includegraphics[trim={0 10.5cm 0 0},clip, width=0.68\linewidth]{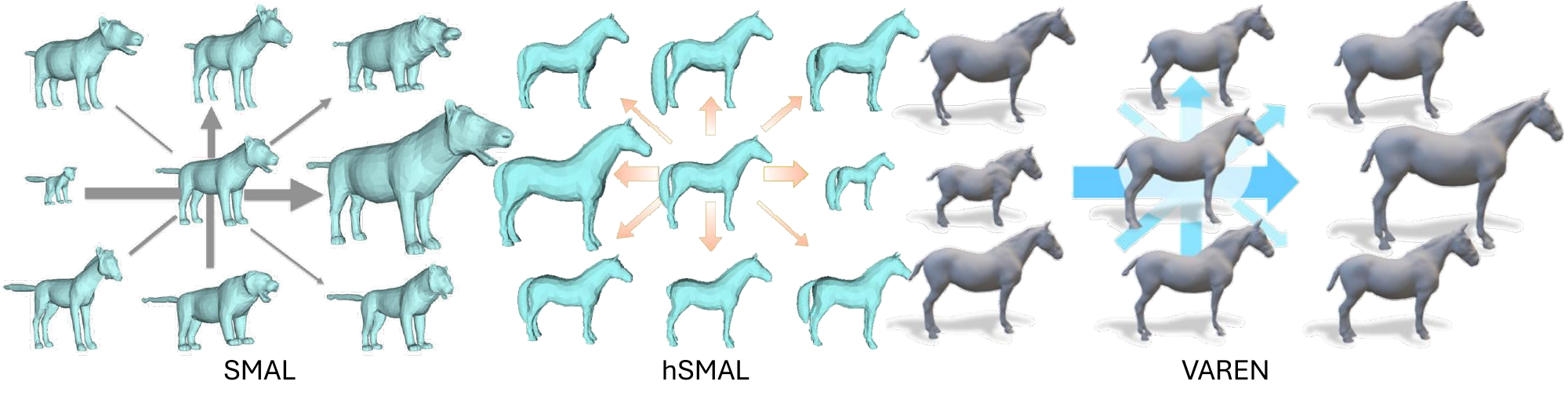}}\\ 
     \textbf{(b)} PCA spaces of SMAL~\cite{zuffi2017menagerie},  hSMAL~\cite{li2021hsmal}, and VAREN~\cite{zuffi2024varen}.
    \end{tabular}
    
\caption{\textbf{(a)} and \textbf{(b)} show deformation-based representation of animals. \textbf{(c)} shows that the PCA spaces of SMAL~\cite{zuffi2017menagerie}, which combines multiple quadruped species, and hSMAL~\cite{li2021hsmal} and VAREN~\cite{zuffi2024varen}, which are specifically designed for horses. For each example, we show the mean shape in the center and the first four principal components. The width of the arrow represents the order of the component. The deviations are shown at $\pm2$ standard deviations from the mean.}
\label{fig:deformation-based}
\end{figure*}

Explicit representations such as polygonal meshes, point clouds, and continuous neural surfaces result in shape spaces of high dimension. Thus, regressing such representations from RGB images or videos would require large neural networks with a large number of parameters, making them very difficult to train. To address this issue, some methods define the 3D shape of an animal as the deformation of a canonical template. The 3D reconstruction problem then reduces to that of estimating the deformation field that aligns the template to the observed images or videos. Another class of methods learns a parametric (or statistical shape) model that represents a low-dimensional embedding of the 3D shape. Parametric methods can be viewed as a special case of template deformation-based methods, where the template is set to be the statistical mean shape learned from exemplars, and the deformation field is constrained by a deformation basis (or principal directions of deformation) learned from the data.

These approaches have shown remarkable performance when reconstructing human body and body part shapes. Thus, their extension to animals has been actively pursued by the community. In what follows, we will provide, in~\Cref{sec:parametric_representation}, a general formulation that unifies these two representations. We will then discuss how these representations are learned from 3D and 4D data (\Cref{sec:learning_parameteric_models}). We will review in~\Cref{sec:reconstruction_pmodels} the methods that use these representations for the 3D reconstruction of the shape, pose, and motion of animals. Finally,~\Cref{sec:joint_statistical_reconstruction} focuses on methods that learn parametric models of animals directly from in-the-wild RGB images and videos.

\subsection{Representation}
\label{sec:parametric_representation}

The key idea (see~\Cref{fig:deformation-based}) is to represent any 3D shape  $\neuralsurfacefunc$ as the deformation of a  template $\mean{\neuralsurfacefunc}$, \ie
\begin{equation}
    \neuralsurfacefunc= \transformation(\mean{\neuralsurfacefunc}  + \deformation).
    \label{eq:template_deformation}
\end{equation}

\noi Here, $\transformation$ is a nonlinear deformation while  $\deformation$  is a displacement field applied to each point on the template shape $\mean{\neuralsurfacefunc}$. The former models large deformations such as those due to articulated motion, while the latter models identity-specific shape and fine-grained surface details such as muscle deformations and surface wrinkles.  Existing representations differ in the type of the template $\mean{\neuralsurfacefunc} $ and how the deformations  $\transformation$ and $\deformation$ are defined.

\textbf{The template $\mean{\neuralsurfacefunc}$} can be either \textbf{(1)} a basic geometric primitive such as a sphere or an ellipse~\cite{yang2021lasr,wu2023dove,aygun2024saor}, \textbf{(2)} an exemplar from the animal species of interest~\cite{badger2020bird,kulkarni2020articulation,kokkinos2022learning}, or \textbf{(3)} the average 3D shape of the animal species of interest learned from a set of exemplars~\cite{Kanazawa2015Learning3D,Kanazawa2018CategorySpecific,wu2023magicpony}. Specifically, let $\{\neuralsurfacefunc_i\}_{i=1}^{\nshapes}$ denote a collection of 3D animal shapes. To enforce correspondence across all shapes, each $\neuralsurfacefunc_i$ is first registered to a predefined template shape in a neutral pose. An inverse Linear Blend Skinning (LBS) transform is applied to bring each registered shape $\neuralsurfacefunc_i^{\mathrm{reg}}$ to a canonical rest pose $\neuralsurfacefunc_i^{\mathrm{rest}}$ of the template. The mean shape $\mean{\neuralsurfacefunc}$ can then be computed as the average of the normalized shapes $\neuralsurfacefunc_i^{\mathrm{rest}}$.


\textbf{The detailed deformation $\deformation$} can be defined in two ways. The \textbf{first}  class of methods follows a free-form deformation approach and defines  $\deformation$ as a displacement field, \ie the displacements of the points on $\neuralsurfacefunc$~\cite{kokkinos2022learning}; see~\Cref{fig:deformation-based}-(a). However, not every deformation field leads to a valid 3D animal shape. Thus,  existing methods constrain $\deformation$ to maintain local rigidity and physical consistency during deformation. A commonly used approach is to control the deformation using handles, following Laplace surface editing~\cite{sorkine2004laplacian,kokkinos2022learning}, while imposing constraints such as  As-Rigid-As-Possible (ARAP) deformations~\cite{sorkine2007rigid,Kanazawa2015Learning3D,li2020online,yang2021lasr,wu2023dove}, symmetry~\cite{Kanazawa2015Learning3D,Zuffi2019Safari}, and stiffness. One can also guide the deformation using positional constraints, \eg by enforcing the deformations to match user-specified keypoints~\cite{Kanazawa2015Learning3D}.

Unlike free-form deformations, the \textbf{second} class of methods learns the subspace of plausible deformations; see~\Cref{fig:deformation-based}-(b). In other words, a deformation $\deformation$ is defined as a linear combination of an orthonormal deformation basis $\left(\eigenvector_1, \dots, \eigenvector_d\right)$, \ie:
\begin{equation} 
    \deformation = \sum_{k=1}^{d}\alpha_k \eigenvector_k, \text{ } \alpha_k \in \real.
    \label{eq:pca_deformation}
\end{equation}

\noi By setting $\transformation$ to identity and using the deformation field of Eqn.~\eqref{eq:pca_deformation}, Eqn.~\eqref{eq:template_deformation} is equivalent to Active Shape Models (ASM) introduced by Cootes \etal~\cite{Cootes1995ASM,cootes:eccv1998,cootes:pami2001} for the analysis of planar objects and later extended to  morphable models for the analysis of the shape of 3D objects such as 3D human faces~\cite{blanz1999morphable} and bodies~\cite{allen2003space}. 

To allow the shape to deviate from the statistical model, some methods append an additional deformation field $\deformation_{\text{r}}$, which can be applied to the shape in a neutral pose, \ie:
\begin{equation}
    \neuralsurfacefunc= \transformation(\mean{\neuralsurfacefunc}  + \deformation_{\text{s}} + \deformation_{\text{r}}).
    \label{eq:template_deformation_refinement1}
\end{equation}

\noi or after posing the shape:
\begin{equation}
    \neuralsurfacefunc= \transformation(\mean{\neuralsurfacefunc}  + \deformation_{\text{s}}) + \deformation_{\text{r}}.
    \label{eq:template_deformation_refinement2}
\end{equation}

\noi In both cases, $\deformation_{\text{s}}$ is defined using Eqn.~\eqref{eq:pca_deformation}.

\textbf{The coarse deformation field $\transformation$ (\Cref{fig:deformation-based}-(c)).} While highly popular, displacement fields, including morphable models,  are only suitable for 3D objects that undergo small elastic deformations. However, natural objects such as humans and animals undergo large and complex articulated motions. These are captured using the coarse deformation field $\transformation$. It is defined using a kinematic chain, \ie a set of joints linked together with bones, forming a 3D skeleton. Each bone $i$ has a degree of influence  $w_i$ on the points $\meanpoint$ on the template shape. In other words, a point $\meanpoint$ on the template shape is transformed using the kinematic chain into a new point $\point$ as follows:
\begin{equation}
    \point = \transformation(\meanpoint) = \sum_i \transformation_i w_i \meanpoint \text{ with } \transformation_i \in \SEthree \text{ and } w_i\in\real.
\end{equation}

\noi The weight $w_i\in\real$ defines the corresponding skinning weight for the $i-$th bone. This is referred to as LBS. 

A popular model that exploits this formulation is the  Skinned Multi-Person Linear Model (SMPL) of Loper \etal~\cite{loper2015smpl}, which:
\begin{itemize}
    \item Decomposes $\deformation$ into an identity-dependent shape, which is a morphable model~\cite{allen2003space} but learned from pose-normalized body shapes, and non-rigid pose-dependent shape details.

    \item Drives the nonlinear transformation $\transformation$ using a skeletal rig composed of $\nsmplposes = 23$ joints. A $k$-th joint,  represented with a $3\times 3$ rotation matrix   $\pose_k \in \rotations$, defines the relative rotation of the $k-$th body part with respect to its parent part in the kinematic tree. In total, the pose is defined using $9 \times \nsmplposes + 9$ parameters, \ie  $9$ for each part in the skeletal rig plus $9$ for the root orientation.
\end{itemize}

\noi SMPL and its variants, such as SMPL-X~\cite{Pavlakos_2019_CVPR}, have been specifically designed for human bodies. Their main advantage is that their low-dimensional pose and shape parameters make it easier for deep neural networks to learn the high-dimensional human body meshes. Thus, they have recently been extended to represent and reconstruct animal body shapes and poses. For instance, Zuffi \etal~\cite{zuffi2017menagerie} introduced the Skinned Multi-Animal Linear (SMAL) model to jointly represent the 3D shape and pose of multiple quadruped animal species. The model is learned from 3D scans of $41$ different quadruped animal toys. Instead of using a single holistic template, Liu \etal~\cite{liu2024lepard} decompose an articulated animal shape into parts, following its kinematic structure. The representation is then fitted to an observation by learning local deformations of each primitive surface to match the image evidence. 

\begin{figure}[t]
    \centering
     \includegraphics[trim={7cm 5.3cm 5cm 0cm},clip,width=\linewidth]{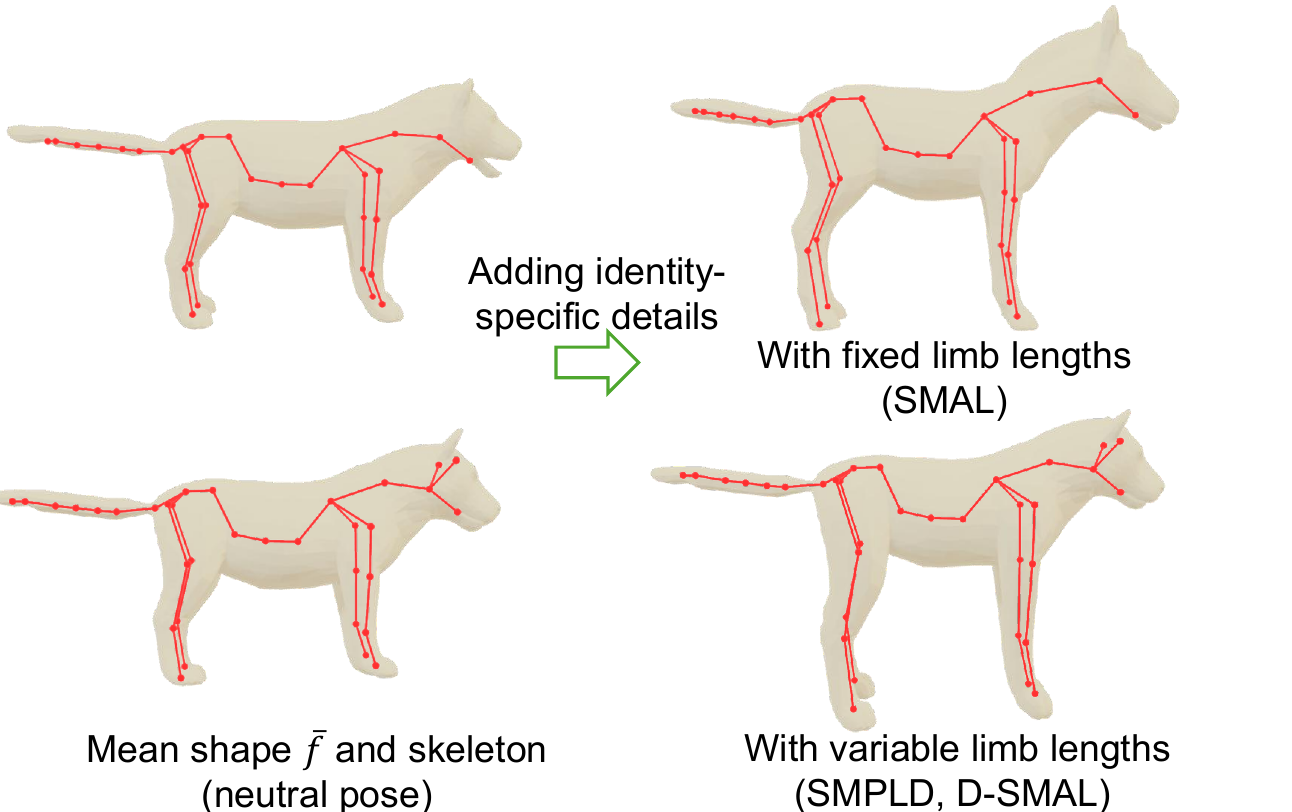}     
    \caption{Illustration of the difference between SMAL model~\cite{zuffi2017menagerie} and models such as SMBLD~\cite{biggs2020who} and D-SMAL~\cite{ruegg2023bite}, which augment SMAL model with variable bone/limb lengths to increase expressiveness.}
    
    \label{fig:smal_bonelength}
\end{figure}

Since its introduction, SMAL~\cite{zuffi2017menagerie} has been extended, customized, and fine-tuned to fit specific animal species such as quadruped animals, birds, and even underwater creatures such as dolphins; see~\Cref{fig:smal_bonelength}.

\vspace{6pt}
\noi \textbf{(1) Quadruped animals.} Unlike humans, quadruped animals exhibit large intra-class variability. In particular, since SMAL is generic,  the shape space is not constrained enough to prevent undesired, unrealistic shapes, \eg a dog should not ever have the proportions of a hippo or a horse. Also, the shape space is too limited and not expressive enough to represent subtle, but important species-specific variations (\eg the shapes and articulations of different dog’s ears). To address these issues, several papers train SMAL on specific quadruped species, \eg dogs as in  D-SMAL~\cite{ruegg2023bite}.

Also, the body parts can significantly vary in scale within and across different species, \eg dog and horse breeds. To account for this, SMBLD~\cite{biggs2020who}, BARC~\cite{ruegg2022barc}, and Tan \etal~\cite{tan2024distilling} for dogs, and hSMAL~\cite{li2021hsmal} and VAREN~\cite{zuffi2024varen} for horses add part scale parameters to independently scale different body parts of quadruped animals to better capture the diverse shapes and body proportions of breeds. In particular, 
hSMAL~\cite{li2021hsmal}  uses a new template, skeleton structure, and shape space learned from the 3D scans of toys representing $37$ different breeds and ages of horses. Zuffi \etal~\cite{zuffi2024varen} extended hSMAL to hSMAL+  by improving its resolution from $1,500$  to $3,647$ vertices and adding body part scaling defined in the model shape space.  hSMAL+ adjusts the model's rest pose to match the posture of real horses better. It also adds linear scaling of a set of body parts, such as limbs. The model is trained on a large number of 3D scans of real horses.


\vspace{6pt}
\noi \textbf{(2) Birds.}  Badger \etal~\cite{badger2020bird} introduced Skinned Linear Articulated Bird Model, the first parameterized avian mesh model that adapts SMAL to birds. Compared to SMPL and SMAL, the method does not have access to 3D groundtruth data to learn the variations in the shape of birds.  Instead,  the method uses a predefined bird template as the mean shape. It includes an additional degree of freedom per joint to model the distance between parent and child joints, thereby capturing variation in the relative length proportions of the body and limb segments.  Also, when birds perch, their wings fold in on themselves. This large deformation cannot be well captured by the LBS of a single bird  model.  Badger \etal~\cite{badger2020bird}  overcome this limitation by using two templates with identical mesh topology, bones, skinning weights, and keypoints but in two different initial poses: one for birds with their wings outstretched and another for birds with their wings folded.

\vspace{6pt}
\noi \textbf{(3) Dolphins.} Baieri \etal~\cite{baieri2025model}  proposed an articulated dolphin model to enable metric reconstruction of the 3D shape and motion   of wild dolphins from monocular videos. Different from SMPL and SMAL, the  dolphin body template is decomposed into parts, with deformation spaces defined for each part. Thus, unlike SMPL and SMAL, the shape variables in the dolphin model are a matrix rather than a vector. To ensure that the resulting model is smoothly connected, the method computes joint displacements to connect each part to its parent in the kinematic tree.


\subsection{Learning parametric shape models}
\label{sec:learning_parameteric_models}

Parametric animal shape models~\cite{zuffi2017menagerie,Zuffi2018LionsTigersBears,Zuffi2019Safari,li2021hsmal} provide strong priors that significantly facilitate the 3D reconstruction process. However, these models need to be learned, in an offline step, from 3D data such as 3D meshes and scans, and then fitted to image(s) or video(s) for 3D reconstruction. For instance, the template shape $\mean{\neuralsurfacefunc}$ of Eqn.~\eqref{eq:template_deformation}  and the deformation basis $\left(\eigenvector_1, \dots, \eigenvector_d\right)$ of Eqn.~\eqref{eq:pca_deformation}, referred to as eigenshapes, can be learned from a population of 3D objects $\{\neuralsurfacefunc_1, \dots, \neuralsurfacefunc_\nshapes\}$ using Principal Component Analysis (PCA).  We refer the reader to the surveys of Laga \etal~\cite{laga2018survey} and Egger \etal~\cite{egger20203dmorphable} for a detailed review of these statistical shape models.

Learning parametric  shape models from 3D data is only possible for generic objects, whose 3D models are widely available, and for cooperative subjects such as humans, which can be brought in large numbers to 3D scanning studios. This is not the case with animals, which are non-cooperative.  Some papers overcome the 3D data scarcity issue by scanning animal toys and statuettes, as done in the SMAL model~\cite{zuffi2017menagerie} and its variants~\cite{Zuffi2018LionsTigersBears,li2021hsmal}. However, such statistical 3D models, which are learned from limited 3D data, do not capture the entire shape variability and are limited to a few common species such as horses and dogs.  Other papers, \eg VAREN~\cite{zuffi2024varen}, learn a realistic 3D articulated shape model of horses, by putting effort into gathering a large dataset of real 3D scans. While the effort is worth it for such domestic animals of high importance, it does not scale to other species, especially species in the wild. Thus, recent techniques focus on the joint 3D reconstruction and statistical shape model learning, in a self-supervised manner, from RGB images and videos. These techniques will be discussed in~\Cref{sec:joint_statistical_reconstruction}.


Finally, unlike free-form deformation-based representations, parametric models such as SMAL and its variants are category-specific, meaning that one needs to train one SMAL model per animal species since the body structure of animals and the range of motions they can perform significantly differ within and across species. Also, these methods implicitly assume that the shape variability follows a Gaussian distribution while the articulated motion is governed by a skeleton rig of a fixed structure that is known in advance. Thus, they cannot represent shape, pose, and articulation variability across species.

\subsection{Reconstruction}
\label{sec:reconstruction_pmodels}


Once a parametric shape model is learned offline using groundtruth 3D data, the 3D reconstruction process reduces to that of fitting the model to the input RGB image(s) and/or video(s). This is done either via optimization~\cite{Kanazawa2015Learning3D,zuffi2017menagerie,Zuffi2018LionsTigersBears,li2021hsmal} (\Cref{sec:parametric_models_optimization}) or by training neural networks that regress the shape and pose parameters~\cite{Zuffi2019Safari,badger2020bird,li2021coarse} (\Cref{sec:parametric_models_learning}).

\subsubsection{Optimization-based methods}
\label{sec:parametric_models_optimization}

Optimization-based methods such as~\cite{Kanazawa2015Learning3D,zuffi2017menagerie,Zuffi2018LionsTigersBears,li2021hsmal} estimate the articulated pose and shape parameters by optimizing an objective function that is a weighted sum of a data term and a regularization term. The data term, which measures the discrepancy between the estimated 3D shape and the groundtruth 3D shape, is defined using 2D cues such as 2D keypoint reprojection loss~\cite{Zuffi2018LionsTigersBears} 
and a silhouette loss~\cite{Zuffi2018LionsTigersBears}. 
The regularization term is used to impose some constraints. For instance, Zuffi \etal~\cite{Zuffi2018LionsTigersBears} 
\begin{itemize}
    \item Observe that the optimization is better behaved if one estimates different shapes for each frame while adding a penalty on their differences across frames.

    \item Encourage the camera's focal length to be greater than $500$ and the 3D shape to be in front of the camera.
    
    \item Limit the location of the joints.
    
    \item Penalize the deviation of the poses and the shape from the distribution of the poses and shapes estimated from SMAL training data. 

    \item Enforce As-Rigid-As-Possible deformations~\cite{Zuffi2018LionsTigersBears}.
    
    \item Impose symmetry with respect to the main axis of the animal body~\cite{Zuffi2018LionsTigersBears}, 
    
    \item Encourage smooth reconstructions by adding a Laplacian smoothing term~\cite{Zuffi2018LionsTigersBears}.

\end{itemize}

\noi SMAL with Refinement (SMALR)~\cite{Zuffi2018LionsTigersBears} follows this approach to recover the instance-specific shape of an animal from multiple views, where the animal may be in a different pose in each image.  It first fits the SMAL model to each of the input images. It then estimates the vertex displacement field $\deformation_r$ of Eqn.~\eqref{eq:template_deformation_refinement1}, which   
refines the shape in a canonical pose, allowing it to deviate from SMAL. This way, the recovered 3D shape, when articulated, better explains the image evidence in all views. The displacement field $\deformation_r$ is estimated by minimizing an objective function (while maintaining pose and shape parameters fixed) composed of a keypoint loss and a silhouette loss while enforcing As-Rigid-As-Possible (ARAP) deformation, symmetry with respect to the main axis of the animal body, and Laplacian smoothing. 


\subsubsection{Leaning-based methods}
\label{sec:parametric_models_learning}

Learning-based methods train deep neural networks that directly regress shape, pose, and appearance. Compared to optimization-based methods, learning-based techniques are usually faster but require training on large datasets. Kanazawa \etal~\cite{Kanazawa2018CategorySpecific} and Zuffi \etal~\cite{Zuffi2019Safari} were among the first to propose a learning-based technique for fitting a template to image observations of animals. Kanazawa \etal~\cite{Kanazawa2018CategorySpecific} represent the template use triangular meshes and uses an MLP to regress the vertex displacements. 
Zuffi \etal~\cite{Zuffi2019Safari} fitted the parametric SMAL model to RGB images. It  trains a neural network composed of an encoder that maps an input into a latent feature vector, followed by multiple heads that decode the feature vector into a texture map, a displacement field $\deformation_{s} $, a 3D pose, and camera parameters. The shape decoder predicts the displacement field $\deformation_{s}$ using a fully connected layer that computes a feature vector $\featurevector_c$ followed by a linear layer that predicts $\deformation_{s} = \weights\featurevector_c + \bias$ where $W$ are the weights of the layer and $\bias$ the bias. The weights $\weights$ are initialized using SMAL blendshapes.  The pose prediction module is a linear layer that outputs a vector of 3D poses as relative joint angles, expressed as Rodrigues vectors. The network is trained end-to-end using a mask loss, a 2D keypoint reprojection loss, and  3D supervision loss terms. Also, the prediction of the texture map enables the method to perform unsupervised per-instance optimization (refinement), exploiting the learned feature space and a photometric loss.

Subsequent methods follow the same idea but introduce additional blocks or incorporate changes to address specific issues. For instance, Aygun \etal~\cite{aygun2024saor} deformed a sphere, using an MLP, and  articulate the
deformed shape using linear skinning~\cite{lewis2023pose} in a skeleton-free manner~\cite{liao2022skeleton} to obtain the final shape. As most natural objects exhibit
bilateral symmetry, similar to~\cite{Kanazawa2018CategorySpecific}, the method only deforms the vertices of the zero-centered initial shape vertices that are located on the positive side of the xy-plane. It then reflects the deformation for the vertices on the negative side.

Ruegg \etal~\cite{ruegg2022barc} observed that dogs are not only highly articulated but exhibit a wide range of shapes and appearances across breeds. Thus, in addition to predicting the neutral shape parameters from a latent representation of the input image, they also predict the breed class. They also introduce a breed triplet loss, similar to the one used in person identity~\cite{taigman2014deepface,schroff2015facenet}, as well as an auxiliary breed classification loss. The former is used to capture intra-breed variations by applying the triplet loss on the latent encoding in the shape branch. The latter is a standard cross-entropy loss used to bias the estimation towards recognizable, breed-specific shapes.

Instead of driving the template deformation using skeletons, Kokkinos \etal~\cite{kokkinos2022learning} used a small number of local, learnable handles whose deformations are predicted using an MLP. During training, the method exploits temporal information, via motion-based cycle loss, to enforce consistency across consecutive 3D reconstructions.

Some methods add a refinement block to enable coarse-to-fine reconstruction. Li \etal~\cite{li2021coarse} refined the SMAL model-based reconstruction using a Graph Convolutional Network (GCN) that predicts a displacement field to accurately align the vertices of the reconstructed mesh to the observations; see Eqns.~\eqref{eq:template_deformation_refinement1} and~\eqref{eq:template_deformation_refinement2}. 
Badger \etal~\cite{badger2020bird} refined the 3D reconstruction using an optimization approach similar to SMPLify~\cite{bogo2016keep}. However, unlike SMPLify~\cite{bogo2016keep}, they fit the model to semantic keypoints, rather than joint locations, automatically detected using a keypoint and mask estimation network.  BARC+~\cite{ruegg2023bite}, a dog-specific 3D reconstruction method,  extends BARC~\cite{ruegg2022barc} by using D-SMAL model~\cite{ruegg2023bite}, instead of SMAL. The network consists of a stacked hourglass that predicts 2D keypoints and a segmentation mask, followed by shape and pose prediction branches, where the former predicts 3D shape parameters and the camera viewpoint (translation and focal length). The latter predicts pose parameters. BITE~\cite{ruegg2023bite}  adds a refinement stage to increase the accuracy of the BARC+ predictions. To that end, the
predicted 3D mesh is projected to the image, giving rise to
2D keypoints and a silhouette. These are combined with
the intermediate keypoint and silhouette predictions from
BARC+~\cite{ruegg2023bite}  and with the image itself, thus implicitly specifying the residual error of those intermediate predictions. All these intermediate results are fed into a 2D encoder, followed by fully connected heads that output refined 3D pose and camera parameters. For images in which the dog is on
flat ground, refinement is supervised with a loss that encourages an anatomically and physically consistent placement
of the 3D model on a local ground plane.

A major limitation of these methods is that they are species-specific. Unlike humans, animals exhibit high intraclass and interclass variability, which is difficult to capture using a single statistical model such as SMAL or deep neural networks. A few papers have attempted to reconstruct multiple species within a single network, but are limited to quadruped animals. For instance, AniMer~\cite{lyu2025animer}, inspired by HMR2.0 for humans~\cite{goel2023humans}, encodes the input images using a Vision Transformer (ViT) and decodes them into SMAL parameters using a transformer-based decoder. However, unlike HMR2.0~\cite{goel2023humans}, the transformer is made family-aware using a family-supervised contrastive learning scheme. In particular, animals exhibit at least two levels of shape differences: inter-family level and intra-family level. For example, dogs share similar shapes with each other, yet they share distinct shapes
with cows. Also, different dog breeds can significantly differ in shape, \eg they can have different body posture and leg lengths. To address this, Ruegg \etal~\cite{ruegg2022barc} also predicted the breed class. During training, it uses   breed triplet loss   to capture intra-breed variations. AniMer~\cite{lyu2025animer}, on the other hand, employs a learnable class token to represent the animal family. The token, together with a minibatch of images, are fed into a ViT encoder followed by an MLP head to generate the animal family feature from the class token. The method then applies supervised contrastive learning to the family feature.

Compared to other methods, AniMer~\cite{lyu2025animer} is able to handle a wide range of quadruped animal species. However, similar to model-based methods, its solution space is limited to the pose and shape space of the SMAL model. 

Note that the 3D pose of an animal, represented in the form of either 3D locations or rotations of the skeleton joints,  is usually regressed using MLPs~\cite{Zuffi2019Safari,biggs2020who,aygun2024saor,tan2024distilling}.  Wu \etal~\cite{wu2023magicpony} and Li \etal~\cite{li2024learning}, on the other hand, used a transformer to account for the inter-dependency of the bones implied by the underlying skeleton. Specifically, they use a small transformer network~\cite{vaswani2017attention} that takes as input tokens and outputs bone rotations parameterised as Euler angles.

\textbf{Finally}, estimating the camera parameters such as translation, rotation, and focal length is a critical first step for learning 3D shapes. Existing methods commonly use MLPs~\cite{Kanazawa2018CategorySpecific, Zuffi2019Safari, li2020self, biggs2020who, aygun2024saor}. Different from these works is the approach of Wu \etal~\cite{wu2023magicpony}, which  employs a convolutional encoder to generate four viewpoint rotation hypotheses. The network also predicts, for each hypothesis,  a score that is used to evaluate the probability that the hypothesis is the best of the four options. Note  that, when multi-view images or video streams of the scene are available, one can  use common Structure-from-Motion techniques, such as COLMAP, to estimate the camera's intrinsic and extrinsics.

\subsection{Joint reconstruction and template learning} 
\label{sec:joint_statistical_reconstruction}
While powerful, parametric shape models need to be learned from 3D and 4D ground-truth data, which is difficult to collect in the case of animals.  Some methods address this issue by assuming that a hand-crafted representative 3D template for the animal category is available~\cite{kulkarni2020articulation,badger2020bird,kokkinos2022learning}. Others attempt to learn the template $\mean{\neuralsurfacefunc}$ and the articulation from a large collection of images or videos, and then regress an instance-specific displacement field $\deformation$ and an articulated motion $\transformation$ that deform the template to the image(s) or video(s) of a given animal~\cite{Kanazawa2018CategorySpecific,li2020self,li2020online}.

For example, Kanazawa \etal~\cite{Kanazawa2018CategorySpecific} treated the mean shape $\meanshape$ as a learned bias term for the predicted  $\deformation$. Thus, it can be estimated, during the training phase, from a large number of images in the same way as the other network parameters. The training process is guided by sparse keypoint annotations and segmentation masks. The method also assumes that the objects are symmetric and imposes this constraint during training. This approach has been adopted and extended in many ways. For example, instead of using a single template, Li \etal~\cite{li2020online} defined the template as a weighted combination of $n$ shape bases $\{\mean{\neuralsurfacefunc}_i \}_{i=1}^n$, \ie $\mean{\neuralsurfacefunc}  = \sum_{i=1}^n \beta_i\mean{\neuralsurfacefunc} _i$, obtained by applying K-Means clustering to all the meshes reconstructed by~\cite{Kanazawa2018CategorySpecific}. Thus,  the framework is completely free of 3D supervision.  The method relaxes the symmetry constraint, allowing the reconstruction of animals in complex poses, \eg when an animal rotates its head. It also relaxes the reliance on keypoint annotations by introducing semantic consistency constraints, \ie enforcing consistency between a canonical semantic map, which defines partwise segmentation learned at the category level in 3D,  and an instance’s part segmentation in the 2D space.

MagicPonny~\cite{wu2023magicpony} represents the template using its continuous Signed Distance Field (SDF) and uses an MLP to learn a category-wise template shape shared across all instances. The method predicts the displacement field using an MLP (with symmetry constraints) and the bone rotations from an input image. The input image is encoded using DINO-ViT~\cite{caron2021dino}, which produces a set of keys and output tokens. The keys (converted to a global key using a small MLP) are used to predict the object's displacement field using an MLP, while the outputs are used to predict its appearance. The bone rotations are predicted using a network from the keys and global key produced by the DINO-ViT. The use of a transformer allows to account for the inter-bones dependencies implied by the underlying skeleton.

Extending these ideas, Li \etal~\cite{li2024learning} introduced the Semantic Bank of Skinned Models (SBSM), a method that automatically discovers a small set of base animal shapes by learning a statistical model from 2D Internet images. In particular, their method learns a pan-category deformable 3D model for more than one hundred quadruped species. The approach employs a skinned model representation and leverages pre-trained unsupervised image features from DINO-ViT to automatically identify and group similar animals. The resulting semantic bank serves as a collection of base shapes from which detailed deformations are estimated and refined through a fitting process that adapts the model to individual input images.

\section{Implicit representations} 
\label{sec:implicit_representations}

\begin{figure*}[!t]
    \centering
    \begin{tabular}{@{}c@{}c@{}c@{}}
        
         \includegraphics[width=0.28\linewidth]{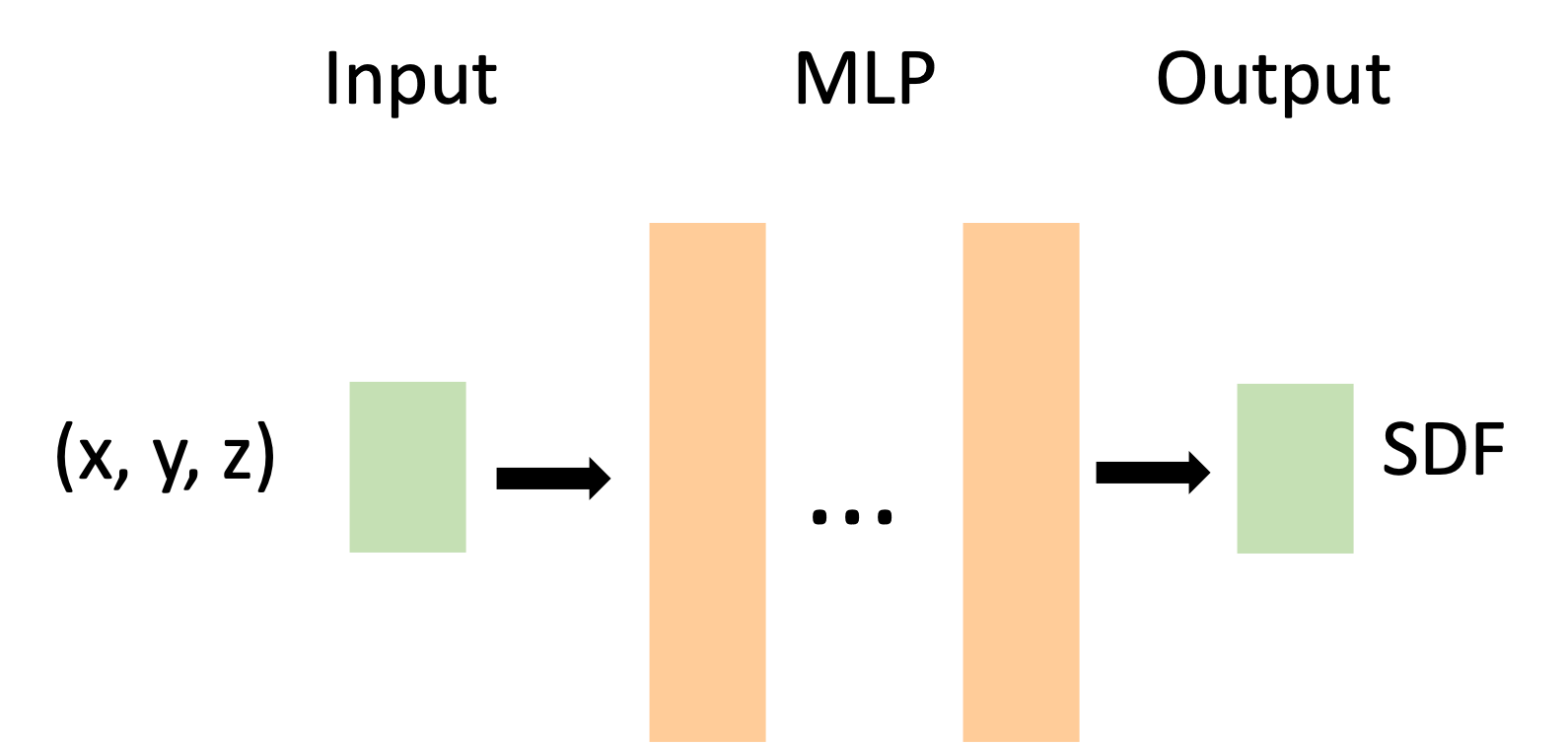} &
         \includegraphics[width=0.44\linewidth]{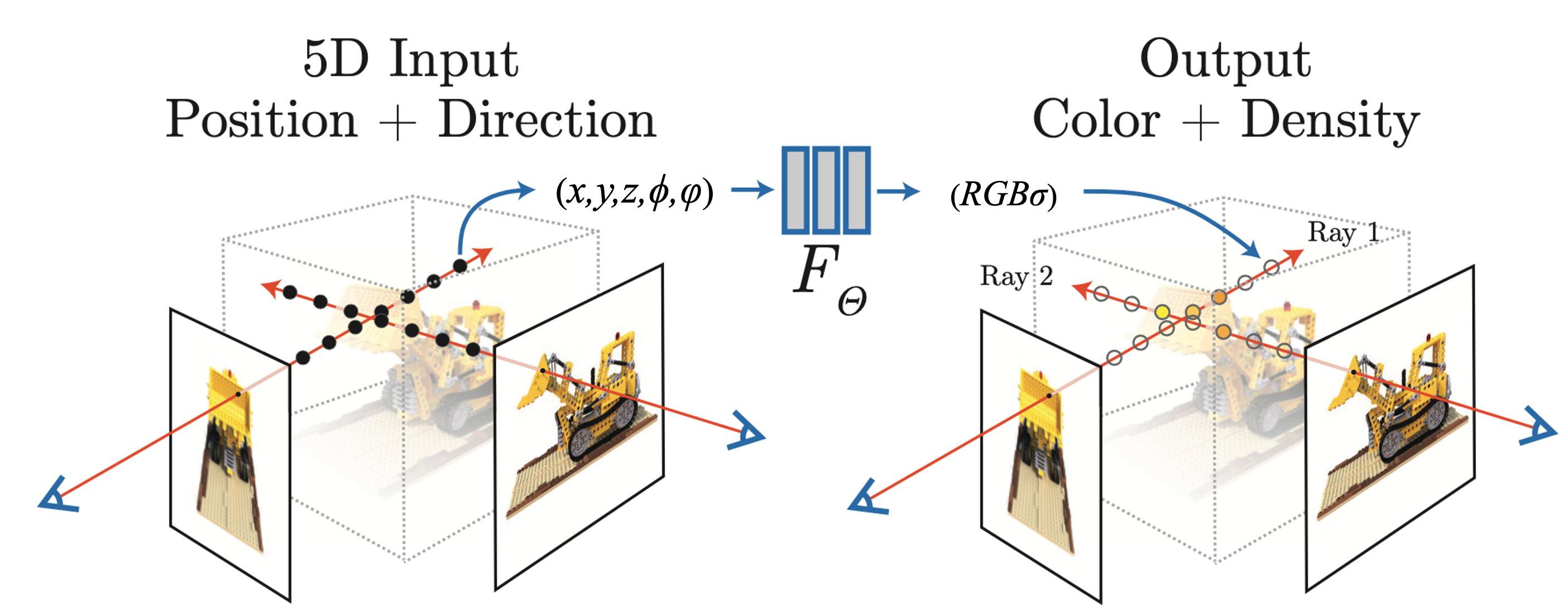} &
         \includegraphics[width=0.28\linewidth]{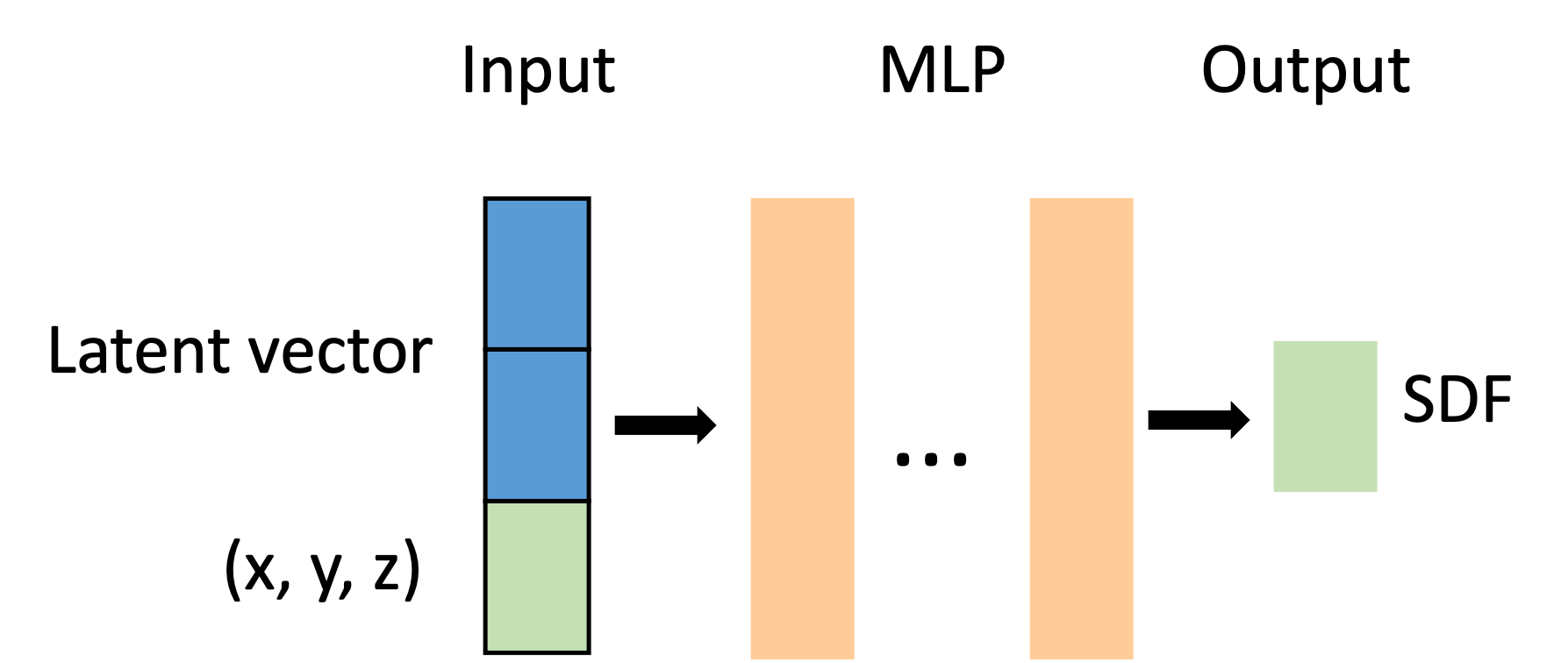}\\
         \textbf{(a)} Shape-specific SDF. & \textbf{(b)} Neural Radiance Fields (NeRF). &
         \textbf{(c)} Shape-agnostic SDF/NeRF.   
    \end{tabular}
    \caption{Neural architectures for signed distance fields representation. \textbf{(a)} Shape-specific methods train an MLP that takes a 3D point (or its frequency encoding) and regresses the SDF value at that point. \textbf{(b)} NeRF further takes as input the viewing direction and regresses the appearance. \textbf{(c)} Shape/scene-agnostic methods further condition the MLP on a latent vector learned in an auto-decoding fashion to enable the representation of multiple objects using a single neural network.}
    \label{fig:SDFNeuralField}
\end{figure*}

%


Explicit representations only represent shape boundaries, \ie they cannot represent internal shape details. Discrete explicit representations can represent shapes of arbitrary topologies, but their accuracy depends on the resolution of the discretization. Continuous ones, on the other hand, are resolution agnostic but require the parameterization of a given shape using a fixed domain $\domain$. Thus, they can only handle low-genus, usually genus-0, surfaces and cannot handle topological changes. 

In this section, we focus on implicit volumetric representations, which can represent 3D shapes of arbitrary topologies, both discretely and continuously. We will first discuss the different  implicit representations (\Cref{sec:volumetric_implicit_representations}) and then present the various 3D animal reconstruction methods that are based on these representations (\Cref{sec:volumetric_reconstruction}).

\subsection{Geometry and appearance representation}
\label{sec:volumetric_implicit_representations}

Implicit representations define the geometry of an object as a continuous volumetric function of the form  $\neuralsurfacefunc: \domain \to \real$ where   $\domain \subset \rthree$. Two types of implicit functions have been used in the literature: the Signed Distance Function (SDF) and the volume occupancy probabilities.

The SDF of a point $\point \in \domain$ is defined as the distance of that point to the closest surface point. It is set to be positive when $\point$ is outside the shape and negative if it is inside.  The surface of the shape is then the zero-level set of $\neuralsurfacefunc$, \ie $
    \shape = \{ \point \in \domain \mid \neuralsurfacefunc(\point) = 0 \}$. 
To handle partial surfaces and surfaces with holes, Li \etal~\cite{li2021complete} used Truncated Signed Distance Fields (TSDF) instead of the SDF. In TSDF, distance values are truncated to a maximum threshold, meaning only distances within a certain range from the surface are recorded, while distances beyond this range are set to the maximum threshold.

The geometry of a 3D shape or scene can also be represented using occupancy probabilities of the form $\neuralsurfacefunc: \domain \subset \rthree \rightarrow [0, 1]$. The function $\neuralsurfacefunc$ maps a point $\point \in \rthree$ to a probability value that indicates the likelihood of $\point$ being inside the shape. The surface of the shape is then defined by the decision boundary at a threshold $\threshold$, typically set to $0.5$, \ie $\shape = \{ \point \in \domain \mid \neuralsurfacefunc(\point) = \threshold \}$. Unlike SDFs and TSDFs, occupancy probabilities can handle uncertainties in data, such as noise or partial observations.


The representation can be further extended to represent \textbf{appearance}, \ie color. Since color is a function of the point of interest $\point$ and the viewing direction $\viewdir\in \stwo$ from which the point is viewed,  the overall representation becomes a 5D function of the form $\neuralsurfacefunc: \domain \times \stwo \to \real \times [0, 1]^3$. It maps a point $\point\in \domain \subset \rthree$ and a viewing direction $\viewdir \in \stwo$ to its geometry and RGB color.  Termed Neural Radiance Fields (NeRF)~\cite{mildenhall2020nerf,mildenhall2021nerf}, this representation allows novel view synthesis from arbitrary viewpoints. The original formulation represents geometry using volume density. Thus, it is not efficient in 3D reconstruction. NeuS~\cite{wang2021neus} extended this formulation by replacing density with SDF, allowing accurate 3D reconstruction and novel view synthesis.

The function $\neuralsurfacefunc$ can be represented in a discrete way by storing its values in the voxels of a volumetric grid. However, memory requirements increase cubically with the resolution, and thus, they are not suitable for representing 3D shapes with fine geometric details. This is particularly problematic with radiance fields since the domain of the function $\neuralsurfacefunc$ is of dimension five. 
Several works have shown that the implicit representation $\neuralsurfacefunc$ can be treated as a continuous function by parameterizing it using Radial Basis Functions~\cite{carr2001reconstruction,kojekine2004surface,ohtake2005multi}, or MLPs~\cite{piperakis2001affine,piperakis20013d,Mescheder2018OccupancyNL}; see~\Cref{fig:SDFNeuralField}. Whether discrete or continuous, an implicit representation can be converted on the fly into an explicit mesh via Differentiable Marching Tetrahedra (DMTet)~\cite{shen2021deep}.

One main limitation of this representation is that it is object/scene-specific. In other words, one needs to train a separate neural network for every 3D shape or scene; see~\Cref{fig:SDFNeuralField}-(a)-(b). Park \etal~\cite{Park2019DeepSDFLC} addressed this issue by conditioning the SDF on a latent code $\latentcode$ that is specific to each shape; see~\Cref{fig:SDFNeuralField}-(c). In other words, for a given shape of index $i$:
\begin{equation}
    \neuralsurfacefunc_{\Params}(\point,  [\viewdir]; \latentcode_i) = \neuralsurfacefunc_i(\point, [\viewdir]).
\end{equation}

\noi Here,  $\Params$ denotes the learnable parameters of the representation. When dealing with 3D geometry, $\neuralsurfacefunc_{\Params}(\point, \latentcode_i)$ represents the predicted SDF value at the point $\point$, conditioned on the latent code $\latentcode_i$. When dealing with appearance, $\neuralsurfacefunc_{\Params}$ further takes as input the viewing direction $\viewdir$. To train such a network, DeepSDF~\cite{Park2019DeepSDFLC} uses an auto-decoder architecture. Unlike traditional auto-encoders,  the auto-decoder removes the encoder. Instead, it initializes a latent vector $\latentcode_i$ for each shape and optimizes it directly during training, alongside the network parameters. This formulation allows learning a continuous latent space. Thus, it enables exploring the shape variability by navigating in the latent space. 

Neural representations have also been extended to represent \textbf{dynamic objects and scenes} in two ways; The \first class of methods~\cite{sinha2023common} maps the temporal observations into a canonical space, using non-rigid deformation fields, and then uses NeRF~\cite{mildenhall2020nerf,mildenhall2021nerf} to represent the animal shape in the canonical space. The deformation field can be parameterized using MLPs. The \second class of methods~\cite{jiang2023consistent4d} treats the shape of a deforming object as a time-parameterized function, \ie 
$\neuralsurfacefunc_\Params: \domain \subset \rthree \times \stwo \times \rplus \rightarrow \rthree \times [0, 1]$, where the last dimension in the input domain refers to time. 

Originally proposed for the representation of generic 3D  objects and human body shapes, neural representations such as DeepSDF~\cite{Park2019DeepSDFLC} and NeRF~\cite{mildenhall2020nerf,mildenhall2021nerf} are becoming popular for representing the  shape and appearance of 3D~\cite{yang2022banmo} and 4D (\ie moving) animals~\cite{sinha2023common,jiang2023consistent4d,cheng2023virtual,tan2024distilling}.

\subsection{Reconstruction methods}
\label{sec:volumetric_reconstruction}

Early methods use a discrete volumetric grid, which can be decoded from an input embedding and processed using 3D convolutional networks. For instance, Li \etal~\cite{li2021complete} introduced 3DComplete, which takes partial scans converted to TSDF and scene flow in the form of a Volumetric Motion Field (VMF), and extracts a 4D time-space embedding using a 4D encoder. The method then jointly infers the missing 3D geometry and motion field using two decoders operating in parallel: a motion decoder and a shape decoder. The decoding is performed at four hierarchical levels. At each level, the shape decoder predicts the geometry and passes it to the corresponding layer in the motion decoder and the following hierarchy level in the shape decoder.  The method is trained in a supervised manner using a synthetic dataset called DeformingThings4D~\cite{li2021complete}, which consists of $1{,}972$ animation sequences spanning $31$ different animals and humanoid categories with dense 4D annotation. 

While straightforward, 3D decoders are very expensive in terms of memory requirements. Thus, early methods that used these architectures are limited to low-resolution 3D reconstruction, and append upsampling and refinement blocks to recover higher-resolution 3D geometry. 

Continuous implicit neural representations have been extensively used for 3D and 4D animal reconstruction due to their resolution-agnostic property, which enables the representation of fine-grained geometric details, and their ability to be trained in a self-supervised manner. Existing implicit neural representations for 3D and 4D animal reconstruction follow the template deformation-based pipeline described in~\Cref{sec:parametric_models} but represent the template using a neural implicit function learned from RGB images and/or videos. They differ in the way the template is learned and the way it is deformed to match the observations.

Existing methods define the template, or canonical shape, as the 
level set of an occupancy field~\cite{lei2022cadex} or signed distance function~\cite{yang2022banmo,yang2023Reconstructing,wu2023magicpony,jiang2023consistent4d,li2024learning}, parameterized with a neural field. Most methods use a single, species-specific template~\cite{lei2022cadex,yang2022banmo,yang2023Reconstructing,wu2023magicpony,jiang2023consistent4d}. However, these category-specific methods fail to generalize to rare species with limited training images.  3D-Fauna of Li \etal~\cite{li2024learning} addresses this issue by learning, from Internet images,  a single joint model of $100+$ species in one go. The method first learns, using a DinoV2-based encoder~\cite{caron2021dino,oquab2023dinov2}, $K$ basis codes, referred to as a semantic memory bank. The latent code of an input image is projected onto these basis codes to obtain a new code, which is then decoded into a base shape, or template, using an MLP, following MagicPony~\cite{wu2023magicpony}. 

Once a representative template is learned, it is deformed to match the input RGB image~\cite{wu2023magicpony,li2024learning},  RGB video~\cite{yang2022banmo, yang2023Reconstructing,sinha2023common,jiang2023consistent4d,cheng2023virtual}, or a sequence of point clouds~\cite{lei2022cadex} of the animal. Existing methods differ in the way the deformations are represented. Lei \etal~\cite{lei2022cadex} modeled a nonrigid shape as the deformation of a canonical shape in the form of bijective mappings (homeomorphisms) that map the canonical shape to each of the input frames.  The homeomorphisms are defined as functions that map each deformed 3D point to its corresponding canonical 3D point.   The method uses Conditional Real-NonVolume Preserving~\cite{dinh2016density} or the Nonlinear Independent Component Estimation~\cite{dinh2014nice} normalizing flow implementations to learn the homeomorphisms. Yang \etal~\cite{yang2022banmo} followed a similar idea as CADEX~\cite{lei2022cadex}, but the forward and backward mappings are driven with species-specific skeleton bones. The method estimates both the skinning weights and the bone deformations using MLPs.

Other methods, similar to those described in~\Cref{sec:parametric_representation}, decompose the template deformation into skeleton-based deformations and surface-based displacements. Both are predicted using MLPs\cite{wu2023magicpony,yang2023Reconstructing,li2024learning}.  These methods learn animatable 3D animal models that capture both skeletal articulations and surface details. At runtime, they deform the template, using the learned deformation field, and pose it, using the learned skeleton, to match the input image. The deformations are estimated using MLPs conditioned on the latent code of the input image extracted from DINOv2's self-supervised encoder. For example, Yang \etal~\cite{yang2023Reconstructing} took as input a video collection of animals within the same species and a predefined skeleton structure. It then learns \textbf{(1)} video-specific canonical shape composed of a template, joint locations, skinning weights, and morphology code, and \textbf{(2)} the deformation of the canonical shape model within the video. Yang \etal~\cite{yang2023Reconstructing} defined the video-specific template following deepSDF~\cite{Park2019DeepSDFLC}, \ie an MLP that estimates the SDF at each point $\point$ in the canonical space, conditioned on the morphology code. Similarly, the joint locations and the skinning weights are estimated using MLPs conditioned on the morphology code. The model is then fitted to the input video by learning a deformation field composed of skeleton deformation, to capture large-scale coarse deformations, and surface-based deformation field following CADEX~\cite{lei2022cadex}. 

Finally, acquiring multi-view videos of animals to supervise dynamic NeRF-based techniques is challenging. Instead,  Jiang \etal~\cite{jiang2023consistent4d} exploited diffusion models, namely Zero123~\cite{liu2023zero1to3}, to generate novel images of the subject from other viewpoints and use them as supervisory signals for K-planes~\cite{fridovich2023k}-based dynamic NeRF. While promising, the performance of this type of methods degrades the farther the animal is from the neutral pose. Thus, this suggests that what these methods learn is actually symmetry.

\section{Gaussian avatars}
\label{sec:gaussian_avatars}

Implicit neural representations (\Cref{sec:implicit_representations}) can achieve high reconstruction quality but suffer from slow training and rendering because of the need to query deep and wide MLPs multiple times. On the other hand, explicit representations (Sections~\ref{sec:explicit_representations} and~\ref{sec:parametric_models}) are more efficient to render and easier to deform. They, however, often have sub-optimal quality and are restricted by the fixed mesh topology of the template, or constrained by using too many points. Gaussian Splatting (GS)-based methods~\cite{kerbl20233d,Dalal2024GaussianS3} overcome the weaknesses of both methods by explicitly approximating the implicit radiance using Gaussian Mixture Models (GMM). This section reviews the foundations of Gaussian Splatting-based representations (\Cref{sec:3dgs_representation}) and then discusses their usage in 3D and 4D animal reconstruction (\Cref{sec:3dgs_reconstruction}).

\begin{figure}[t]
    \centering
    \includegraphics[trim={0cm 12.4cm 0cm 1cm},clip,width=\linewidth]{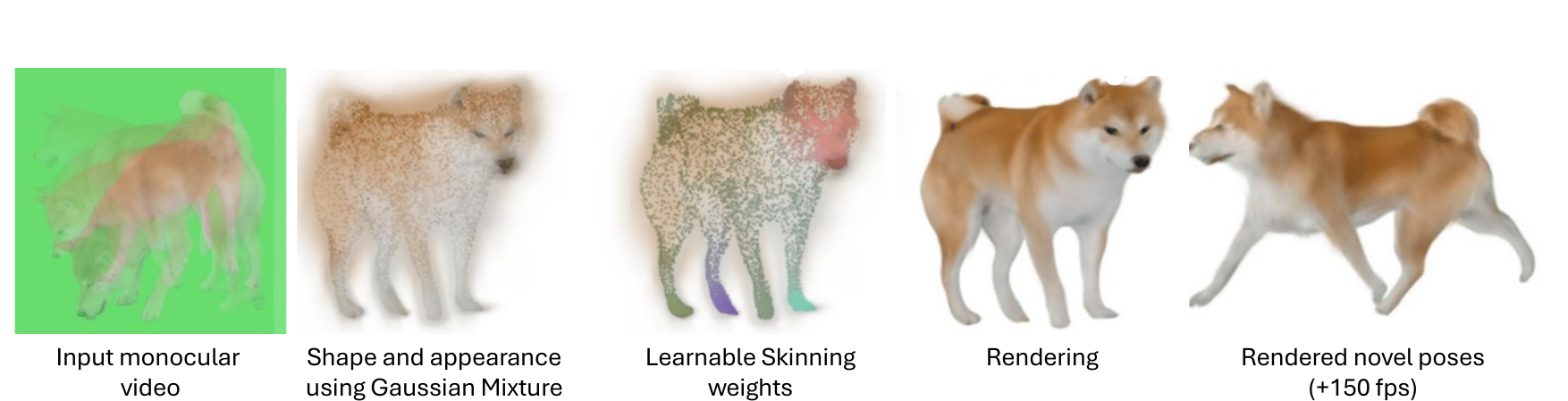}
    \caption{Gaussian Avatars of dogs. Image adapted from GART~\cite{lei2024gart}.}
    \label{fig:GaussianAvatars_GART}
\end{figure}

\subsection{Representation}
\label{sec:3dgs_representation}

3D Gaussian Splatting (3DGS)~\cite{kerbl20233d,Dalal2024GaussianS3}   combines neural fields and classical point-based graphics~\cite{zwicker2002ewa}. It represents a 3D scene using a set of Gaussians, each defined by its location $\meanpoint$ in the 3D space $\rthree$ and a covariance matrix   $\cov$.

To ensure that the covariance matrices have a physical meaning, Kerbl \etal~\cite{kerbl20233d} decomposed them  into a scaling matrix $\scalingmatrix$ and a rotation matrix $\rotation$:   
\begin{equation}
    \cov= \rotation \scalingmatrix \scalingmatrix^\top\rotation^\top.
\end{equation}

\noi Geometrically, this is equivalent to an oriented ellipsoid in 3D space. For rendering, the color at a given pixel $\pixel$ is computed by blending all the 3D Gaussians whose projections (or splats) along the viewing direction under consideration overlap the pixel $\pixel$:
\begin{equation}
    \thecolor = \sum_{i=1}^{\gaussianscount}\thecolor_i\alpha_i\prod_{j=1}^{i-1}(1 - \alpha_i).
\end{equation}

\noi Here,   $\thecolor_i$ is the view-dependent color of the $i$-th point along the ray through the pixel $\pixel$. It is modeled using spherical harmonics. The blending weight $\alpha_i$ is the learned opacity $\opacity_i$ of the Gaussian $\gaussian_i$ weighted by  its splat, \ie the 2D Gaussian $\gaussian'_i$ defined as the projection of $\gaussian_i$ onto the image plane:
\begin{equation}
    \alpha_i = \opacity_i \times \gaussian'(\pixel; \meanpixel_i', \cov_i'),
\end{equation}

\noi where $\meanpixel_i'$ and $\cov_i'$ are the 2D mean and 2D covariance matrix of $\gaussian'_i$. To learn a representation of a 3D scene, 3DGS~\cite{kerbl20233d} follows a strategy similar to NeRF~\cite{mildenhall2020nerf,mildenhall2021nerf}, in which novel views of a given scene are rendered and compared to groundtruth images using a photometric loss. The learnable parameters of the representation, \ie the center, rotation,  scales, opacity, and view-dependent color of each Gaussian, can then be optimized through back-propagation. 


3DGS-based representations have been used in modeling general dynamic scenes, where the scenes do not have specific structures (\ie articulation)~\cite{yang2023real,luiten2024dynamic,wu20244d,yang2024deformable}. Several papers started to explore their usage in building animatable avatars of animals from RGB images and videos~\cite{lei2024gart,cho2025dogrecon,zhai2025taga}.

\subsection{Reconstruction}
\label{sec:3dgs_reconstruction}

\begin{figure*}
    \centering
    \includegraphics[trim={0 3cm 0 0},clip,width=\linewidth]{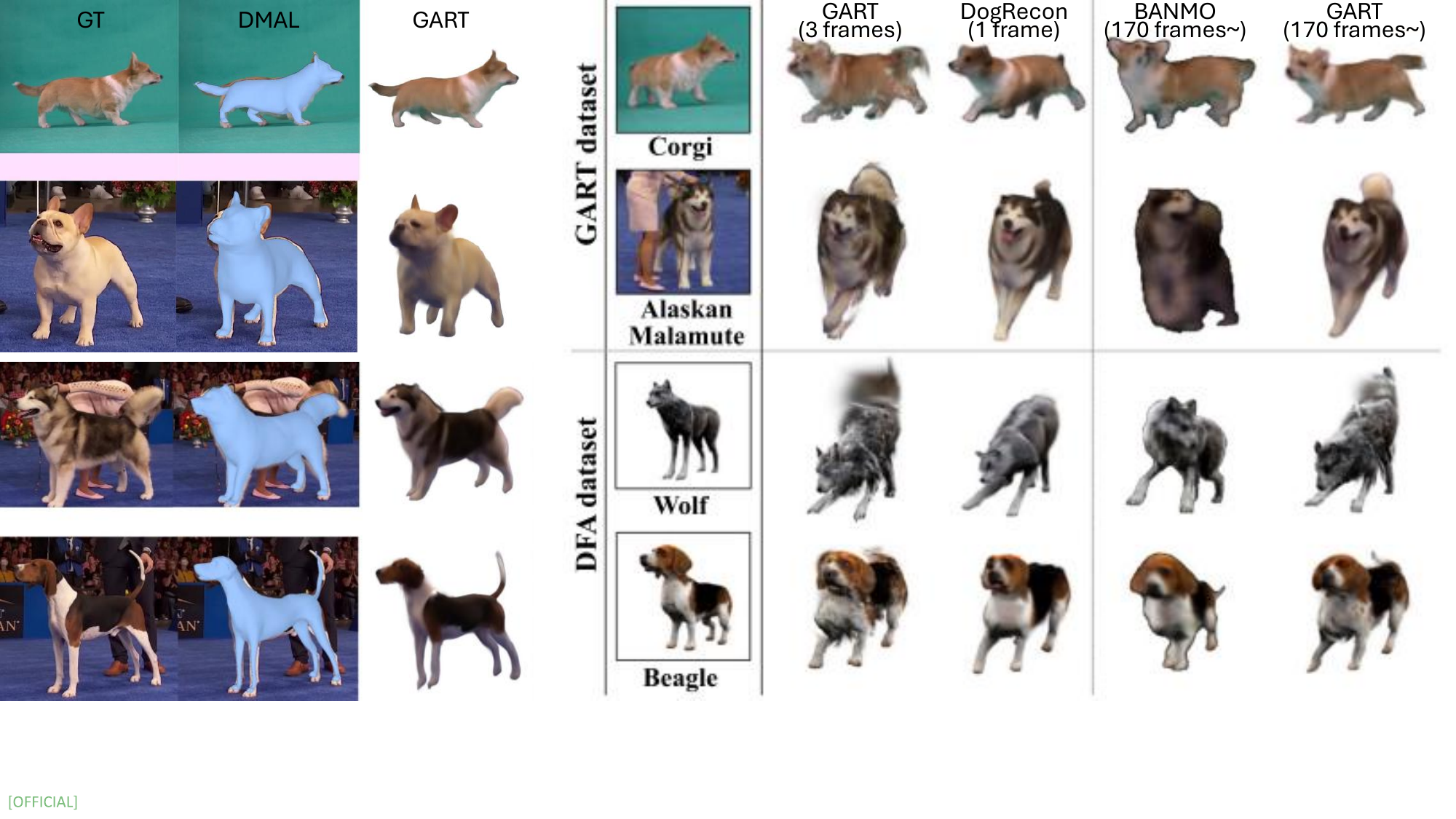}
    \caption{Comparison of Gaussian Avatar-based 3D reconstruction and rendering of dogs. Images adapted from GART~\cite{lei2024gart} and DogRecon~\cite{cho2025dogrecon}.}
    \label{fig:GaussianAvatars_comparison}
\end{figure*}

The literature on 3DGS for 3D animal reconstruction is rather sparse. The few existing methods~\cite{lei2024gart,cho2025dogrecon,zhai2025taga} that aim to recover animatable 3D animal shapes either from images~\cite{cho2025dogrecon} or videos~\cite{lei2024gart,zhai2025taga} combine the Gaussian Splatting-based representation with animatable models such as  SMPL~\cite{bogo2016keep} or D-SMAL~\cite{ruegg2023bite}, the SMPL extension to quadruped animals. 

Formally, the 3D shape of an animal is defined as the deformation of a template shape where the deformations are driven by a skeleton and LBS weights. Unlike the methods described in Sections~\ref{sec:parametric_models} and~\ref{sec:implicit_representations}, the template is represented using a mixture of 3D Gaussians~\cite{lei2024gart,cho2025dogrecon,zhai2025taga}. Each Gaussian is defined in the space of the template, and is characterized by its position, scale,  rotation, opacity, and view-dependent color~\cite{lei2024gart,zhai2025taga}. The template is then deformed with learnable skinning; see~\Cref{fig:GaussianAvatars_GART}.  However, unlike traditional SMPL and SMAL-based methods, the skinning weights are defined on a coarse voxel grid, and the skinning weight at a given surface point is determined through trilinear interpolation of its neighboring grid cells. Voxel-based skinning fields have several advantages. \textbf{First}, they enhance training stability by constraining the skinning field to predefined vertices. \textbf{Second}, skinning constraints enable smooth propagation of reliable skinning weights from sampled bone points to the surrounding 3D space, and \textbf{finally}, the linear interpolation enhances smoothness.

While  3DGS-based methods for animatable 3D animal reconstruction follow this pipeline,  introduced in~\cite{lei2024gart}, they differ in a few aspects. \First, GART~\cite{lei2024gart} builds animatable 3D models from an RGB video. DogRecon~\cite{cho2025dogrecon} builds animatable 3D dog models from a single RGB image. It takes an input RGB image and then generates multi-view images of the object using an image-conditioned diffusion model~\cite{liu2023zero1to3}. It then uses GART~\cite{lei2024gart} to fit a deformable GMM to the given image with the predicted D-SMAL~\cite{ruegg2023bite} pair and the generated images. To robustly reconstruct the 3D Gaussian dog, the method utilizes a sampling weight that can handle poorly generated images.

\Second, GART~\cite{lei2024gart} and DogRecon~\cite{cho2025dogrecon} use an articulated template while  TAGA~\cite{zhai2025taga} is template-free, \ie it does not include mesh vertices or skinning annotations. Instead, it exploits the fact that the 3D Gaussians maintain a bijective mapping between the canonical and observation space through skinning. However, without template priors, forward mapping often captures spurious correlations of adjacent body parts, leading to unrealistic geometric artifacts in the canonical pose. To alleviate this, TAGA~\cite{zhai2025taga}  defines Gaussians with spurious correlations as “Ambiguous Gaussians” and then proposes a new backward mapping strategy that integrates anomaly detection to identify and correct ambiguous Gaussians, \ie Gaussians whose skinning weights deviate significantly from their expected values. To correct the ambiguous Gaussians, the method employs a kNN-based algorithm: for each ambiguous Gaussian, it finds the k-nearest non-ambiguous Gaussians and then computes the corrected weight from the neighbors based on the maximized confidence score.

In terms of computation time, TAGA~\cite{zhai2025taga} takes around $30$ mins to train, GART~\cite{lei2024gart} can be reconstructed via differentiable rendering from monocular videos in  minutes, while DogRecon~\cite{cho2025dogrecon} takes around $6$ mins. \Cref{fig:GaussianAvatars_comparison} compares, visually, 3D reconstructions of dogs obtained using D-SMAL~\cite{ruegg2023bite}, GART~\cite{lei2024gart}, and DogRecon~\cite{cho2025dogrecon}.

\section{From reconstruction to generation}
\label{sec:generation}

Early efforts have focused on 3D reconstruction and modeling from images and videos. Recently, however, we have seen a growing interest in generating animal avatars with realistic appearances and motions from sparse visual observations, textual descriptions, or biological parameters. This has important applications in areas such as digital entertainment, wildlife conservation, animal ecology, and biomechanics. Compared to humans, animal shape and motion generation presents unique challenges due to species diversity, varied morphological structures, different behavioral patterns in response to similar environmental conditions and textual descriptions, and scarcity of 3D motion annotations paired with corresponding and detailed textural descriptions and environment geometry. 

State-of-the-art papers~\cite{cheng2023virtual,yang2024omnimotiongpt,wang2025animo} that tackled these challenges can be classified into two main categories.
The \first class of methods~\cite{cheng2023virtual} trains a Variational Auto-Encoder (VAE) that can generate animal trajectories from a learned latent code. Their advantage  is that the generation process can be conditioned on additional parameters to control the output. For instance, Cheng \etal~\cite{cheng2023virtual} conditioned the generation process on the 3D geometry of the animal's surrounding environment, the starting position and pose, as well as on additional constraints such as the position of the animal's limbs. Cheng \etal~\cite{cheng2023virtual} decomposed the generation process into two stages: the first stage trains a VAE that generates, from a trajectory latent code, the 3D trajectory of the animal given an initial 3D position of the animal, 3D points sampled from the surrounding environment, and the 3D positions of the animal limbs. By trajectory, we refer to the 3D location over time of the root joint of the animal's skeleton. The second stage takes the generated trajectory and the initial pose (\ie the rotation of each joint of the animal's skeleton), and trains another VAE to generate, from a pose latent code, the poses of the animal over time. 

The second class of methods aligns the query (\eg environmental parameters or text prompts) with the 3D motion~\cite{yang2024omnimotiongpt,wang2025animo}. The key idea is to learn a joint latent space such that a 3D motion of an animal and any text prompts that correspond to the same 3D motion map to the same point in the latent space. During training, a motion autoencoder is trained to map a 3D motion of an animal into a latent code, which is then decoded, using a decoder, into the same 3D motion. At the same time, another branch takes text prompts and encodes them using a CLIP text encoder~\cite{radford2021learning}. The network is trained so that the text encoder and the 3D motion encoder are aligned, \ie the text prompts that describe the same 3D motion are encoded into the same latent code as the 3D motion itself.

Animo~\cite{wang2025animo} and OmnimotionGPT~\cite{yang2024omnimotiongpt} follow this pipeline but introduce several modifications to address challenges related to animals. Animo~\cite{wang2025animo} encodes the 3D motion of an animal in two stages. The first stage is a spatial transformer that encodes individual poses. The second stage is a motion transformer that takes the spatially encoded poses and generates a temporal encoding. To handle the large motion variability across species, the spatiotemporal encoding is further enhanced using a species-aware feature modulation module to generate motion tokens. A textual prompt, augmented with the species type and animal attributes, is encoded using CLIP followed by a residual transformer to generate motion tokens. These are then decoded into animal motion using a decoder.

OmnimotionGPT~\cite{yang2024omnimotiongpt} addressed the issue of lack of training data (\ie text prompts paired with ground-truth animal motion). The method aims to transfer prior knowledge learned from human data to the animal domain. The key idea is to jointly train two motion autoencoders, one for human motion (trained on abundant human motion data) and another for animal motion (trained on a limited animal dataset) while optimizing the similarity score among human motion encoding, animal motion encoding, and text CLIP embedding. To generate new animal motion by leveraging motion data, the method takes a text prompt and encodes it using CLIP to an animal latent code. At the same time, it replaces the subject of the query with a person and uses it to generate a human motion, which is then encoded into a human motion latent code. The latter as well as the animal code are concatenated and fed to a decoder that maps it to an animal motion code, which is then decoded using the animal motion decoder into an animal motion.  

\section{Training strategies and datasets} 
\label{Sec: Training Strategies}

This section summarizes the different datasets used for training and evaluating 3D animal reconstruction methods (\Cref{sec:datasets}). It also discusses the different levels of supervision commonly used in 3D animal reconstruction, namely 3D supervision (\Cref{sec:3Dsuperivsion}), self-supervision (\Cref{sec:selfsuperivsion}), and the regularization techniques used to impose some constraints on the 3D reconstruction (\Cref{sec:regulariation}).

\subsection{Datasets}
\label{sec:datasets}
Datasets are the core of deep learning-based 3D reconstruction. \Cref{tab:datasets} provides a comprehensive list of different datasets used for training and evaluating the performance of animal reconstruction methods. We tabulate existing datasets based on \textbf{(1)} the number of classes (\eg species types and breeds) and instances they include, \textbf{(2)} whether they are real or synthesized data, and \textbf{(3)} the type of modalities (\eg images or videos). We also include how the groundtruth 3D models are represented, noting one can use existing and well-established techniques to convert the data into different modalities. For example, polygonal meshes can be easily converted into point clouds or volumetric representations using discretization algorithms. Volumetric representations can also be converted to polygonal meshes using marching cubes algorithm. \Cref{tab:datasets} also discusses whether the datasets include additional annotations such as 2D/3D keypoints, camera poses, etc. 

Several key limitations can be observed from~\Cref{tab:datasets}. \First, real 3D animal data is scarce, most datasets with 3D ground truth rely on synthetic data or scanned statuettes rather than live animals. \Second, multi-view datasets are limited, particularly for real animals in natural environments. \Third, temporal consistency annotations (dynamic 4D data) are rare. Finally, cross-species coverage is unbalanced, while dogs and cats are well-represented, many species have minimal or no representation.

\begin{table*}[t]
\caption{\label{tab:datasets}Overview of various datasets for 3D animal reconstruction ordered chronologically.}
\centering

\setlength{\tabcolsep}{3pt}
 \resizebox{\textwidth}{!}{
\begin{tabular}{@{}p{3cm}p{1.5cm}p{1cm}*{5}{c}*{5}{c}*{8}{c}@{}}
\toprule
\multirow{2}{*}{\textbf{Dataset}} & \multirow{2}{1.5cm}{\centering \textbf{\#Species}\\\textbf{/Breeds}} & \multirow{2}{*}{\textbf{\#Inst.}} & \multirow{2}{*}{\textbf{Type}} & \multicolumn{2}{c}{\textbf{Images}} & \multicolumn{2}{c}{\textbf{Videos}} & \multicolumn{5}{c}{\textbf{3D Representation}} & \multicolumn{8}{c}{\textbf{Additional Annotations}} \\
\cmidrule(lr){5-6} \cmidrule(lr){7-8} \cmidrule(lr){9-13} \cmidrule(lr){14-21}
& & & & \textbf{Mono} & \textbf{Multi} & \textbf{Mono} & \textbf{Multi} & \textbf{Dyn} & \textbf{Mesh} & \textbf{Param} & \textbf{P.C.} & \textbf{Voxel} & \textbf{Key} & \textbf{Mask} & \textbf{Depth} & \textbf{Pose} & \textbf{Cam} & \textbf{Flow} & \textbf{Skel} & \textbf{Text} \\
\midrule
    
\textbf{TOSCA}~\cite{tosca2008} & 9 & 80 & \synthdata\animalhuman & \xmarkred & \xmarkred & \xmarkred & \xmarkred & \xmarkred & \cmarkgreen & \xmarkred & \xmarkred & \xmarkred & \xmarkred & \xmarkred & \xmarkred & \xmarkred & \xmarkred & \xmarkred & \xmarkred & \xmarkred \\


\textbf{TigDog}~\cite{tigdog2016} & 3 & 110k & \realdata\animalonly & \xmarkred & \xmarkred & \cmarkgreen & \xmarkred & \xmarkred & \xmarkred & \xmarkred & \xmarkred & \xmarkred & \cmarkgreen & \xmarkred & \xmarkred & \xmarkred & \xmarkred & \xmarkred & \xmarkred & \xmarkred \\

\textbf{SMAL}~\cite{zuffi2017menagerie} & 5 & 49 & \realdata\animalonly & \xmarkred & \xmarkred & \xmarkred & \xmarkred & \xmarkred & \cmarkgreen & \cmarkgreen & \xmarkred & \xmarkred & \xmarkred & \xmarkred & \xmarkred & \xmarkred & \xmarkred & \xmarkred & \xmarkred & \xmarkred \\

\textbf{SMALR}~\cite{Zuffi2018LionsTigersBears} & 5 & 7{,}588 & \realdata\animalonly & \xmarkred & \cmarkgreen & \xmarkred & \cmarkgreen & \cmarkgreen & \cmarkgreen & \cmarkgreen & \xmarkred & \xmarkred & \cmarkgreen & \cmarkgreen & \xmarkred & \xmarkred & \cmarkgreen & \xmarkred & \xmarkred & \xmarkred \\

\textbf{Animal\text{-}Pose}~\cite{animalpose2019} & 5 & 6{,}000 & \realdata\animalonly & \cmarkgreen & \xmarkred & \xmarkred & \xmarkred & \xmarkred & \xmarkred & \xmarkred & \xmarkred & \xmarkred & \cmarkgreen & \xmarkred & \xmarkred & \xmarkred & \xmarkred & \xmarkred & \xmarkred & \xmarkred \\

\textbf{Penn Aviary}~\cite{badger2020bird} & birds & 6{,}300 & \realdata\animalonly & \xmarkred & \cmarkgreen & \xmarkred & \xmarkred & \xmarkred & \cmarkgreen & \xmarkred & \xmarkred & \xmarkred & \cmarkgreen & \cmarkgreen & \xmarkred & \xmarkred & \cmarkgreen & \xmarkred & \xmarkred & \xmarkred \\

\textbf{StanfordExtra}~\cite{biggs2020who} & dogs/120 & 12{,}000 & \realdata\animalonly & \cmarkgreen & \xmarkred & \xmarkred & \xmarkred & \xmarkred & \xmarkred & \xmarkred & \xmarkred & \xmarkred & \cmarkgreen & \cmarkgreen & \xmarkred & \xmarkred & \xmarkred & \xmarkred & \xmarkred & \xmarkred \\

\textbf{RGBD\text{-}Dog}~\cite{rgbddog2020} & dogs/5 & 25 & \realdata\animalonly & \cmarkgreen & \xmarkred & \cmarkgreen & \xmarkred & \cmarkgreen & \cmarkgreen & \xmarkred & \cmarkgreen & \xmarkred & \xmarkred & \xmarkred & \cmarkgreen & \cmarkgreen & \cmarkgreen & \xmarkred & \cmarkgreen & \xmarkred \\

\textbf{AP\text{-}10K}~\cite{ap10k2021} & 54 & 10{,}015 & \realdata\animalonly & \cmarkgreen & \xmarkred & \xmarkred & \xmarkred & \xmarkred & \xmarkred & \xmarkred & \xmarkred & \xmarkred & \cmarkgreen & \xmarkred & \xmarkred & \xmarkred & \xmarkred & \xmarkred & \xmarkred & \xmarkred \\

\textbf{Horse\text{-}30}~\cite{horse30_2021} & horses & 8{,}114 & \realdata\animalonly & \cmarkgreen & \xmarkred & \cmarkgreen & \xmarkred & \xmarkred & \xmarkred & \xmarkred & \xmarkred & \xmarkred & \cmarkgreen & \xmarkred & \xmarkred & \xmarkred & \xmarkred & \xmarkred & \xmarkred & \xmarkred \\

\textbf{SyDog}~\cite{sydog2021} & dogs & 32k & \synthdata\animalonly & \cmarkgreen & \xmarkred & \xmarkred & \xmarkred & \xmarkred & \xmarkred & \xmarkred & \xmarkred & \xmarkred & \cmarkgreen & \cmarkgreen & \xmarkred & \xmarkred & \xmarkred & \xmarkred & \xmarkred & \xmarkred \\

\textbf{AcinoSet}~\cite{acinoset2021} & cheetahs & 8{,}522 & \realdata\animalonly & \xmarkred & \cmarkgreen & \xmarkred & \xmarkred & \cmarkgreen & \xmarkred & \xmarkred & \xmarkred & \xmarkred & \cmarkgreen & \cmarkgreen & \xmarkred & \cmarkgreen & \cmarkgreen & \xmarkred & \xmarkred & \xmarkred \\

\textbf{DT4D}~\cite{li2021complete} & 31 & 1{,}972 & \synthdata\animalhuman & \xmarkred & \xmarkred & \xmarkred & \xmarkred & \cmarkgreen & \cmarkgreen & \xmarkred & \xmarkred & \cmarkgreen & \xmarkred & \xmarkred & \xmarkred & \xmarkred & \xmarkred & \xmarkred & \xmarkred & \xmarkred \\

\textbf{APT\text{-}36K}~\cite{apt36k2022} & 30 & 36{,}000 & \realdata\animalonly & \xmarkred & \xmarkred & \cmarkgreen & \xmarkred & \cmarkgreen & \xmarkred & \xmarkred & \xmarkred & \xmarkred & \cmarkgreen & \xmarkred & \xmarkred & \xmarkred & \xmarkred & \xmarkred & \xmarkred & \xmarkred \\

\textbf{Animal Kingdom}~\cite{ng2022animal} & 850 & 33k & \realdata\animalonly & \xmarkred & \xmarkred & \cmarkgreen & \xmarkred & \xmarkred & \xmarkred & \xmarkred & \xmarkred & \xmarkred & \cmarkgreen & \xmarkred & \xmarkred & \xmarkred & \xmarkred & \xmarkred & \xmarkred & \xmarkred \\

\textbf{DFA}~\cite{luo2022artemis} & 9 & 122k & \synthdata\animalonly & \xmarkred & \xmarkred & \xmarkred & \cmarkgreen & \cmarkgreen & \xmarkred & \xmarkred & \xmarkred & \cmarkgreen & \xmarkred & \cmarkgreen & \xmarkred & \xmarkred & \cmarkgreen & \xmarkred & \cmarkgreen & \xmarkred \\

\textbf{Animal3D}~\cite{Xu2023Animal3DAC} & 40 & 3{,}400 & \realdata\animalonly & \cmarkgreen & \xmarkred & \xmarkred & \xmarkred & \xmarkred & \cmarkgreen & \cmarkgreen & \xmarkred & \xmarkred & \cmarkgreen & \cmarkgreen & \xmarkred & \cmarkgreen & \xmarkred & \xmarkred & \xmarkred & \xmarkred \\

\textbf{CoP3D}~\cite{sinha2023common} & 2 & 600k & \realdata\animalonly & \xmarkred & \xmarkred & \cmarkgreen & \xmarkred & \xmarkred & \xmarkred & \xmarkred & \xmarkred & \xmarkred & \xmarkred & \cmarkgreen & \xmarkred & \xmarkred & \cmarkgreen & \xmarkred & \xmarkred & \xmarkred \\

\textbf{MV\text{-}SyDog}~\cite{mvsydog2023} & dogs & 1k & \synthdata\animalonly & \xmarkred & \xmarkred & \xmarkred & \cmarkgreen & \cmarkgreen & \xmarkred & \xmarkred & \xmarkred & \xmarkred & \cmarkgreen & \xmarkred & \cmarkgreen & \cmarkgreen & \cmarkgreen & \xmarkred & \cmarkgreen & \xmarkred \\

\textbf{SyDog\text{-}Video}~\cite{sydogvideo2023} & dogs & 87{,}500 & \synthdata\animalonly & \xmarkred & \xmarkred & \cmarkgreen & \xmarkred & \xmarkred & \xmarkred & \xmarkred & \xmarkred & \xmarkred & \cmarkgreen & \cmarkgreen & \xmarkred & \xmarkred & \xmarkred & \xmarkred & \xmarkred & \xmarkred \\

\textbf{DigitalLife3D}~\cite{digitallife3d} & 20+ & 163 & \realdata\animalonly & \xmarkred & \xmarkred & \xmarkred & \xmarkred & \xmarkred & \cmarkgreen & \xmarkred & \xmarkred & \xmarkred & \xmarkred & \xmarkred & \xmarkred & \xmarkred & \xmarkred & \xmarkred & \xmarkred & \xmarkred \\


\textbf{D\text{-}SMAL}~\cite{ruegg2023bite} & dogs & 65 & \realdata\animalonly & \cmarkgreen & \xmarkred & \xmarkred & \xmarkred & \xmarkred & \cmarkgreen & \cmarkgreen & \xmarkred & \xmarkred & \xmarkred & \xmarkred & \xmarkred & \xmarkred & \xmarkred & \xmarkred & \xmarkred & \xmarkred \\

\textbf{3D\text{-}Fauna}~\cite{li2024learning} & 128 & 78k & \realdata\animalonly & \cmarkgreen & \xmarkred & \xmarkred & \xmarkred & \xmarkred & \xmarkred & \xmarkred & \xmarkred & \xmarkred & \xmarkred & \cmarkgreen & \xmarkred & \xmarkred & \xmarkred & \xmarkred & \xmarkred & \xmarkred \\
 
\textbf{AnimalML3D}~\cite{yang2024omnimotiongpt} & 36 & 1{,}240 & \synthdata\animalonly & \xmarkred & \xmarkred & \xmarkred & \xmarkred & \cmarkgreen & \cmarkgreen & \cmarkgreen & \xmarkred & \xmarkred & \xmarkred & \xmarkred & \xmarkred & \cmarkgreen & \xmarkred & \xmarkred & \cmarkgreen & \cmarkgreen \\


\textbf{DigiDogs}~\cite{digidogs2024} & dogs & 27{,}900 & \synthdata\animalonly & \xmarkred & \xmarkred & \cmarkgreen & \xmarkred & \cmarkgreen & \cmarkgreen & \cmarkgreen & \xmarkred & \xmarkred & \cmarkgreen & \cmarkgreen & \cmarkgreen & \cmarkgreen & \cmarkgreen & \xmarkred & \cmarkgreen & \xmarkred \\

\textbf{AniMo4D}~\cite{wang2025animo} & 114 & 78{,}149 & \synthdata\animalonly & \xmarkred & \xmarkred & \xmarkred & \xmarkred & \cmarkgreen & \cmarkgreen & \xmarkred & \xmarkred & \xmarkred & \xmarkred & \xmarkred & \xmarkred & \cmarkgreen & \xmarkred & \xmarkred & \cmarkgreen & \cmarkgreen \\

\textbf{CtrlAni3D}~\cite{lyu2025animer} & 10 & 9{,}700 & \synthdata\animalonly & \cmarkgreen & \xmarkred & \xmarkred & \xmarkred & \xmarkred & \cmarkgreen & \cmarkgreen & \xmarkred & \xmarkred & \cmarkgreen & \cmarkgreen & \xmarkred & \xmarkred & \xmarkred & \xmarkred & \xmarkred & \xmarkred \\

\bottomrule
\end{tabular}
}
\vspace{2pt}
\textit{Symbols: \realdata/\synthdata~indicate real/synthetic data; \animalonly/\animalhuman~indicate animal-only/animal\&human. Abbreviations: 3D Representation columns: Dyn=Dynamic/4D, Mesh=Triangular meshes, Param=Parametric models, P.C.=Point clouds, Voxel=Voxel grids. Additional Annotations: Key=Keypoints, Mask=Segmentation masks, Depth=Depth maps, Pose=3D pose, Cam=Camera parameters, Flow=Optical flow, Skel=Skeleton annotations, Text=Text descriptions.}
\end{table*}

\subsection{3D Supervision}
\label{sec:3Dsuperivsion}
During training, supervised techniques require 3D ground-truth, which can be paired or unpaired with the 2D images. While they achieve high accuracy,    obtaining high-quality 3D annotations for animals is inherently challenging. A practical approach to acquiring 3D geometric annotations is through the use of synthetic data and data augmentation techniques. For example, Zuffi \etal~\cite{Zuffi2019Safari} utilized the SMALR model to render a set of $57$ zebra images, generating $10$ different individual zebra models.

The loss functions for 3D supervision are designed to minimize discrepancies between predicted and ground-truth 3D shapes, using differentiable metrics to facilitate gradient-based optimization. In the following formulations, we denote by $\distance(\cdot, \cdot)$ a general distance metric, such as the L1 or L2 norm.

In general, when using explicit representations, the geometric loss function $\loss_{\text{geom}}$ measures the error between predicted and ground-truth 3D positions:
\begin{equation}
\loss_{\text{geom}} = \frac{1}{\nfeature_\point} \sum_{i=1}^{\nfeature_\point} \distance(\point_i, \predpoint_i),
\end{equation}

\noi where $\point_i$ and $\predpoint_i$ are the ground-truth and predicted 3D points, respectively, and $\nfeature_\point$ is the total number of vertices. Commonly used losses include the $\lone$ and $\ltwo$ metrics, and the  Chamfer Distance (CD). The latter compares two point sets by measuring the average closest-point distances between them. It is defined as:
\begin{equation}
    \begin{split}
        \loss_{\text{Cham}} = \frac{1}{|\pointset|} \sum_{\point \in \pointset} \min_{\predpoint \in \predpointset} \distance(\point, \predpoint) 
         + \frac{1}{|\predpointset|} \sum_{\predpoint \in \predpointset} \min_{\point \in \pointset} \distance(\predpoint, \point),
    \end{split}
\end{equation}

\noi where $\pointset$ and $\predpointset$ are the ground-truth and predicted point sets, and $|\pointset|$ and $|\predpointset|$ denote their sizes. This loss ensures that each point in one set is close to some point in the other set, enhancing overall shape similarity.

For implicit representations, loss functions measure the discrepancy between the predicted and ground-truth volumetric values, \eg the SDF, TSDF or occupancy values or their logarithmic transform~\cite{li2021complete} ,  over the 3D space. In the case of binary occupancy grids, the loss function $\loss_{\text{occ}}$ is often defined using binary cross-entropy between the predicted occupancy probabilities and the ground truth~\cite{li2021complete, lei2022cadex}.

\subsection{Weakly supervised and self-supervised Learning} 
\label{sec:selfsuperivsion}

Weakly supervised techniques use sparse 3D annotations,  such as 3D keypoints, that are easy to acquire. Self-supervised learning, on the other hand, uses 2D cues such as keypoints, silhouettes (foreground masks $\mask$), and even images themselves as supervisory signals. Compared to full 3D supervision, these cues can be obtained either through manual annotations or automated detection methods. Many existing methods guide 3D reconstruction using manual annotations in the form of 2D keypoints~\cite{Kanazawa2015Learning3D}, 2D keypoints and silhouettes~\cite{zuffi2017menagerie}, and/or prior knowledge in the form of statistical shape models~\cite{Zuffi2019Safari}. The latter removes the reliance on 2D annotations, but the statistical shape models need to be learned from 3D exemplars in an offline training phase.  3D models of animals are, however,  challenging to acquire. Zuffi \etal~\cite{Zuffi2019Safari} overcome this issue by using scans of animal statuettes. Thus, the statistical models learned from such data do not capture the rich inter-class and intra-class variability in shape,  pose, and motion of animals in their natural environments. 


The key idea is that instead of measuring the loss in terms of discrepancy between the reconstructed 3D model and the groundtruth 3D model, one can project, during training, the predicted 3D model onto images of the same scene but captured from different viewing angles, and then use the discrepancy between the projections and the 2D observations as a training loss. The projection can be implemented using either perspective~\cite{Zuffi2018LionsTigersBears, Zuffi2019Safari, badger2020bird, biggs2020who, ruegg2022barc} or weak-perspective~\cite{Kanazawa2018CategorySpecific, tulsiani2020implicit, li2020online} methods.

A commonly used loss   is the \textbf{keypoint reprojection loss}, which measures the discrepancy between the projections of  predicted 3D keypoints and  ground-truth 2D keypoints:
\begin{equation}
\loss_{\text{kp}} = \frac{1}{\nfeature_\text{kp}} \sum_{j=1}^{\nfeature_\text{kp}} \distance\left( x_j, \project\left( \point_j; \cameraparams \right) \right),
\end{equation}

\noi where $x_j$ is the ground-truth 2D keypoint, $\point_j$ is the corresponding 3D keypoint in the predicted model, $\project$ is the projection function parameterized by camera parameters $\cameraparams$, and $\nfeature_\text{kp}$ is the total number of keypoints. Minimizing this loss encourages the projected 3D keypoints to align with the observed 2D keypoints.

In addition to keypoints, \textbf{silhouettes}, represented as foreground masks $\mask$, provide valuable information for 2D supervision. The silhouette reprojection loss evaluates how closely the silhouette of the projected 3D model matches the observed 2D silhouette. A commonly used silhouette loss is the Intersection-over-Union (IoU) between the predicted mask $\hat{\mask}$ and the ground-truth mask $\mask$:
\begin{equation}
\loss_{\text{silh}} = 1 - \frac{|\mask \cap \hat{\mask}|}{|\mask \cup \hat{\mask}|},
\end{equation}

\noi where $|\cdot|$ denotes the number of pixels in the set. Minimizing this loss encourages the predicted silhouette to closely match the ground truth in the image.

Alternatively, the silhouette reprojection loss can be formulated using distance transforms, which measure the proximity between the predicted and ground-truth silhouette boundaries:
\begin{equation}
    \begin{split}
        \loss_{\text{silh}} =  \frac{1}{|\mask|} \sum_{\pixel \in \mask} D_{\hat{\mask}}(\pixel) 
         + \frac{1}{|\hat{\mask}|} \sum_{\pixel' \in \hat{\mask}} D_{\mask}(\pixel').
    \end{split}
\end{equation}

\noi Here, $D_{\hat{\mask}}(\pixel)$ is the distance from pixel $\pixel \in \mask$ to the boundary of $\hat{\mask}$, and $D_{\mask}(\pixel')$ is the distance from pixel $\pixel' \in \hat{\mask}$ to the boundary of $\mask$. This loss ensures that the predicted silhouette contour aligns closely with the ground-truth contour.

\textbf{Perceptual loss}~\cite{Zuffi2019Safari, yang2021lasr, aygun2024saor, sinha2023common} measures the similarity between the predicted and reference images based on feature representations extracted from a pre-trained network:
\begin{equation}
\loss_{\text{perc}} = \frac{1}{\nfeature} \sum_{\featurevector} \distance\left( F_{\params}(I_{\text{ref}})(\featurevector), F_{\params}(\tilde{I}_{\text{ref}})(\featurevector) \right).
\end{equation}

\noi Here, $F_{\params}$ denotes the feature extraction function from a pre-trained network (\eg VGG or ResNet), $I_{\text{ref}}$ is the reference image, $\tilde{I}_{\text{ref}}$ is the reconstructed image, and the sum is over all features $\featurevector$. By comparing feature maps, this loss is robust to appearance variations. It also captures high-level perceptual differences not evident in pixel-level comparisons.

\textbf{Semantic consistency loss}~\cite{yao2021lassie, yao2023artic3d, yao2023hi-lassie} aims to align semantic features between the predicted 3D model and the reference representations. It is formulated as:
\begin{equation}
    \begin{split}
        \loss_{\text{sem}} = \sum_{j} \bigg( & \frac{1}{|\pixels_j|} \sum_{\pixel \in \pixels_j} \min_{\point \in \vertices_j} \distance\left( \pixel, \project_j(\point) \right) \\
        & + \frac{1}{|\vertices_j|} \sum_{\point \in \vertices_j} \min_{\pixel \in \pixels_j} \distance\left( \project_j(\point), \pixel \right) \bigg),
    \end{split}
\end{equation}

\noi where $\pixels_j$ is the set of pixels in the 2D image for instance $j$, $\vertices_j$ is the set of corresponding points on the 3D model, and $\project_j(\vertex)$ denotes the projection of 3D point $\point$ onto the 2D plane. Minimizing this loss encourages spatial alignment between the projected 3D model and the 2D image.

To incorporate semantic features, the distance $\distance(\pixel, \project_j(\point))$ can be augmented with geometric and semantic alignment:
\begin{equation}
\distance\left( \pixel, \project_j(\point) \right) = \distance\left( \pixel, \project_j(\point) \right) + \alpha \distance\left( K_j(\pixel), Q(\point) \right),
\end{equation}

\noi where $K_j(\pixel)$ represents the semantic features at pixel $\pixel$ in the 2D image, $Q(\point)$ denotes the semantic features at vertex $\point$ in the 3D model, and $\alpha$ is a weighting factor that balances the contributions of geometric and semantic distances.

\subsection{Regularization} 
\label{sec:regulariation}

Regularization techniques play a crucial role in enhancing the quality of 3D reconstructions by ensuring that the models are geometrically coherent and visually plausible. 

Examples of regularizers include \textbf{As-Rigid-As-Possible loss}~\cite{sorkine2007rigid,Kanazawa2015Learning3D, li2020online, yang2021lasr, wu2023dove}, which aims to preserve local rigidity during deformation by minimizing distortions in the reconstructed shape:
\begin{equation}
    \begin{split}
        \loss_{\text{ARAP}} = \frac{1}{\nfeature_\text{\point}} \sum_{i=1}^{\nfeature_\text{\point}} \sum_{j \in \neighborhood(i)} \distance\left( (\point_i' - \point_j'), \right. 
        \left. \rotation_i (\point_i - \point_j) \right),
    \end{split}
\end{equation}

\noi where $\nfeature_\text{\point}$ is the number of points, $\point_i$ and $\point_j$ are the positions of 3D points $i$ and $j$ in the original shape, $\point_i'$ and $\point_j'$ are the corresponding positions in the deformed shape, $\rotation_i$ is the optimal rotation matrix for vertex $i$, and $\neighborhood(i)$ denotes the set of neighboring vertices of vertex $i$. Minimizing this loss maintains local geometric relationships, ensuring transformations are as rigid as possible.

Another common regularizer is the \textbf{Laplacian smoothness loss}~\cite{yang2021lasr, yao2021lassie, yao2023artic3d, wu2023dove, yao2023hi-lassie}, which encourages smoothness by penalizing deviations of each vertex from the average position of its neighbors:
\begin{equation}
\loss_{\text{lap}} = \frac{1}{\nfeature_\text{\vertex}} \sum_{i=1}^{\nfeature_\text{\vertex}} \distance\left( \vertex_i', \frac{1}{|\neighborhood(i)|} \sum_{j \in \neighborhood(i)} \vertex_j' \right),
\end{equation}

\noi where $\vertex_i'$ is the $i$-th vertex in the deformed mesh, and $|\neighborhood(i)|$ is the number of its neighbors. This regularization reduces surface irregularities and enforces continuity across the mesh.

The \textbf{normal smoothness loss}~\cite{yao2023artic3d, jiang2023consistent4d, wu2023dove, yao2023hi-lassie, aygun2024saor} ensures that adjacent faces in the mesh align smoothly by minimizing the angular difference between their normals:
\begin{equation}
\loss_{\text{norm}} = \frac{1}{\nfeature} \sum_{(\face_i, \face_j) \in \faces} \distance\left( \normalvec_{\face_i}, \normalvec_{\face_j} \right),
\end{equation}

\noi where $\normalvec_{\face_i}$ and $\normalvec_{\face_j}$ are the normals of adjacent faces $\face_i$ and $\face_j$ that share an edge, $\faces$ is the set of all such face pairs, and $\nfeature$ represents the total number of these pairs. Here, $\distance(\cdot, \cdot)$ measures the angular difference between normals, and is computed using the cosine of the angle between the two normal vectors:
\begin{equation}
\distance\left( \normalvec_{\face_i}, \normalvec_{\face_j} \right) = 1 - \frac{\normalvec_{\face_i} \cdot \normalvec_{\face_j}}{\| \normalvec_{\face_i} \| \| \normalvec_{\face_j} \|}.
\end{equation}

\noi Minimizing this loss reduces sharp changes in the orientation of adjacent faces, resulting in  smooth reconstructions.

\section{Discussion} 
\label{sec:discussion}

\begin{table*}[t]
\caption{\label{Tab: Taxonomy and comparison}Taxonomy and comparison of 3D animal reconstruction methods.}
\centering
\setlength{\tabcolsep}{3pt}
 \resizebox{\textwidth}{!}{
\begin{tabular}{@{}p{2.8cm}p{1.0cm}p{3.0cm}p{3.0cm}p{3.0cm}p{2.5cm}p{2.7cm}@{}}
\toprule
\textbf{Method} & \textbf{Year} & \textbf{Input} & \textbf{Representation} & \textbf{Technique} & \textbf{Supervision Level} & \textbf{Notation} \\
\midrule
 Kanazawa \etal~\cite{Kanazawa2015Learning3D} & 2015 & 2D keypoints & Template & Opt.-based & N/A & - \\
 Zuffi \etal~\cite{zuffi2017menagerie} & 2017 & Parameters & SMAL & Opt.-based & N/A & - \\
 Zuffi \etal~\cite{Zuffi2018LionsTigersBears} & 2018 & 2D keypoints and mask & SMALR & Opt.-based & N/A & - \\

 Kanazawa \etal~\cite{Kanazawa2018CategorySpecific} & 2018 & 2D image & Template & Enc.-dec. & Weakly-supv. & 2D(kp+mask) \\
 Zuffi \etal~\cite{Zuffi2019Safari} & 2019 & 2D image & SMALST & Enc.-dec. & Supervised & 3D + 2D(kp, mask) \\
 
 Biggs \etal~\cite{biggs2019creatures} & 2018 & 2D image & SAMLify & Enc.-dec. & Weakly-supv. & 2D(kp+mask) \\
 Badger \etal~\cite{badger2020bird} & 2020 & 2D image & Template & Enc.-dec. & Weakly-supv. & 2D(kp+mask) \\
 Tulsiani \etal~\cite{tulsiani2020implicit} & 2020 & 2D image & Neural surface & Enc.-dec. & Weakly-supv. & 2D(mask) \\
 Biggs \etal~\cite{biggs2020who} & 2020 & 2D image & SMALD & Enc.-dec. & Weakly-supv. & 2D(kp+mask) \\
 Li \etal~\cite{li2020self} & 2020 & 2D image & Template & Enc.-dec. & Self-supv. & - \\
 Li \etal~\cite{li2020online} & 2020 & Monocular video & Template & Enc.-dec. & Weakly-supv. & 2D(kp+mask) \\
 Kulkarni \etal~\cite{kulkarni2020articulation} & 2020 & 2D image & Template & Enc.-dec. & Weakly-supv. & 2D(kp+mask) \\
 Yang \etal~\cite{yang2021lasr} & 2021 & Video sequence & Template & Opt.-based & N/A & 2D(kp+mask) \\
 Li \etal~\cite{li2021complete} & 2021 & partial TSDF and VMF & TSDF and VMF & Enc.-dec. & Supervised & 3D(TSDF, VMF) \\
 Yang \etal~\cite{yang2021viser} & 2021 & Monocular video & Triangle Mesh & Auto-dec. & Self-supv. & - \\
 Yao \etal~\cite{yao2021lassie} & 2021 & Sparse images & Neural surface & Enc.-dec. & Self-supv. & - \\
 Li \etal~\cite{li2021hsmal} & 2021 & Parameters & hSMAL & Opt.-based & N/A & - \\
 
 Lei \etal~\cite{lei2022cadex} & 2022 & Sparse point clouds & Occupancy network & Enc.-dec. & Supervised & 3D(occupancy field) \\
 
 Kokkinos \etal~\cite{kokkinos2022learning} & 2022 & Monocular video & Template & Enc.-dec. & Weakly-supv. & 2D(kp+mask) \\
 
 Yang \etal~\cite{yang2022banmo} & 2022 & Monocular videos & NeRF & Auto-dec. & Self-supv. & - \\
 
 Rüegg \etal~\cite{ruegg2022barc} & 2022 & 2D image & SMAL model for dogs & Enc.-dec. & Self-supv. & - \\
 
 Wu \etal~\cite{wu2022casa} & 2023 & Monocular video & Template & Enc.-dec. & Weakly-supv. & 2D(mask+optical flow) \\
 
 Wu \etal~\cite{wu2023magicpony} & 2023 & 2D image & SDF-mesh & Auto-dec. \& enc.-dec. & Weakly-supv. & 2D(mask) \\
 
 Sinha \etal~\cite{sinha2023common} & 2023 & multi-view video & NeRF & Auto-dec. & Self-supv. & - \\
 
 Liu \etal~\cite{liu2024one} & 2023 & 2D image & SDF & Auto-dec. & Self-supv. & - \\
 
 Yao \etal~\cite{yao2023artic3d} & 2023 & 2D images & Neural surface & Enc.-dec. & Self-supv. & - \\
 
 Yao \etal~\cite{yao2023hi-lassie} & 2023 & 2D images & Neural surface & Enc.-dec. & Self-supv. & - \\
 
 Yang \etal~\cite{yang2023Reconstructing} & 2023 & Monocular video & SDF & Auto-dec. & Self-supv. & - \\
 
 Jiang \etal~\cite{jiang2023consistent4d} & 2023 & Monocular video & NeRF & Auto-dec. & Self-supv. & - \\
 
 Wu \etal~\cite{wu2023dove} & 2023 & Monocular video & Template & Enc.-dec. & Self-supv. & - \\

  Cheng \etal~\cite{cheng2023virtual} & 2023 & Monocular video & NeRF & Auto-dec. \& enc.-dec. & Self-supv. & - \\
 
 Rüegg \etal~\cite{ruegg2023bite} & 2023 & 2D image & SMAL model for dogs & Enc.-dec. & Self-supv. & - \\
 
 Aygün \etal~\cite{aygun2024saor} & 2024 & 2D image & Template & Enc.-dec. & Self-supv. & - \\
 
 Zuffi \etal~\cite{zuffi2024varen} & 2024 & Parameters & Template & Opt.-based & N/A & - \\
 
 Li \etal~\cite{li2024learning} & 2024 & 2D image & SDF-mesh & Auto-dec. \& enc.-dec. & Self-supv. & - \\
 
 Tan \etal~\cite{tan2024distilling} & 2024 & Monocular video & Template & Enc.-dec. & Supervised & 3D(pseudo-gt) \\
 
 Zeng \etal~\cite{zeng2024stag4d} & 2024 & Monocular video & Gaussian splatting & 3DGS & N/A & - \\
 
 Lei \etal~\cite{lei2024gart} & 2024 & Monocular video & Gaussian splatting & 3DGS & Self-supv. & - \\
 
 Liu \etal~\cite{liu2024lepard} & 2024 & 2D images & Template & Enc.-dec. & Self-supv. & - \\
 
 Cho \etal~\cite{cho2025dogrecon} & 2025 & 2D image & Gaussian splatting & 3DGS & Self-supv. & - \\
 
 Zhai \etal~\cite{zhai2025taga} & 2025 & Monocular video & Gaussian splatting & 3DGS & Self-supv. & - \\
 
 Lyu \etal~\cite{lyu2025animer} & 2025 & 2D image & Template & Enc.-dec. & Self-supv. & - \\
 
 Kaye \etal~\cite{kaye2025dualpm} & 2025 & 2D images & PointMap & Enc.-dec. & Self-supv. & - \\
 
 Nizamani \etal~\cite{nizamani2025dynamic} & 2025 & 4D data & Neural surface & Enc.-dec. & Self-supv. & - \\
\bottomrule
\end{tabular}
}

\vspace{2pt}
\textit{Abbreviations: Opt.-based: Optimization-based. Enc.-dec.: Encoder-decoder. Auto-dec.: Auto-decoder. Self-supv.: Self-supervised. Weakly-supv.: Weakly-supervised. N/A: Not Applicable.}
\end{table*}

\begin{figure*}[t]
    \centering
    \includegraphics[width=\textwidth]{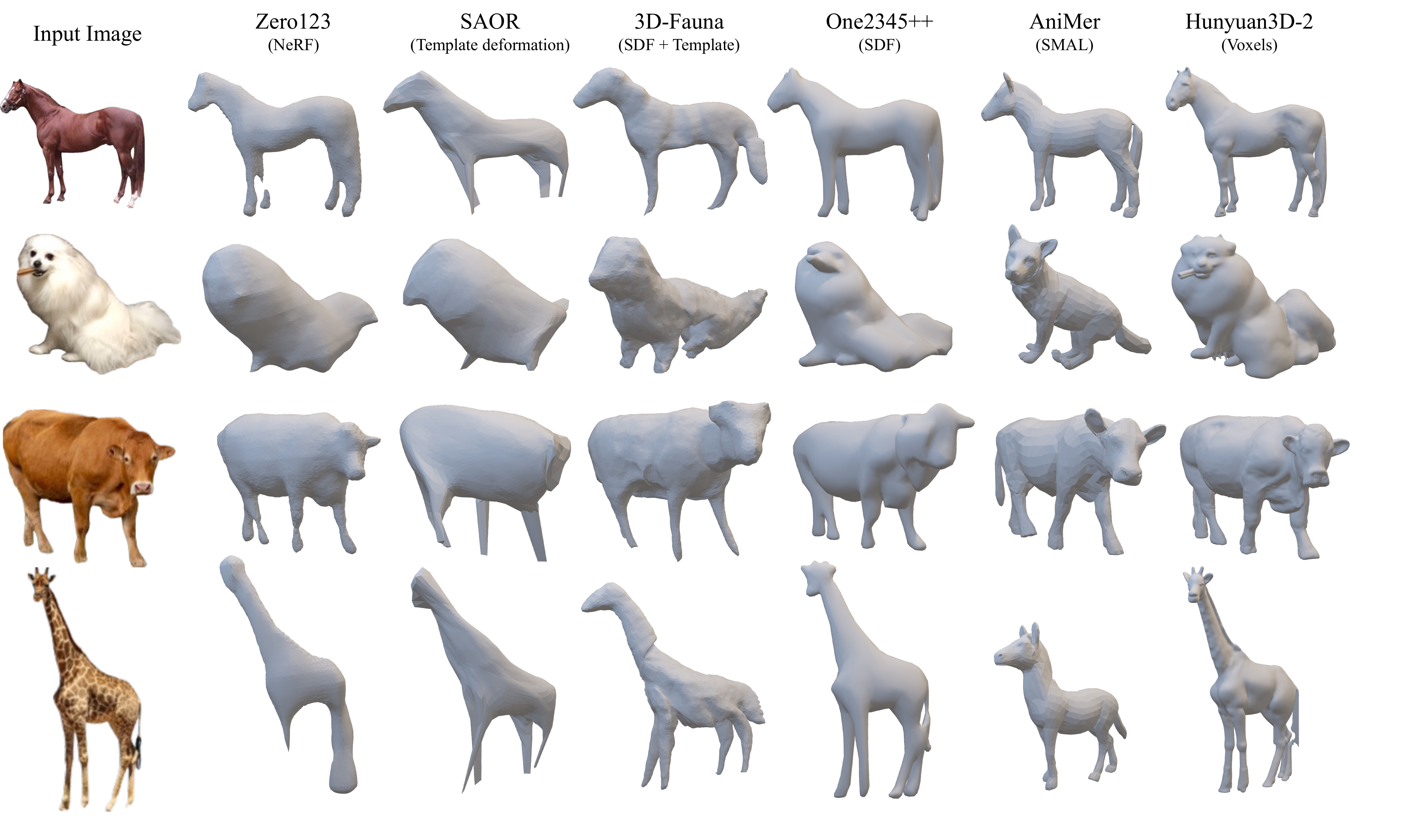}
    \caption{Qualitative comparison of geometric detail in 3D animal reconstruction, including Zero123~\cite{liu2023zero1to3} (2023), SAOR~\cite{aygun2024saor} (2024), 3D Fauna~\cite{li2024learning} (2024), One2345++~\cite{liu2024one2345} (2024), AniMer~\cite{lyu2025animer} (2025), and Hunyuan3D-2~\cite{hunyuan3d22025tencent} (2025).}
    \label{fig:comparison_reconstruction}
\end{figure*}

\begin{figure*}[t]
    \centering
    \includegraphics[width=\textwidth]{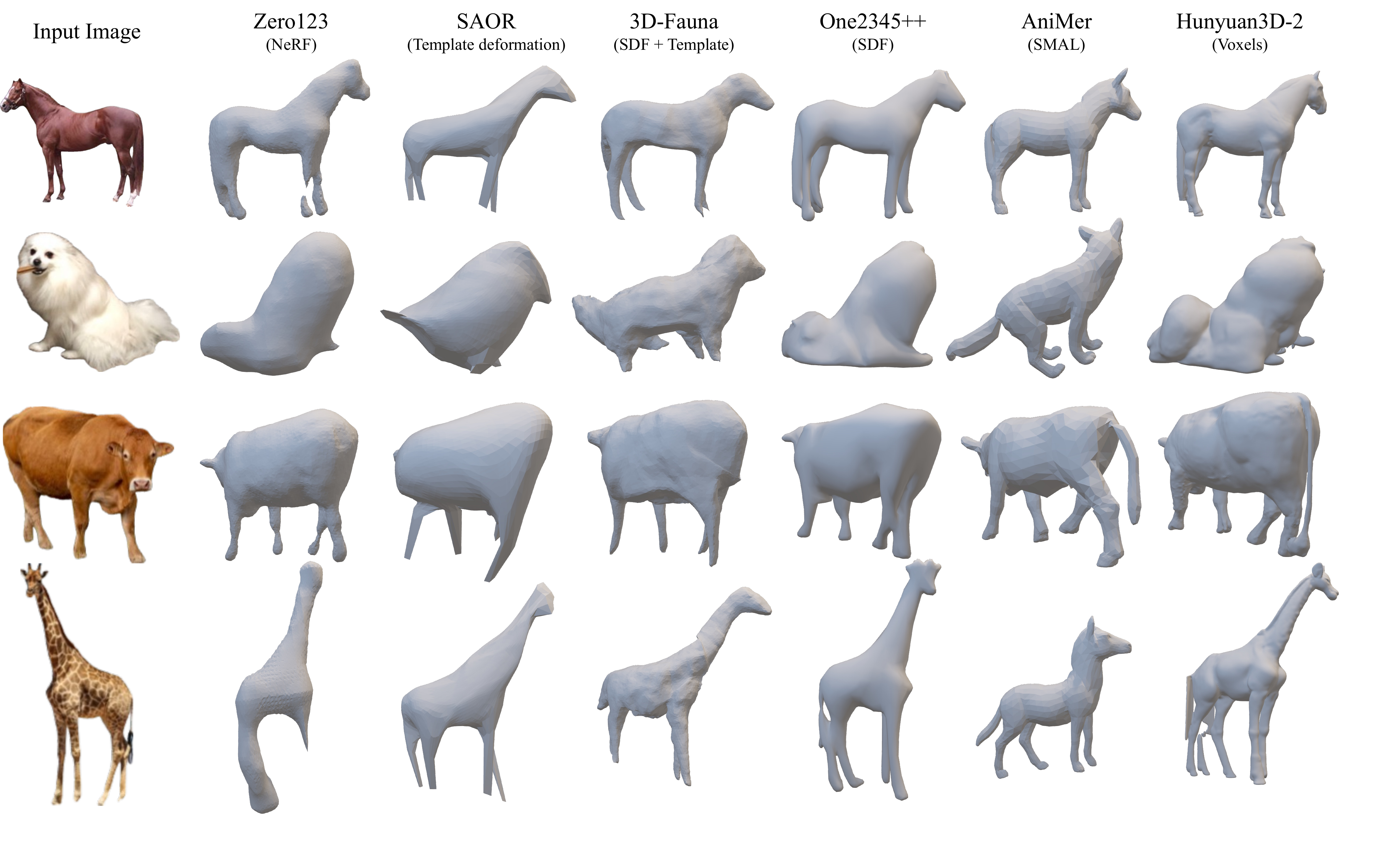}
    \caption{The same results as in~\Cref{fig:comparison_reconstruction}, rendered from a different viewpoint (back view).}
    \label{fig:comparison_reconstructionback}
\end{figure*}

\Cref{Tab: Taxonomy and comparison} provides a summary of the key methods covered in this survey, illustrating how the field of 3D reconstruction of animals has evolved over time. To perform a qualitative analysis, we selected a subset of representative methods and tested them on a variety of real-world scenarios. We evaluated their ability to be applied to monocular image reconstruction while covering a diverse range of species, with a particular emphasis on quadrupeds; see Figures~\ref{fig:comparison_reconstruction} and~\ref{fig:comparison_reconstructionback}.  We selected six state-of-the-art methods that span the different representations covered in this survey. 

Among the \textbf{template-based methods}, we selected 3D-Fauna~\cite{li2024learning}, SAOR~\cite{aygun2024saor}, and AniMer~\cite{lyu2025animer}:
\begin{itemize}
    \item 3D Fauna~\cite{li2024learning}  employs an explicit representation through a parametric model and is fast, with an average runtime of approximately $1.8$ seconds per image.  
    
    \item SAOR~\cite{aygun2024saor} uses explicit representation via template (sphere) deformation. It offers greater flexibility in handling various shapes but is slower as it requires around $7$ seconds per image. 
    
    \item  AniMer~\cite{lyu2025animer}, which uses  the SMAL model, is remarkably fast as it requires less than $3$ seconds per image.
\end{itemize}

\noi We have also evaluated three \textbf{template-free methods}, namely Zero123~\cite{liu2023zero1to3}, One2345++~\cite{liu2024one2345}, and Hunyuan3D-2~\cite{hunyuan3d22025tencent}:
\begin{itemize}
    \item  Zero123~\cite{liu2023zero1to3} uses an implicit representation based on NeRF, which is known for high fidelity but requires significantly higher computation time, averaging around $5$ minutes per image.

    \item  One2345++~\cite{liu2024one2345} also employs an implicit representation via SDF, with a particular focus on consistent multi-view generation to improve 3D reconstruction quality. 

    \item Hunyuan3D-2~\cite{hunyuan3d22025tencent} is a foundational model that uses volumetric representations. It is generic as it can be used for the 3D reconstruction of arbitrary 3D objects. Thus, it generalizes well to a wide range of animal species. In terms of computation time, it requires around $100$ seconds per image.
\end{itemize}

\noi Most methods (3D-fauna, SAOR, Zero123, AniMer and Hunyuan3D-2) were executed locally on a PC equipped with an NVIDIA GeForce RTX 3090 GPU, with 24 GB of VRAM and running Ubuntu $22.04$. However, One2345++ does not have publicly available source code, and thus its results were obtained via the official demonstration website. Consequently, the processing time for this method cannot be directly compared with the others, and thus, our evaluation of this method focused solely on the accuracy of its reconstruction results.

To ensure a fair comparison and minimize the influence of different segmentation methods on the reconstruction quality, all input images were preprocessed using the same segmentation approach to remove the background. This preprocessing step ensures that the observed differences in reconstruction results reflect the actual reconstruction capabilities of the methods rather than variations in image segmentation quality.

To evaluate  the performance of these methods, we selected four distinct real-world scenarios (see Figures~\ref{fig:comparison_reconstruction} and~\ref{fig:comparison_reconstructionback}):
\begin{itemize}
    \item The first scenario represents the ideal case, where the animal is in a standing pose, and the species is well-known, allowing for optimal conditions for reconstruction. 
    
    \item The second scenario focuses on animals with highly detailed surfaces, such as dogs that have regions with complex fur details, particularly the head and tail. 
    
    \item The third scenario involves multiple animals, specifically examining the challenges posed by overlapping head regions. 
    
    \item The last scenario tests the methods on animals that differ significantly from the common quadruped morphology, such as giraffes, to evaluate their generalization to species variation.
\end{itemize}

\noi The comparison reveals fundamental trade-offs between template-based and template-free approaches. Template-based methods such as AniMer, which rely on parametric models such as SMAL, are fast and produce anatomically plausible results since the deformation of the template is restricted to be within the space of plausible shapes.  These methods, however, fail to accurately capture instance-specific details. They also do not generalize to species that are not within the data used to learn the parametric model. As shown in the giraffe example of Figures~\ref{fig:comparison_reconstruction} and~\ref{fig:comparison_reconstructionback}, SMAL-based methods produce poor results for species outside their template coverage. Even for supported species, template-based methods cannot capture fine details such as the fur texture visible in the dog examples.

Template-free approaches such as Hunyuan3D-2 demonstrate opposite characteristics. While they are slower and occasionally produce anatomical artifacts such as extra limbs (as seen in the giraffe example), they excel at capturing fine-grained details and handling diverse animal morphologies. 

Unlike other methods, Zero123 and One2345++  take as input a monocular RGB image, generate additional images of the animal from different viewing angles, and perform the reconstruction from the generated multi-views. Thus, the quality of the reconstruction depends on the quality of the generated views. This is illustrated in  Figures~\ref{fig:comparison_reconstruction} and~\ref{fig:comparison_reconstructionback}, where we can see that Zero123 misses thin parts such as animal limbs. This is due to the fact that these parts are missing in the generated multi-views. This problem is addressed in  One2345++ by improving the quality of the synthesized views using diffusion models. This results in more complete reconstructions.


\section{Open challenges} 
\label{Sec:Research gaps}

Despite the significant advances that we have seen in the past five years,  there are several challenges that need further exploration. In this section, we provide an overview of typical challenges and highlight future research directions.

\vspace{6pt}
\noi\textbf{Reconstruction accuracy.} Deep learning models for the 3D reconstruction of animals have made notable progress in recent years. However, their reconstruction accuracy, particularly in terms of recovering complex surface details, is still inferior to their human and generic object counterpart. Most of the existing deep learning-based techniques produce globally consistent 3D models but lack detailed surface features. An example is shown in~\Cref{fig:comparison_reconstruction} where we can see that all existing methods fail to reconstruct the hairy dog.
This limitation is exacerbated when using templates or training data that do not capture the geometric details of real animals~\cite{zuffi2017menagerie, biggs2020who, li2021hsmal}. We hypothesize that this is due to the fact that  \textbf{(1)} current methods lack the representation capability and \textbf{(2)} self-supervising mechanisms, which use reprojection losses, are insufficient for learning detailed 3D cues.  Recent advances in foundation models such as Hunyuan3D-2~\cite{hunyuan3d22025tencent} have demonstrated promising results in capturing fine-grained details through their comprehensive training on large-scale, diverse datasets. However, their generic nature limits their ability to capture species-specific anatomical priors. These models could be leveraged to generate high-quality multi-view supervision or to synthesize detailed surface textures that current methods struggle to capture. Incorporating traditional multi-view stereo (MVS) methods, or lessons learned from these mathematically and geometrically motivated methods, into the learning-based methods can be one step forward for addressing the accuracy issue.    


\vspace{6pt}
\noi\textbf{Occlusions and multiple interactions.} Reconstructing animals in scenes involving multiple subjects or dynamic interactions presents another critical challenge. Existing methods often focus on single-animal scenarios and struggle with occlusions that occur when parts of an animal or multiple animals overlap or interact with each other~\cite{yang2021viser, yao2023hi-lassie}. Accurate 3D reconstruction in these contexts requires sophisticated algorithms capable of inferring hidden surfaces and maintaining temporal coherence in dynamic environments~\cite{jiang2023consistent4d, li2020online}. While deep learning-based methods excel in inferring hidden parts, they require large training data, which is difficult to acquire in the case of animals, especially when dealing with uncommon and rare ones. Additionally, methods designed for single animals fail to scale to multiple interacting subjects, highlighting the need for techniques that can manage complex occlusions and varying poses while preserving geometric detail. For example, as shown in~\Cref{fig:comparison_reconstruction}, overlapping cows in the image lead to inaccurate deformations in the head area of the models.

\vspace{6pt}
\noi\textbf{Generalization across species.} A significant challenge in 3D animal reconstruction is achieving generalization across various species. Many existing methods are tailored to specific species or breeds, using predefined models that do not easily adapt to different anatomical structures~\cite{yao2021lassie, yao2023hi-lassie, yao2023artic3d}. Approaches that fail to generalize effectively to a diverse range of species or that depend on breed-specific information limit their applicability and accuracy when applied to unseen species~\cite{ruegg2022barc, li2024learning}. Developing a unified, highly accurate method capable of learning skeletal and surface structures across diverse species would significantly benefit various fields of science, including wildlife monitoring and preservation.

\vspace{6pt}
\noi\textbf{Metric 3D reconstruction.} Incorporating real-world measurement units into 3D reconstruction models remains an underexplored area, yet it is crucial for applications requiring precise dimensions, such as biomechanical analysis or species-specific studies. Most current monocular image/video-based reconstruction techniques fail to recover real-world scales, limiting their applicability in contexts where accurate physical measurements are essential. Developing methods that inherently include or can easily adjust to real-world measurements would enhance the practical utility of 3D reconstruction~\cite{li2020self}.

\vspace{6pt}
\noi\textbf{Interdisciplinary collaborations.} Most existing datasets are collected without input from biologists, resulting in limited coverage of morphological variation and anatomical accuracy. Collaboration with biologists is essential to create comprehensive datasets that not only capture species diversity but also incorporate biological knowledge such as anatomical constraints. Such biologically-informed datasets would enable reconstruction models to achieve better accuracy and generalization across species, ensuring that methods are both technically robust and biologically meaningful.

\vspace{6pt}
\noi\textbf{Reconstructing animals and the wild.} Finally, reconstructing animals and their wild environments is crucial for understanding animal behavior and evolution, and for modeling biological processes. This, however, is very challenging due to the complexity and variability in the environments in which animals live. Some recent papers, \eg\cite{kulits2025reconstructing}, started to explore this problem, and we expect to see more research in the future that can reconstruct and model animals, their environments, and the way they interact with each other and with the environment.

\section{Conclusion} 
\label{sec:conclusion}
In recent years, the field of 3D animal shape, pose, and motion reconstruction has experienced significant growth, driven by advancements in deep learning techniques and their applications to non-rigid objects. This survey covers the main representations and approaches that appeared in the past five years. Specifically, we provide an overview of these methods based on their input modalities and their underlying representations such as point maps, neural surfaces, template deformation, neural SDF and NeRF, and 3DGS. We also compare these methods by analyzing their strengths and weaknesses. Finally, we summarize the open challenges aiming to inspire future research. While many previous surveys~\cite{han2019image, laga2020survey, Fahim2021SingleView3R, Jin20203DRU, Dalal2024GaussianS3} have focused on general 3D object reconstruction, they do not address the unique challenges inherent to 3D animals. This paper fills this gap by providing the first comprehensive review of the latest methods beyond parametric models, offering insights into the specific challenges of reconstructing dynamic and diverse animal shapes. It can serve as a reference for researchers and practitioners in this fast-evolving field.

\bibliographystyle{IEEEtran}
\bibliography{main}%
\end{document}

%% file: notations.tex
\newcommand{\cmarkgreen}{\textcolor{green}{\checkmark}}
\newcommand{\xmarkred}{\textcolor{red}{\ding{55}}}

\newcommand{\realdata}{\ding{108}}
\newcommand{\synthdata}{\ding{110}}

\newcommand{\animalonly}{\ding{72}}
\newcommand{\animalhuman}{\ding{119}}

\newcommand{\depthmap}{D}

\newcommand{\shape}{X}             		

\newcommand{\mean}[1]{\bar{#1}}

\newcommand{\point}{p}				
\newcommand{\pointmap}{\textbf{P}}

\newcommand{\predpoint}{\hat{\point}}
\newcommand{\meanpoint}{\mean{\point}}

\newcommand{\pixel}{x}				

\newcommand{\pixels}{\mathbf{\pixel}}
\newcommand{\meanpixel}{\mean{\pixel}}	

\newcommand{\vertex}{\textbf{v}}		
\newcommand{\vertices}{\mathcal{V}}	
\newcommand{\faces}{\mathcal{F}}		
\newcommand{\nfaces}{n}
\newcommand{\face}{\text{T}}
 
\newcommand{\pointset}{S}			
\newcommand{\predpointset}{\hat{\pointset}}

\newcommand{\neuralsurfacefunc}{f}  
  
\newcommand{\params}{\theta}			
\newcommand{\Params}{\Theta}			

\newcommand{\featurevector}{\textbf{x}}   
\newcommand{\latentcode}{\textbf{z} }

\newcommand{\featuredim}{c}			
\newcommand{\nfeature}{N}           

\newcommand{\objectivefunc}{\mathcal{L}}
\newcommand{\loss}{\objectivefunc}			
\newcommand{\weights}{W}				
\newcommand{\bias}{b}

\newcommand{\ie}{\emph{i.e., }}
\newcommand{\eg}{\emph{e.g., }}
\newcommand{\etal}{\emph{et al.}}

\newcommand{\noi}{\noindent}

\newcommand{\real}{\mathbb{R}}      	

\newcommand{\rthree}{\real^3}

\newcommand{\rplus}{\real^+}

\newcommand{\cameraparams}{\alpha}	

\newcommand{\eigenvector}{\Lambda}			

\newcommand{\gaussian}{g}
\newcommand{\gaussianscount}{N}

\newcommand{\rotation}{\textbf{R}}			

\newcommand{\rotations}{SO(3)}
\newcommand{\SEthree}{SE(3)}

\newcommand{\domain}{\mathcal{D}}		
\newcommand{\inputdomain}{\domain_{in}}
\newcommand{\outputdomain}{\domain_{out}}

\newcommand{\geometryspace}{\mathcal{G}} %
\newcommand{\appearancespace}{\mathcal{A}} %

\newcommand{\sphere}{\mathbb{S}}            
\newcommand{\stwo}{\sphere^2}				
\newcommand{\domainpoint}{s}             

\newcommand{\ltwo}{L_2}						
\newcommand{\lone}{L_1}						

\newcommand{\deformation}{\delta}			

\newcommand{\width}{W}						
\newcommand{\height}{H}						

\newcommand{\neighborhood}{\mathcal{N}}		

\newcommand{\distance}{d}

\newcommand{\threshold}{\epsilon}

\newcommand{\First}{\textbf{First}}
\newcommand{\Second}{\textbf{Second}}
\newcommand{\Third}{\textbf{Third}}

\newcommand{\first}{\textbf{first}~}
\newcommand{\second}{\textbf{second}~}

\newcommand{\meansurf}{\mean{\shape}}
\newcommand{\meanshape}{\meansurf}
\def\cov{K}                     
\newcommand{\nshapes}{n}
\newcommand{\scalingmatrix}{\text{S}}

\newcommand{\pose}{\theta}
\newcommand{\nsmplposes}{K}

\newcommand{\project}{\pi}

\newcommand{\opacity}{\sigma}
\newcommand{\thecolor}{c}

\newcommand{\npoints}{N}

\newcommand{\viewdir}{\textbf{d}}

\newcommand{\normalvec}{\textbf{n}}

\newcommand{\transformation}{T}

\newcommand{\thetime}{t}

\newcommand{\nvert}{n}

\newcommand{\mask}{\mathcal{M}}

\newcommand{\conditions}{\mathcal{C}}

%% file: main.bbl
\begin{thebibliography}{100}
\providecommand{\url}[1]{#1}
\csname url@samestyle\endcsname
\providecommand{\newblock}{\relax}
\providecommand{\bibinfo}[2]{#2}
\providecommand{\BIBentrySTDinterwordspacing}{\spaceskip=0pt\relax}
\providecommand{\BIBentryALTinterwordstretchfactor}{4}
\providecommand{\BIBentryALTinterwordspacing}{\spaceskip=\fontdimen2\font plus
\BIBentryALTinterwordstretchfactor\fontdimen3\font minus \fontdimen4\font\relax}
\providecommand{\BIBforeignlanguage}[2]{{%
\expandafter\ifx\csname l@#1\endcsname\relax
\typeout{** WARNING: IEEEtran.bst: No hyphenation pattern has been}%
\typeout{** loaded for the language `#1'. Using the pattern for}%
\typeout{** the default language instead.}%
\else
\language=\csname l@#1\endcsname
\fi
#2}}
\providecommand{\BIBdecl}{\relax}
\BIBdecl

\bibitem{sinha2023common}
S.~Sinha, R.~Shapovalov, J.~Reizenstein, I.~Rocco, N.~Neverova, A.~Vedaldi, and D.~Novotný, ``{Common Pets in 3D: Dynamic New-View Synthesis of Real-Life Deformable Categories},'' \emph{IEEE/CVF CVPR}, pp. 4881--4891, 2023.

\bibitem{Kanazawa2015Learning3D}
A.~Kanazawa, S.~Z. Kovalsky, R.~Basri, and D.~W. Jacobs, ``{Learning 3D Deformation of Animals from 2D Images},'' \emph{Computer Graphics Forum}, vol.~35, 2015.

\bibitem{yang2023Reconstructing}
G.~Yang, C.~Wang, D.~R. Narapureddy, and D.~Ramanan, ``{Reconstructing Animatable Categories from Videos},'' \emph{2023 IEEE/CVF CVPR}, pp. 16\,995--17\,005, 2023.

\bibitem{cheng2023virtual}
Y.-C. Cheng, C.~H. Lin, C.~Wang, Y.~Kant, S.~Tulyakov, A.~Schwing, L.~Gui, and H.-Y. Lee, ``{Virtual Pets: Animatable Animal Generation in 3D Scenes},'' \emph{ArXiv}, vol. abs/2312.14154, 2023.

\bibitem{li2024learning}
Z.~Li, D.~Litvak, R.~Li, Y.~Zhang, T.~Jakab, C.~Rupprecht, S.~Wu, A.~Vedaldi, and J.~Wu, ``{Learning the 3D Fauna of the Web},'' \emph{IEEE/CVF CVPR}, vol. abs/2401.02400, 2024.

\bibitem{yang2022banmo}
G.~Yang, M.~Vo, N.~Neverova, D.~Ramanan, A.~Vedaldi, and H.~Joo, ``{BANMo: Building Animatable 3D Neural Models from Many Casual Videos},'' \emph{IEEE/CVF CVPR}, pp. 2853--2863, 2022.

\bibitem{yang2021viser}
G.~Yang, D.~Sun, V.~Jampani, D.~Vlasic, F.~Cole, C.~Liu, and D.~Ramanan, ``{Viser: Video-specific surface embeddings for articulated 3D shape reconstruction},'' \emph{Advances in Neural Information Processing Systems}, vol.~34, pp. 19\,326--19\,338, 2021.

\bibitem{han2019image}
X.-F. Han, H.~Laga, and M.~Bennamoun, ``Image-based 3d object reconstruction: State-of-the-art and trends in the deep learning era,'' \emph{IEEE transactions on pattern analysis and machine intelligence}, vol.~43, no.~5, pp. 1578--1604, 2021.

\bibitem{laga2020survey}
H.~Laga, L.~V. Jospin, F.~Boussaid, and M.~Bennamoun, ``A survey on deep learning techniques for stereo-based depth estimation,'' \emph{IEEE transactions on pattern analysis and machine intelligence}, vol.~44, no.~4, pp. 1738--1764, 2022.

\bibitem{Fahim2021SingleView3R}
G.~Fahim, K.~Amin, and S.~Zarif, ``Single-view 3d reconstruction: A survey of deep learning methods,'' \emph{Comput. Graph.}, vol.~94, pp. 164--190, 2021.

\bibitem{Jin20203DRU}
Y.~Jin, D.~Jiang, and M.~Cai, ``3d reconstruction using deep learning: a survey,'' \emph{Commun. Inf. Syst.}, vol.~20, pp. 389--413, 2020.

\bibitem{Dalal2024GaussianS3}
A.~Dalal, D.~Hagen, K.~G. Robbersmyr, and K.~M. Knausg{\aa}rd, ``Gaussian splatting: 3d reconstruction and novel view synthesis, a review,'' \emph{ArXiv}, vol. abs/2405.03417, 2024.

\bibitem{Tretschk2022StateOT}
E.~Tretschk, N.~Kairanda, R.~MallikarjunB., R.~Dabral, A.~Kortylewski, B.~Egger, M.~Habermann, P.~Fua, C.~Theobalt, and V.~Golyanik, ``{State of the Art in Dense Monocular Non‐Rigid 3D Reconstruction},'' \emph{Computer Graphics Forum}, vol.~42, 2022.

\bibitem{yunus2024recent}
R.~Yunus, J.~E. Lenssen, M.~Niemeyer, Y.~Liao, C.~Rupprecht, C.~Theobalt, G.~Pons-Moll, J.-B. Huang, V.~Golyanik, and E.~Ilg, ``Recent trends in 3d reconstruction of general non-rigid scenes,'' \emph{Computer Graphics Forum}, p. e15062, 2024.

\bibitem{blanz1999morphable}
V.~Blanz and T.~Vetter, ``A morphable model for the synthesis of 3d faces,'' \emph{Proceedings of the 26th annual conference on Computer graphics and interactive techniques}, pp. 187--194, 1999.

\bibitem{egger20203dmorphable}
B.~Egger, W.~A. Smith, A.~Tewari, S.~Wuhrer, M.~Zollhoefer, T.~Beeler, F.~Bernard, T.~Bolkart, A.~Kortylewski, S.~Romdhani \emph{et~al.}, ``{3D morphable face models—past, present, and future},'' \emph{ACM TOG}, vol.~39, no.~5, pp. 1--38, 2020.

\bibitem{loper2015smpl}
M.~Loper, N.~Mahmood, J.~Romero, G.~Pons-Moll, and M.~J. Black, ``{SMPL}: A skinned multi-person linear model,'' \emph{ACM Trans. Graphics (Proc. SIGGRAPH Asia)}, vol.~34, no.~6, pp. 248:1--248:16, Oct. 2015.

\bibitem{yang2024depth}
L.~Yang, B.~Kang, Z.~Huang, X.~Xu, J.~Feng, and H.~Zhao, ``Depth anything: Unleashing the power of large-scale unlabeled data,'' \emph{ArXiv}, vol. abs/2401.10891, 2024.

\bibitem{yao2021lassie}
C.~Yao, W.-C. Hung, Y.~Li, M.~Rubinstein, M.~Yang, and V.~Jampani, ``{LASSIE: Learning Articulated Shapes from Sparse Image Ensemble via 3D Part Discovery},'' \emph{Advances in Neural Information Processing Systems}, vol. abs/2207.03434, 2022.

\bibitem{yao2023artic3d}
C.-H. Yao, A.~Raj, W.-C. Hung, M.~Rubinstein, Y.~Li, M.-H. Yang, and V.~Jampani, ``{Artic3D: Learning robust articulated 3Dshapes from noisy web image collections},'' \emph{Advances in Neural Information Processing Systems}, vol.~36, 2024.

\bibitem{yao2023hi-lassie}
C.~Yao, W.-C. Hung, Y.~Li, M.~Rubinstein, M.-H. Yang, and V.~Jampani, ``{Hi-LASSIE: High-Fidelity Articulated Shape and Skeleton Discovery from Sparse Image Ensemble},'' \emph{IEEE/CVF CVPR}, pp. 4853--4862, 2022.

\bibitem{liu2024one}
M.~Liu, C.~Xu, H.~Jin, L.~Chen, M.~Varma~T, Z.~Xu, and H.~Su, ``{One-2-3-45: Any single image to 3D mesh in 45 seconds without per-shape optimization},'' \emph{Advances in Neural Information Processing Systems}, vol.~36, 2024.

\bibitem{liu2023zero1to3}
R.~Liu, R.~Wu, B.~V. Hoorick, P.~Tokmakov, S.~Zakharov, and C.~Vondrick, ``{Zero-1-to-3: Zero-shot One Image to 3D Object},'' 2023.

\bibitem{liu2024one2345}
M.~Liu, R.~Shi, L.~Chen, Z.~Zhang, C.~Xu, X.~Wei, H.~Chen, C.~Zeng, J.~Gu, and H.~Su, ``{One-2-3-45++: Fast Single Image to 3D Objects with Consistent Multi-View Generation and 3D Diffusion},'' \emph{ArXiv}, vol. abs/2311.07885, 2023.

\bibitem{li2021hsmal}
C.~Li, N.~Ghorbani, S.~Broomé, M.~Rashid, M.~J. Black, E.~Hernlund, H.~Kjellström, and S.~Zuffi, ``{hSMAL: Detailed Horse Shape and Pose Reconstruction for Motion Pattern Recognition},'' 2021.

\bibitem{caron2021dino}
M.~Caron, H.~Touvron, I.~Misra, H.~J{\'e}gou, J.~Mairal, P.~Bojanowski, and A.~Joulin, ``{Emerging properties in self-supervised vision transformers},'' \emph{Proceedings of the IEEE/CVF international conference on computer vision}, pp. 9650--9660, 2021.

\bibitem{oquab2023dinov2}
M.~Oquab, T.~Darcet, T.~Moutakanni, H.~Vo, M.~Szafraniec, V.~Khalidov, P.~Fernandez, D.~Haziza, F.~Massa, A.~El-Nouby \emph{et~al.}, ``{DINOv2: Learning robust visual features without supervision},'' \emph{arXiv preprint arXiv:2304.07193}, 2023.

\bibitem{li2021complete}
Y.~Li, H.~Takehara, T.~Taketomi, B.~Zheng, and M.~Nie{\ss}ner, ``{4DComplete: Non-Rigid Motion Estimation Beyond the Observable Surface},'' \emph{IEEE/CVF ICCV}, pp. 12\,686--12\,696, 2021.

\bibitem{wu2023dove}
S.~Wu, T.~Jakab, C.~Rupprecht, and A.~Vedaldi, ``{DOVE}: Learning deformable 3d objects by watching videos,'' \emph{IJCV}, 2023.

\bibitem{Kanazawa2018CategorySpecific}
A.~Kanazawa, S.~Tulsiani, A.~A. Efros, and J.~Malik, ``{Learning category-specific mesh reconstruction from image collections},'' \emph{ECCV}, pp. 371--386, 2018.

\bibitem{kulkarni2020articulation}
N.~Kulkarni, A.~K. Gupta, D.~F. Fouhey, and S.~Tulsiani, ``{Articulation-Aware Canonical Surface Mapping},'' \emph{2020 IEEE/CVF Conference on Computer Vision and Pattern Recognition (CVPR)}, pp. 449--458, 2020.

\bibitem{tulsiani2020implicit}
S.~Tulsiani, N.~Kulkarni, and A.~K. Gupta, ``{Implicit Mesh Reconstruction from Unannotated Image Collections},'' \emph{ArXiv}, vol. abs/2007.08504, 2020.

\bibitem{li2020self}
X.~Li, S.~Liu, K.~Kim, S.~De~Mello, V.~Jampani, M.-H. Yang, and J.~Kautz, ``{Self-supervised single-view 3D reconstruction via semantic consistency},'' \emph{ECCV}, pp. 677--693, 2020.

\bibitem{wu2022casa}
Y.~Wu, Z.~Chen, S.~Liu, Z.~Ren, and S.~Wang, ``Casa: Category-agnostic skeletal animal reconstruction,'' \emph{Advances in Neural Information Processing Systems}, vol.~35, pp. 28\,559--28\,574, 2022.

\bibitem{ruegg2022barc}
N.~Rueegg, S.~Zuffi, K.~Schindler, and M.~J. Black, ``{BARC: Learning to Regress 3D Dog Shape from Images by Exploiting Breed Information},'' \emph{2022 IEEE/CVF Conference on Computer Vision and Pattern Recognition (CVPR)}, pp. 3866--3874, 2022.

\bibitem{Zuffi2019Safari}
S.~Zuffi, A.~Kanazawa, T.~Berger-Wolf, and M.~J. Black, ``{Three-D Safari: Learning to Estimate Zebra Pose, Shape, and Texture From Images “In the Wild”},'' \emph{IEEE/CVF ICCV}, pp. 5358--5367, 2019.

\bibitem{biggs2020who}
B.~Biggs, O.~Boyne, J.~Charles, A.~Fitzgibbon, and R.~Cipolla, ``Who left the dogs out? 3d animal reconstruction with expectation maximization in the loop,'' in \emph{European Conference on Computer Vision}.\hskip 1em plus 0.5em minus 0.4em\relax Springer, 2020, pp. 195--211.

\bibitem{liu2024lepard}
D.~Liu, A.~Stathopoulos, Q.~Zhangli, Y.~Gao, and D.~Metaxas, ``{LEPARD: Learning Explicit Part Discovery for 3D Articulated Shape Reconstruction},'' \emph{Advances in Neural Information Processing Systems}, vol.~36, 2024.

\bibitem{wu2023magicpony}
S.~Wu, R.~Li, T.~Jakab, C.~Rupprecht, and A.~Vedaldi, ``{MagicPony: Learning Articulated 3D Animals in the Wild},'' \emph{IEEE CVPR}, pp. 8792--8802, 2022.

\bibitem{lassner2021pulsar}
C.~Lassner and M.~Zollh{\"o}fer, ``Pulsar: Efficient sphere-based neural rendering,'' \emph{2021 IEEE/CVF Conference on Computer Vision and Pattern Recognition (CVPR)}, pp. 1440--1449, 2021.

\bibitem{pfister2000surfels}
H.~Pfister, M.~Zwicker, J.~van Baar, and M.~H. Gross, ``Surfels: surface elements as rendering primitives,'' \emph{Proceedings of the 27th annual conference on Computer graphics and interactive techniques}, 2000.

\bibitem{mihajlovic2021deepsurfels}
M.~Mihajlovic, S.~Weder, M.~Pollefeys, and M.~R. Oswald, ``Deepsurfels: Learning online appearance fusion,'' \emph{2021 IEEE/CVF Conference on Computer Vision and Pattern Recognition (CVPR)}, pp. 14\,519--14\,530, 2020.

\bibitem{kaye2025dualpm}
B.~Kaye, T.~Jakab, S.~Wu, C.~Rupprecht, and A.~Vedaldi, ``{DualPM: Dual Posed-Canonical Point Maps for 3D Shape and Pose Reconstruction},'' \emph{IEEE/CVF CVPR}, 2025.

\bibitem{wang2024dust3r}
S.~Wang, V.~Leroy, Y.~Cabon, B.~Chidlovskii, and J.~Revaud, ``{DUSt3R: Geometric 3d vision made easy},'' in \emph{IEEE/CVF CVPR}, 2024, pp. 20\,697--20\,709.

\bibitem{leroy2024grounding}
V.~Leroy, Y.~Cabon, and J.~Revaud, ``{Grounding image matching in 3D with MASt3R},'' in \emph{European Conference on Computer Vision}.\hskip 1em plus 0.5em minus 0.4em\relax Springer, 2024, pp. 71--91.

\bibitem{zhang2025monst3r}
J.~Zhang, C.~Herrmann, J.~Hur, V.~Jampani, T.~Darrell, F.~Cole, D.~Sun, and M.-H. Yang, ``{MONSt3R: A simple approach for estimating geometry in the presence of motion},'' \emph{ICLR}, 2025.

\bibitem{newcombe2015dynamicfusion}
R.~A. Newcombe, D.~Fox, and S.~M. Seitz, ``Dynamicfusion: Reconstruction and tracking of non-rigid scenes in real-time,'' in \emph{IEEE/CVF CVPR}, 2015, pp. 343--352.

\bibitem{zhang2023tale}
J.~Zhang, C.~Herrmann, J.~Hur, L.~Polania~Cabrera, V.~Jampani, D.~Sun, and M.-H. Yang, ``A tale of two features: Stable diffusion complements dino for zero-shot semantic correspondence,'' \emph{Advances in Neural Information Processing Systems}, vol.~36, pp. 45\,533--45\,547, 2023.

\bibitem{zhang2021ners}
J.~Zhang, G.~Yang, S.~Tulsiani, and D.~Ramanan, ``{NeRS: Neural reflectance surfaces for sparse-view 3D reconstruction in the wild},'' \emph{Advances in Neural Information Processing Systems}, vol.~34, pp. 29\,835--29\,847, 2021.

\bibitem{morreale2021neural}
L.~Morreale, N.~Aigerman, V.~G. Kim, and N.~J. Mitra, ``{Neural Surface Maps},'' \emph{IEEE CVPR}, pp. 4639--4648, 2021.

\bibitem{williamson2024neural}
R.~Williamson and N.~J. Mitra, ``{Neural Geometry Processing via Spherical Neural Surfaces},'' in \emph{Computer Graphics Forum}.\hskip 1em plus 0.5em minus 0.4em\relax Wiley Online Library, 2024, p. e70021.

\bibitem{nizamani2025dynamic}
A.~Nizamani, H.~Laga, G.~Wang, F.~Boussaid, M.~Bennamoun, and A.~Srivastava, ``{Dynamic Neural Surfaces for Elastic 4D Shape Representation and Analysis},'' \emph{IEEE CVPR}, 2025.

\bibitem{zuffi2017menagerie}
S.~Zuffi, A.~Kanazawa, D.~W. Jacobs, and M.~J. Black, ``{3D Menagerie: Modeling the 3D Shape and Pose of Animals},'' \emph{2017 IEEE Conference on Computer Vision and Pattern Recognition (CVPR)}, pp. 5524--5532, 2017.

\bibitem{zuffi2024varen}
S.~Zuffi, Y.~Mellbin, C.~Li, M.~Hoeschle, H.~Kjellstr{\"o}m, S.~Polikovsky, E.~Hernlund, and M.~J. Black, ``Varen: Very accurate and realistic equine network,'' in \emph{IEEE/CVF CVPR}, 2024, pp. 5374--5383.

\bibitem{yang2021lasr}
G.~Yang, D.~Sun, V.~Jampani, D.~Vlasic, F.~Cole, H.~Chang, D.~Ramanan, W.~T. Freeman, and C.~Liu, ``{LASR: Learning Articulated Shape Reconstruction from a Monocular Video},'' \emph{IEE/CVF CVPR}, pp. 15\,975--15\,984, 2021.

\bibitem{aygun2024saor}
M.~Aygun and O.~M. Aodha, ``{SAOR: Single-View Articulated Object Reconstruction},'' \emph{IEEE/CVF CVPR}, 2024.

\bibitem{badger2020bird}
M.~Badger, Y.~Wang, A.~Modh, A.~Perkes, N.~Kolotouros, B.~Pfrommer, M.~F. Schmidt, and K.~Daniilidis, ``{3D Bird Reconstruction: a Dataset, Model, and Shape Recovery from a Single View},'' \emph{ECCV}, vol. 12363, pp. 1--17, 2020.

\bibitem{kokkinos2022learning}
F.~Kokkinos and I.~Kokkinos, ``{Learning monocular 3D reconstruction of articulated categories from motion},'' \emph{IEEE/CVF CVPR}, pp. 1737--1746, 2021.

\bibitem{sorkine2004laplacian}
O.~Sorkine, D.~Cohen-Or, Y.~Lipman, M.~Alexa, C.~R{\"o}ssl, and H.-P. Seidel, ``Laplacian surface editing,'' in \emph{Proceedings of the 2004 Eurographics/ACM SIGGRAPH symposium on Geometry processing}, 2004, pp. 175--184.

\bibitem{sorkine2007rigid}
O.~Sorkine and M.~Alexa, ``As-rigid-as-possible surface modeling,'' \emph{Eurographics Association}, p. 109–116, 2007.

\bibitem{li2020online}
X.~Li, S.~Liu, S.~De~Mello, K.~Kim, X.~Wang, M.-H. Yang, and J.~Kautz, ``{Online adaptation for consistent mesh reconstruction in the wild},'' \emph{Advances in Neural Information Processing Systems}, vol.~33, pp. 15\,009--15\,019, 2020.

\bibitem{Cootes1995ASM}
T.~Cootes, C.~Taylor, D.~Cooper, and J.~Graham, ``Active shape models-their training and application,'' \emph{CVIU}, vol.~61, no.~1, pp. 38 -- 59, 1995.

\bibitem{cootes:eccv1998}
T.~F. Cootes, G.~J. Edwards, and C.~J. Taylor, ``Active appearance models,'' \emph{Computer Vision—ECCV’98: 5th European Conference on Computer Vision Freiburg, Germany, June 2--6, 1998 Proceedings, Volume II 5}, pp. 484--498, 1998.

\bibitem{cootes:pami2001}
T.~Cootes, G.~Edwards, and C.~Taylor, ``Active appearance models,'' \emph{IEEE PAMI}, vol.~23, no.~6, pp. 681--685, 2001.

\bibitem{allen2003space}
B.~Allen, B.~Curless, and Z.~Popovi{\'c}, ``The space of human body shapes: reconstruction and parameterization from range scans,'' \emph{ACM TOG}, vol.~22, no.~3, pp. 587--594, 2003.

\bibitem{Pavlakos_2019_CVPR}
G.~Pavlakos, V.~Choutas, N.~Ghorbani, T.~Bolkart, A.~A.~A. Osman, D.~Tzionas, and M.~J. Black, ``{Expressive Body Capture: 3D Hands, Face, and Body From a Single Image},'' \emph{IEEE CVPR}, June 2019.

\bibitem{ruegg2023bite}
N.~R{\"u}egg, S.~Tripathi, K.~Schindler, M.~J. Black, and S.~Zuffi, ``{BITE: Beyond priors for improved three-D dog pose estimation},'' in \emph{IEEE/CVF CVPR}, 2023, pp. 8867--8876.

\bibitem{tan2024distilling}
J.~Tan, G.~Yang, and D.~Ramanan, ``{Distilling Neural Fields for Real-Time Articulated Shape Reconstruction},'' \emph{IEEE/CVF CVPR}, pp. 4692--4701, 2023.

\bibitem{baieri2025model}
D.~Baieri, R.~Cicciarella, M.~Kr{\"u}tzen, E.~Rodol{\`a}, and S.~Zuffi, ``Model-based metric 3d shape and motion reconstruction of wild bottlenose dolphins in drone-shot videos,'' \emph{arXiv preprint arXiv:2504.15782}, 2025.

\bibitem{Zuffi2018LionsTigersBears}
S.~Zuffi, A.~Kanazawa, and M.~J. Black, ``{Lions and Tigers and Bears: Capturing Non-Rigid, 3D, Articulated Shape from Images},'' \emph{IEEE Conference on Computer Vision and Pattern Recognition (CVPR)}, 2018.

\bibitem{laga2018survey}
H.~Laga, ``{A survey on nonrigid 3D shape analysis},'' \emph{Academic Press Library in Signal Processing, Volume 6}, pp. 261--304, 2018.

\bibitem{li2021coarse}
C.~Li and G.~H. Lee, ``{Coarse-to-fine Animal Pose and Shape Estimation},'' \emph{Advances in Neural Information Processing Systems}, vol.~34, pp. 11\,757--11\,768, 2021.

\bibitem{lewis2023pose}
J.~P. Lewis, M.~Cordner, and N.~Fong, ``Pose space deformation: a unified approach to shape interpolation and skeleton-driven deformation,'' in \emph{Seminal Graphics Papers: Pushing the Boundaries, Volume 2}, 2023, pp. 811--818.

\bibitem{liao2022skeleton}
Z.~Liao, J.~Yang, J.~Saito, G.~Pons-Moll, and Y.~Zhou, ``Skeleton-free pose transfer for stylized 3d characters,'' in \emph{European Conference on Computer Vision}.\hskip 1em plus 0.5em minus 0.4em\relax Springer, 2022, pp. 640--656.

\bibitem{taigman2014deepface}
Y.~Taigman, M.~Yang, M.~Ranzato, and L.~Wolf, ``Deepface: Closing the gap to human-level performance in face verification,'' in \emph{IEEE/CVF CVPR}, 2014, pp. 1701--1708.

\bibitem{schroff2015facenet}
F.~Schroff, D.~Kalenichenko, and J.~Philbin, ``Facenet: A unified embedding for face recognition and clustering,'' in \emph{IEEE/CVF CVPR}, 2015, pp. 815--823.

\bibitem{bogo2016keep}
F.~Bogo, A.~Kanazawa, C.~Lassner, P.~Gehler, J.~Romero, and M.~J. Black, ``{Keep it SMPL: Automatic estimation of 3D human pose and shape from a single image},'' \emph{ECCV}, pp. 561--578, 2016.

\bibitem{lyu2025animer}
J.~Lyu, T.~Zhu, Y.~Gu, L.~Lin, P.~Cheng, Y.~Liu, X.~Tang, and L.~An, ``Animer: Animal pose and shape estimation using family aware transformer,'' in \emph{Proceedings of the Computer Vision and Pattern Recognition Conference}, 2025, pp. 17\,486--17\,496.

\bibitem{goel2023humans}
S.~Goel, G.~Pavlakos, J.~Rajasegaran, A.~Kanazawa, and J.~Malik, ``Humans in 4d: Reconstructing and tracking humans with transformers,'' in \emph{Proceedings of the IEEE/CVF International Conference on Computer Vision}, 2023, pp. 14\,783--14\,794.

\bibitem{vaswani2017attention}
A.~Vaswani, N.~Shazeer, N.~Parmar, J.~Uszkoreit, L.~Jones, A.~N. Gomez, {\L}.~Kaiser, and I.~Polosukhin, ``Attention is all you need,'' \emph{Advances in neural information processing systems}, vol.~30, 2017.

\bibitem{mildenhall2020nerf}
B.~Mildenhall, P.~P. Srinivasan, M.~Tancik, J.~T. Barron, R.~Ramamoorthi, and R.~Ng, ``{NeRF: Representing Scenes as Neural Radiance Fields for View Synthesis},'' in \emph{ECCV}.\hskip 1em plus 0.5em minus 0.4em\relax Springer, 2020, pp. 405--421.

\bibitem{mildenhall2021nerf}
B.~Mildenhall, P.~P. Srinivasan, M.~Tancik, J.~T. Barron, R.~Ramamoorthi, and R.~Ng, ``Nerf: Representing scenes as neural radiance fields for view synthesis,'' \emph{Communications of the ACM}, vol.~65, no.~1, pp. 99--106, 2021.

\bibitem{wang2021neus}
P.~Wang, L.~Liu, Y.~Liu, C.~Theobalt, T.~Komura, and W.~Wang, ``Neus: Learning neural implicit surfaces by volume rendering for multi-view reconstruction,'' \emph{NeurIPS}, 2021.

\bibitem{carr2001reconstruction}
J.~C. Carr, R.~K. Beatson, J.~B. Cherrie, T.~J. Mitchell, W.~R. Fright, B.~C. McCallum, and T.~R. Evans, ``Reconstruction and representation of 3d objects with radial basis functions,'' in \emph{Proceedings of the 28th annual conference on Computer graphics and interactive techniques}, 2001, pp. 67--76.

\bibitem{kojekine2004surface}
N.~Kojekine, V.~Savchenko, and I.~Hagiwara, ``Surface reconstruction based on compactly supported radial basis functions,'' in \emph{Geometric modeling: techniques, applications, systems and tools}.\hskip 1em plus 0.5em minus 0.4em\relax Springer, 2004, pp. 217--231.

\bibitem{ohtake2005multi}
Y.~Ohtake, A.~Belyaev, M.~Alexa, G.~Turk, and H.-P. Seidel, ``Multi-level partition of unity implicits,'' in \emph{Acm Siggraph 2005 Courses}, 2005, pp. 173--es.

\bibitem{piperakis2001affine}
E.~Piperakis and I.~Kumazawa, ``{Affine transformations of 3D objects represented with neural networks},'' in \emph{Proceedings Third International Conference on 3-D Digital Imaging and Modeling}.\hskip 1em plus 0.5em minus 0.4em\relax IEEE, 2001, pp. 213--223.

\bibitem{piperakis20013d}
E.~Piperakis, I.~Kumazawa, and R.~Piperakis, ``3d object \& light source representation with multi layer feed forward networks,'' \emph{Neural, Parallel \& Scientific Computations}, vol.~9, no.~2, pp. 161--173, 2001.

\bibitem{Mescheder2018OccupancyNL}
L.~M. Mescheder, M.~Oechsle, M.~Niemeyer, S.~Nowozin, and A.~Geiger, ``Occupancy networks: Learning 3d reconstruction in function space,'' \emph{2019 IEEE/CVF Conference on Computer Vision and Pattern Recognition (CVPR)}, pp. 4455--4465, 2018.

\bibitem{shen2021deep}
T.~Shen, J.~Gao, K.~Yin, M.-Y. Liu, and S.~Fidler, ``{Deep marching tetrahedra: a hybrid representation for high-resolution 3d shape synthesis},'' \emph{Advances in Neural Information Processing Systems}, vol.~34, pp. 6087--6101, 2021.

\bibitem{Park2019DeepSDFLC}
J.~J. Park, P.~R. Florence, J.~Straub, R.~A. Newcombe, and S.~Lovegrove, ``Deepsdf: Learning continuous signed distance functions for shape representation,'' \emph{2019 IEEE/CVF Conference on Computer Vision and Pattern Recognition (CVPR)}, pp. 165--174, 2019.

\bibitem{jiang2023consistent4d}
Y.~Jiang, L.~Zhang, J.~Gao, W.~Hu, and Y.~Yao, ``{Consistent4D: Consistent 360{\textdegree} Dynamic Object Generation from Monocular Video},'' \emph{The Twelfth International Conference on Learning Representations}, 2024.

\bibitem{lei2022cadex}
J.~Lei and K.~Daniilidis, ``{CaDeX: Learning Canonical Deformation Coordinate Space for Dynamic Surface Representation via Neural Homeomorphism},'' \emph{IEEE/CVF CVPR}, pp. 6614--6624, 2022.

\bibitem{dinh2016density}
L.~Dinh, J.~Sohl-Dickstein, and S.~Bengio, ``Density estimation using real nvp,'' \emph{arXiv preprint arXiv:1605.08803}, 2016.

\bibitem{dinh2014nice}
L.~Dinh, D.~Krueger, and Y.~Bengio, ``Nice: Non-linear independent components estimation,'' \emph{arXiv preprint arXiv:1410.8516}, 2014.

\bibitem{fridovich2023k}
S.~Fridovich-Keil, G.~Meanti, F.~R. Warburg, B.~Recht, and A.~Kanazawa, ``{K-planes: Explicit radiance fields in space, time, and appearance},'' in \emph{IEEE/CVF CVPR}, 2023, pp. 12\,479--12\,488.

\bibitem{kerbl20233d}
B.~Kerbl, G.~Kopanas, T.~Leimk{\"u}hler, and G.~Drettakis, ``{3D gaussian splatting for real-time radiance field rendering},'' \emph{ACM Trans. Graph.}, vol.~42, no.~4, pp. 139--1, 2023.

\bibitem{lei2024gart}
J.~Lei, Y.~Wang, G.~Pavlakos, L.~Liu, and K.~Daniilidis, ``{GART: Gaussian Articulated Template Models},'' in \emph{IEEE/CVF CVPR}, 2024, pp. 19\,876--19\,887.

\bibitem{zwicker2002ewa}
M.~Zwicker, H.~Pfister, J.~Van~Baar, and M.~Gross, ``Ewa splatting,'' \emph{IEEE Transactions on Visualization and Computer Graphics}, vol.~8, no.~3, pp. 223--238, 2002.

\bibitem{yang2023real}
Z.~Yang, H.~Yang, Z.~Pan, and L.~Zhang, ``Real-time photorealistic dynamic scene representation and rendering with 4d gaussian splatting,'' \emph{arXiv preprint arXiv:2310.10642}, 2023.

\bibitem{luiten2024dynamic}
J.~Luiten, G.~Kopanas, B.~Leibe, and D.~Ramanan, ``Dynamic 3d gaussians: Tracking by persistent dynamic view synthesis,'' in \emph{2024 International Conference on 3D Vision (3DV)}.\hskip 1em plus 0.5em minus 0.4em\relax IEEE, 2024, pp. 800--809.

\bibitem{wu20244d}
G.~Wu, T.~Yi, J.~Fang, L.~Xie, X.~Zhang, W.~Wei, W.~Liu, Q.~Tian, and X.~Wang, ``4d gaussian splatting for real-time dynamic scene rendering,'' in \emph{Proceedings of the IEEE/CVF conference on computer vision and pattern recognition}, 2024, pp. 20\,310--20\,320.

\bibitem{yang2024deformable}
Z.~Yang, X.~Gao, W.~Zhou, S.~Jiao, Y.~Zhang, and X.~Jin, ``Deformable 3d gaussians for high-fidelity monocular dynamic scene reconstruction,'' in \emph{Proceedings of the IEEE/CVF conference on computer vision and pattern recognition}, 2024, pp. 20\,331--20\,341.

\bibitem{cho2025dogrecon}
G.~Cho, C.~Kang, D.~Soon, and K.~Joo, ``{DogRecon: Canine Prior-Guided Animatable 3D Gaussian Dog Reconstruction From A Single Image},'' \emph{International Journal of Computer Vision}, pp. 1--15, 2025.

\bibitem{zhai2025taga}
Z.~Zhai, G.~Chen, W.~Wang, D.~Zheng, and J.~Xiao, ``{TAGA: Self-supervised Learning for Template-free Animatable Gaussian Articulated Model},'' in \emph{Proceedings of the Computer Vision and Pattern Recognition Conference}, 2025, pp. 21\,159--21\,169.

\bibitem{yang2024omnimotiongpt}
Z.~Yang, M.~Zhou, M.~Shan, B.~Wen, Z.~Xuan, M.~Hill, J.~Bai, G.-J. Qi, and Y.~Wang, ``{OmniMotionGPT: Animal Motion Generation with Limited Data},'' \emph{IEEE/CVF CVPR}, pp. 1249--1259, 2024.

\bibitem{wang2025animo}
X.~Wang, K.~Ruan, X.~Zhang, and G.~Wang, ``{AniMo: Species-Aware Model for Text-Driven Animal Motion Generation},'' in \emph{IEEE/CVF CVPR}, 2025, pp. 1929--1939.

\bibitem{radford2021learning}
A.~Radford, J.~W. Kim, C.~Hallacy, A.~Ramesh, G.~Goh, S.~Agarwal, G.~Sastry, A.~Askell, P.~Mishkin, J.~Clark \emph{et~al.}, ``Learning transferable visual models from natural language supervision,'' in \emph{International conference on machine learning}.\hskip 1em plus 0.5em minus 0.4em\relax PmLR, 2021, pp. 8748--8763.

\bibitem{tosca2008}
A.~M. Bronstein, M.~M. Bronstein, and R.~Kimmel, \emph{Numerical geometry of non-rigid shapes}.\hskip 1em plus 0.5em minus 0.4em\relax Springer Science \& Business Media, 2008.

\bibitem{tigdog2016}
L.~Del~Pero, S.~Ricco, R.~Sukthankar, and V.~Ferrari, ``Articulated motion discovery using pairs of trajectories,'' in \emph{IEEE/CVF CVPR (CVPR)}, 2015.

\bibitem{animalpose2019}
J.~Cao, H.~Tang, H.-S. Fang, X.~Shen, C.~Lu, and Y.-W. Tai, ``Cross-domain adaptation for animal pose estimation,'' in \emph{Proceedings of the IEEE/CVF International Conference on Computer Vision}, 2019, pp. 9498--9507.

\bibitem{rgbddog2020}
S.~Kearney, W.~Li, M.~Parsons, K.~I. Kim, and D.~Cosker, ``Rgbd-dog: Predicting canine pose from rgbd sensors,'' in \emph{IEEE/CVF Conference on Computer Vision and Pattern Recognition (CVPR)}, June 2020.

\bibitem{ap10k2021}
H.~Yu, Y.~Xu, J.~Zhang, W.~Zhao, Z.~Guan, and D.~Tao, ``Ap-10k: A benchmark for animal pose estimation in the wild,'' in \emph{Thirty-fifth Conference on Neural Information Processing Systems Datasets and Benchmarks Track (Round 2)}, 2021.

\bibitem{horse30_2021}
A.~Mathis, T.~Biasi, S.~Schneider, M.~Yuksekgonul, B.~Rogers, M.~Bethge, and M.~W. Mathis, ``Pretraining boosts out-of-domain robustness for pose estimation,'' in \emph{Proceedings of the IEEE/CVF winter conference on applications of computer vision}, 2021, pp. 1859--1868.

\bibitem{sydog2021}
M.~Shooter, C.~Malleson, and A.~Hilton, ``Sydog: A synthetic dog dataset for improved 2d pose estimation,'' \emph{arXiv preprint arXiv:2108.00249}, 2021.

\bibitem{acinoset2021}
D.~Joska, L.~Clark, N.~Muramatsu, R.~Jericevich, F.~Nicolls, A.~Mathis, M.~W. Mathis, and A.~Patel, ``Acinoset: a 3d pose estimation dataset and baseline models for cheetahs in the wild,'' in \emph{2021 IEEE international conference on robotics and automation (ICRA)}.\hskip 1em plus 0.5em minus 0.4em\relax IEEE, 2021, pp. 13\,901--13\,908.

\bibitem{apt36k2022}
Y.~Yang, J.~Yang, Y.~Xu, J.~Zhang, L.~Lan, and D.~Tao, ``Apt-36k: A large-scale benchmark for animal pose estimation and tracking,'' \emph{Advances in Neural Information Processing Systems}, vol.~35, pp. 17\,301--17\,313, 2022.

\bibitem{ng2022animal}
X.~L. Ng, K.~E. Ong, Q.~Zheng, Y.~Ni, S.~Y. Yeo, and J.~Liu, ``Animal kingdom: A large and diverse dataset for animal behavior understanding,'' in \emph{Proceedings of the IEEE/CVF conference on computer vision and pattern recognition}, 2022, pp. 19\,023--19\,034.

\bibitem{luo2022artemis}
H.~Luo, T.~Xu, Y.~Jiang, C.~Zhou, Q.~Qiu, Y.~Zhang, W.~Yang, L.~Xu, and J.~Yu, ``{Artemis: articulated neural pets with appearance and motion synthesis},'' \emph{ACM Transactions on Graphics (TOG)}, vol.~41, no.~4, pp. 1--19, 2022.

\bibitem{Xu2023Animal3DAC}
J.~Xu, Y.~Zhang, J.-X. Peng, W.~Ma, A.~Jesslen, P.~Ji, Q.~Hu, J.~Zhang, Q.~Liu, J.~Wang, W.~Ji, C.~Wang, X.~Yuan, P.~Kaushik, G.~Zhang, J.~Liu, Y.~Xie, Y.~Cui, A.~L. Yuille, and A.~Kortylewski, ``Animal3d: A comprehensive dataset of 3d animal pose and shape,'' \emph{IEEE/CVF ICCV}, pp. 9065--9075, 2023.

\bibitem{mvsydog2023}
M.~Shooter, C.~Malleson, and A.~Hilton, ``Digidogs: Single-view 3d pose estimation of dogs using synthetic training data,'' in \emph{Proceedings of the IEEE/CVF Winter Conference on Applications of Computer Vision}, 2024, pp. 80--89.

\bibitem{sydogvideo2023}
M.~Shooter, C.~Malleson, and A.~Hilton, ``Sydog-video: A synthetic dog video dataset for temporal pose estimation,'' \emph{International Journal of Computer Vision}, vol. 132, no.~6, pp. 1986--2002, 2024.

\bibitem{digitallife3d}
{Digital Life Project}, ``{Digital Life 3D},'' \url{https://digitallife3d.org/}.

\bibitem{digidogs2024}
M.~Shooter, C.~Malleson, and A.~Hilton, ``Digidogs: Single-view 3d pose estimation of dogs using synthetic training data,'' in \emph{Proceedings of the IEEE/CVF Winter Conference on Applications of Computer Vision}, 2024, pp. 80--89.

\bibitem{biggs2019creatures}
B.~Biggs, T.~Roddick, A.~Fitzgibbon, and R.~Cipolla, ``{Creatures great and smal: Recovering the shape and motion of animals from video},'' \emph{Asian Conference on Computer Vision}, pp. 3--19, 2019.

\bibitem{zeng2024stag4d}
Y.~Zeng, Y.~Jiang, S.~Zhu, Y.~Lu, Y.~Lin, H.~Zhu, W.~Hu, X.~Cao, and Y.~Yao, ``{STAG4D: Spatial-Temporal Anchored Generative 4D Gaussians},'' \emph{ECCV}, pp. 163--179, 2024.

\bibitem{hunyuan3d22025tencent}
T.~H. Team, ``Hunyuan3d 2.0: Scaling diffusion models for high resolution textured 3d assets generation,'' 2025.

\bibitem{kulits2025reconstructing}
P.~Kulits, M.~J. Black, and S.~Zuffi, ``{Reconstructing Animals and the Wild},'' in \emph{IEEE/CVF CVPR}, 2025, pp. 16\,565--16\,577.

\end{thebibliography}
